\documentclass[conference]{IEEEtran}
\usepackage{times}
\usepackage{graphicx}
\usepackage{subfig}
\usepackage{amsmath}
\usepackage{amssymb}
\usepackage{multirow}
\usepackage{multicol}
\usepackage{balance}
\usepackage{csquotes}
\usepackage{tikz}
\usepackage{booktabs}
\usepackage{comment}
\usepackage{stfloats}
\usepackage[english]{babel}
\usepackage{amsthm}
\theoremstyle{definition}
\newtheorem{definition}{Definition}

\usepackage{float}
\usepackage{algorithm}
\usepackage{algpseudocode}
\usepackage{mathtools}
\usepackage{balance}

\usepackage[numbers]{natbib}
\usepackage{multicol}
\usepackage[bookmarks=true]{hyperref}
\usepackage{soul}
\usepackage{xcolor}

\usepackage{thmtools}
\usepackage{thm-restate}

\newtheorem{assumption}{Assumption}

\renewcommand{\d}{\mathrm{d}}
\newcommand{\E}{\mathbb{E}}
\newcommand{\epsmin}{\epsilon_{\min}}
\newcommand{\epsmax}{\epsilon_{\max}}
\newcommand{\KL}{\mathbb{KL}}
\renewcommand{\H}{\mathbb{H}}

\newcommand{\method}{\textsc{BRIDGeR}}
\newcommand{\para}[1]{\vspace{0.4em}\noindent\textbf{#1}}

\DeclarePairedDelimiter{\norm}{\lVert}{\rVert}

\begin{document}

\title{Don't Start from Scratch: Behavioral Refinement via Interpolant-based Policy Diffusion}
\author{Kaiqi Chen$^{1}$, Eugene Lim$^{1}$, Kelvin Lin$^{1}$, Yiyang Chen$^{1}$, and Harold Soh$^{1,2}$\\$^{1}$Dept. of Computer Science, National University of Singapore.
\\ $^{2}$Smart Systems Institute, NUS.
\\{\small Contact Authors: {\texttt{\{kaiqi, harold\}@comp.nus.edu.sg}}}
}

\maketitle

\begin{abstract}
Imitation learning empowers artificial agents to mimic behavior by learning from demonstrations. Recently, diffusion models, which have the ability to model high-dimensional and multimodal distributions, have shown impressive performance on imitation learning tasks. These models learn to shape a policy by diffusing actions (or states) from standard Gaussian noise. However, the target policy to learn is often significantly different from Gaussian and this mismatch can result in poor performance when using a small number of diffusion steps (to improve inference speed) and under limited data. The key idea in this work is that initiating from a more informative source than Gaussian enables diffusion methods to mitigate the above limitations. We contribute both theoretical results, a new method, and empirical findings that show the benefits of using an informative source policy. Our method, which we call \method, leverages the stochastic interpolants framework to bridge arbitrary policies, thus enabling a flexible approach towards imitation learning. It generalizes prior work in that standard Gaussians can still be applied, but other source policies can be used if available. In experiments on challenging simulation benchmarks and on real robots, \method{} outperforms state-of-the-art diffusion policies. We provide further analysis on design considerations when applying \method. Code for \method{} is available at \url{https://github.com/clear-nus/bridger}.
\end{abstract}

\IEEEpeerreviewmaketitle

\section{Introduction}
\label{sec:intro}

Imitation learning enables robots to learn policies from demonstrations and has been applied to a variety of domains including manipulation \cite{rahmatizadeh2018vision, zhang2018deep, florence2019self}, autonomous driving \cite{pomerleau1988alvinn, bojarski2016end}, and shared autonomy \cite{schaff2020residual, yoneda2023noise}. Recently, there has been a significant interest in the adaptation of diffusion models for imitation learning~\cite{chi2023diffusion, reuss2023goal, hu2023rf}. These deep generative models, which progressively transform Gaussian noise to a policy over a number of diffusion steps, offer practical advantages over classical techniques \cite{shafiullah2022behavior, sharma2018multiple} --- they scale well with the number of dimensions in the action/state spaces (e.g., for visuo-motor learning on a 7-DoF robot arm \cite{chi2023diffusion}) and are able to capture complex multimodal distributions. However, current diffusion methods also require large training datasets and typically have long inference times due to the number of diffusion steps needed to obtain effective action distributions for complex tasks~\cite{reuss2023goal}.

\begin{figure}
    \centering
\includegraphics[width=0.90\columnwidth]{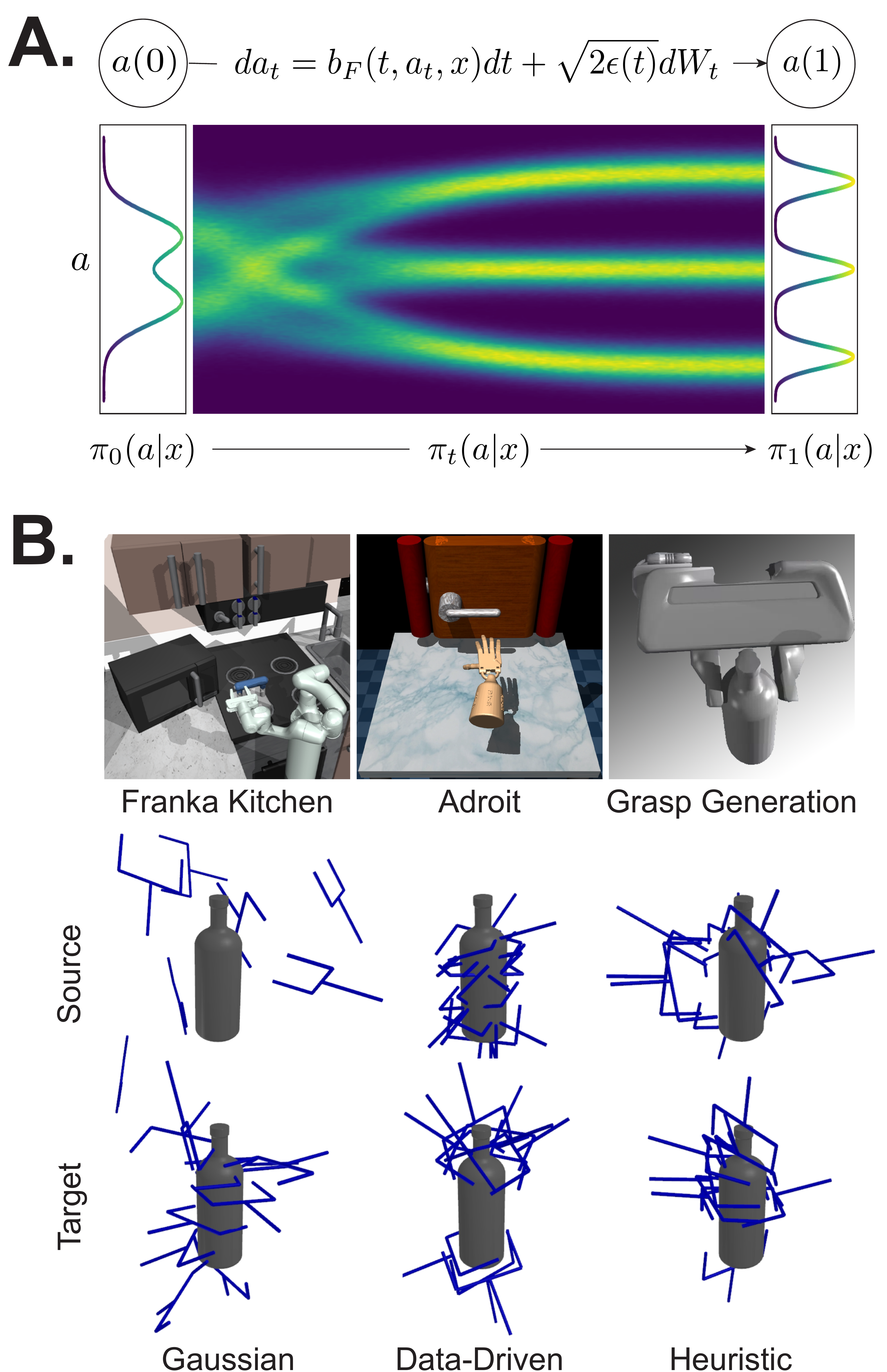}
    \caption{\small (\textbf{A}) Overview of action generation with \method. With trained velocity $b$ and score $s$ functions, \method{} transports the actions from source distribution $\pi_0(a|x)$ to the target distribution $\pi_1(a|x)$ via the forward SDE (Eq. \ref{eq:fsde}). (\textbf{B}) We tested \method{} on challenging robot benchmark tasks and show that using informative source policies enhances performance. For example, in 6-DoF grasp generation, using heuristic or data-driven source policies results in more successful grasps compared to the conventional Gaussian. 
    }
    \label{fig:method}
    \vspace{-1.7em}
\end{figure}

An examination of existing diffusion-style imitation learning reveals a fundamental issue: these models learn to shape a policy starting from standard Gaussian noise, which is often starkly different from the intended policy or action distribution. The key insight in our work is that initiating from Gaussian noise isn't a prerequisite. To explore this, we move beyond the conventional diffusion framework and employ stochastic interpolants~\cite{albergo2023stochastic} for bridging arbitrary densities within finite time (Fig. \ref{fig:method}). This approach allows us to leverage stochastic source policies, enabling the diffusion process to begin from a more informative starting point. Example source policies include  policies hand-crafted using prior knowledge of the task or data-driven policies trained on a similar task.
We find that this shift retains the inherent advantages of diffusion-style imitation learning, but positively impacts inference time and performance. Practically, this leads to faster generation and more accurate robot actions. Our approach also  generalizes prior work in diffusion-based imitation learning since if no source policy is available, simple distributions such as the Gaussian can be used.

In this paper, we first contribute a theoretical analysis of the impact of different source policies in diffusion. In brief, we find that under reasonable assumptions, selecting a better source policy results in better target policies. We then turn to a practical approach for incorporating source policies into diffusion methods. Applying the stochastic interpolants framework~\cite{albergo2023stochastic} to imitation learning, we derive a new method called \method{} (\textbf{B}ehavioral \textbf{R}efinement via \textbf{I}nterpolant-based \textbf{D}iffusion for \textbf{Ge}nerative \textbf{R}obotics). To our knowledge, our work is the first adaptation of this bridging methodology to imitation learning, contrasting with its previous use in simple synthetic tasks \cite{zhou2023denoising} and image generation \cite{zhou2023denoising, liu20232}. In addition to standard neural architecture design for the learnt forward model, the stochastic interpolant framework relies on several critical design choices including the source policy and interpolant. The interpolant dictates how a sampled point transitions from the source to the target distribution, with the transition modulated by noise introduced through time-dependent Gaussian latent variables~\cite{albergo2023stochastic}. Intuitively, the interpolant forms a ``bridge'' or ``guide'' between two policies (e.g., it gradually changes poor robot actions into better ones). 

We contribute a systematic empirical study of the effects of using source policies (and other design elements) on a diverse set of robot tasks, including the Franka kitchen benchmark, grasp generation, and manipulation using a robot hand. Overall, the experimental results coincide with our theoretical findings; Gaussians were seldom the most effective source distribution and surprisingly, even simple heuristic distributions resulted in superior learnt policies compared to the Gaussian. We demonstrate that given a good source policy, \method{} surpasses existing state-of-the-art diffusion policies. Additionally, we discuss the effects of the interpolant function when learning highly multi-modal behaviors. Similar positive results were observed on real-world experiments using two robots: a Franka Emika Panda arm with a two-finger gripper for stable grasping, and a UR5e equipped with a Shadow Dexterous Hand Lite for synthetic wound cleaning. These tasks involved real-world high-dimensional observations (e.g., point clouds and images) and complex actions (22 action dimensions per time-step for the wound cleaning task).

In summary, our work connects distribution bridging to imitation learning, which results in improved performance and addresses inherent limitations of standard diffusion, such as lengthy inference times. We contribute:
\begin{itemize}
\item Theoretical results on the impact of diffusing from source policies of varying quality;
\item A practical method that enables source policies to be used for diffusion-based imitation learning, which leads to better trade-offs between inference speed and performance;
\item A comprehensive empirical study demonstrating the impact of source distributions and interpolant design on outcome quality across various robot tasks.
\end{itemize}
From a broader perspective, our research demonstrates the potential of bridging models in imitation learning. We hope that this work lays the foundation for future imitation learning methods that leverage past policies for lifelong robot learning. 
\section{Preliminaries: Background \& Related Work}
\subsection{Problem Formulation}

In imitation learning~\cite{hussein2017imitation,zheng2022imitation,wang2020generalizing}, we wish to learn a policy from expert demonstrations. 
Let $\pi_1(a|x)$ be an expert policy; it  captures the probability of an expert selecting an action $a$ given an observation $x$. Suppose we have a dataset $\mathcal{D} = \{\smash{x^{(i)}}, \smash{a^{(i)}}\}_{i=1}^N$ drawn from $\pi_1(a|x)$ and let $\pi_0(a|x)$ be a distribution that we can easily sample $a$ from. 

Our goal is to learn a model for transporting actions drawn from $\pi_0$ to $\pi_1$. More concretely, given an observation $x$, let $\pi_t(a|x)$ over time $t\in[0,1]$ be distributions on the \emph{bridge} that links $\pi_{0}(a|x)$ and $\pi_{1}(a|x)$. Let $\hat{\pi}_t$ be the density of the learned distribution over time $t\in[0,1]$. We want the generated target density $\hat{\pi}_1$ to match the ground truth density $\pi_1$.

Similar to recent research~\cite{chi2023diffusion}, our approach primarily generates \emph{action sequences} rather than single-step actions. Upon completing (or partially completing) an action sequence, the model assimilates a new observation to generate the subsequent action sequence. In the following, $a$ denotes an action sequence or a target pose, depending on the context.

\subsection{Diffusion-based Policy Learning}

Diffusion-based policy methods \cite{chi2023diffusion, reuss2023goal} are largely based on Denoising Diffusion Probabilistic Models (DDPM)~\cite{ho2020denoising}. These methods operate by progressively adding noise to an action $a_1$ during a forward process and subsequently, employing a reverse process to learn how to denoise. 
This is achieved by training a neural network $g_\upsilon(a_{t_k}, t_k)$ to predict the noise $z\sim\mathcal{N}(0, I)$ added to the  data sample, within a defined temporal sequence $0=t_0<t_1<\dots<t_{K-1}<t_K=1$. The training utilizes a regression loss,
\begin{equation}
    \mathcal{L}(\upsilon) = \mathbb{E}\left[\norm{z - g_\upsilon(\sqrt{\Bar{\alpha}_k}a_1+\sqrt{1-\Bar{\alpha}_k}z, t_k)}^2\right].
\end{equation}

Upon completion of training, DDPM utilizes initial samples $a_0$ drawn from a standard Gaussian distribution and applies a $K$-step denoising procedure to synthesize the desired output $a_1$,
\begin{equation}
    a_{t_k} = \frac{1}{\sqrt{\alpha_k}}\left(a_{t_{k-1}}-\frac{1-\alpha_k}{\sqrt{1-\Bar{\alpha}_k}}g_\upsilon(a_{t_{k}}, t_{k})\right) + \sigma_k\mathcal{N}(0, I).
\end{equation}

One limitation of the original DDPM is slow sampling; the number of steps $K$ required is typically large (e.g., hundreds to thousands). To improve sample efficiency, diffusion-based policy methods \cite{chi2023diffusion, reuss2023goal} employ Denoising Diffusion Implicit Models (DDIM)~\cite{du2019implicit}, which apply a \emph{deterministic} non-Markovian process to trade-off between sample quality and inference speed. 

\subsection{Related Work}
\label{sec:related_work}

\method{} builds upon a large body of work in imitation learning, specifically behavior cloning. Behavior cloning has had a long history, with early methods adopting supervised learning techniques, principally regression~\cite{bojarski2016end, rahmatizadeh2018vision, rahmatizadeh2018vision, zhang2018deep, florence2019self, ross2011reduction, toyer2020magical}. Although efficient, regression-based methods were generally unable to capture multi-modal behavior. Later methods used classification methods, by discretizing the action space~\cite{shafiullah2022behavior, sharma2018multiple, avigalspeedfolding, wu2020spatial}, to address this limitation but were sensitive to hyperparameters (such as the level of discretization) and struggled with high-precision tasks~\cite{florence2022implicit}. 

Modern imitation methods employ generative models, starting with Gaussian Mixture Models~\cite{mandlekar2022matters}, then transitioning to more powerful Energy-Based Models (EBMs) \cite{datta2023iifl, florence2022implicit, jarrett2020strictly, singh2023revisiting} and Diffusion Models \cite{scheikl2023movement, chi2023diffusion, pearce2023imitating, ng2023diffusion}. Compared to EBMs, diffusion models have demonstrated better training stability~\cite{chi2023diffusion} and effectiveness in generating consistent, multi-modal action sequences for visual-motor control~\cite{scheikl2023movement, chi2023diffusion, pearce2023imitating, ng2023diffusion} and hierarchical, long-horizon planning tasks~\cite{reuss2023multimodal, ha2023scaling, mishra2023generative, chen2023playfusion, xian2023chaineddiffuser}. Recent progress has been made in reducing the computational costs of diffusion methods by adapting the number of diffusion steps~\cite{hu2023rf}  or leveraging DDIM~\cite{reuss2023goal}. 

Unlike the above methods, \method{} is the first to generalize diffusion-type policy learning to exploit source stochastic policies. As we will see, this leads to better performance compared to strong baselines (such as DDIM) when informative policies are available, especially when using a small number of diffusion steps. Since \method{} adapts existing distributions, it can be  seen as a few-shot learner, specifically for imitation learning~\cite{duan2017one, wang2020generalizing}. Existing work on few-shot imitation learning requires either specific prior policies~\cite{finn2017one, yu2018one, li2021meta} or a hierarchical structure~\cite{xu2023xskill}. In contrast, \method{} only requires that we are able to sample from the source policy. For fair comparison against few-shot methods, we adopt residual learning which is used for few-shot reinforcement learning \cite{silver2018residual, alakuijala2021residual, johannink2019residual}, planning \cite{mandlekar2023human}, and shared autonomy \cite{schaff2020residual, yoneda2023noise}. \method{} is also related to recent gradient-flow methods~\cite{ansari2020refining,taunyazov2023refining} but the SDE is over finite time and does not use classifiers to approximate the drift. 
\section{Bridging Policies: Theoretical Considerations}
\label{sec:theory}

The central premise of this work is the use of source policies for diffusion-based imitation learning: we posit that starting with a more informative source density facilitates the shaping of the target density. We examine this hypothesis theoretically in this section. Under reasonable assumptions, we show that a ``good'' source policy can enhance the resultant the target policy up to an addictive factor. Note that these results apply to any diffusion-type model whereby a source distribution is gradually adapted over time to match a target.  

Formally, we denote the ``difference'' between the action distribution $\hat{\pi}$ at time $t$ (conditioned upon observation $x$) and the expert policy $\pi_1$ as
\begin{equation*}
    \phi_{F,\hat{\pi}}(t,x)=F({\hat{\pi}_t(\cdot|x)},{\pi_1(\cdot|x)}).
\end{equation*}
where $F(\cdot,\cdot)$ is a measure of difference between two distributions (e.g. KL divergence $\KL(\cdot,\cdot)$ and cross-entropy $\H(\cdot,\cdot)$). In our setup, we diffuse from a source  distribution $\hat{\pi}$ from time $t=0$ to $t=1$, and would like $\phi_{F,\hat{\pi}}(1,x)$ to be small.

\begin{assumption}
    \label{ass:contbounded}
    There exist constants $\epsmax>\epsmin>0$ such that for all $t,x$,
    \begin{equation*}
        0\geq-\epsmin\geq \partial_t\phi_{F,\hat{\pi}}(t,x)\geq-\epsmax.
    \end{equation*}
\end{assumption}
We can interpret $\epsmin\,\d t$ and $\epsmax\,\d t$ as the minimum and maximum improvement in the differences towards $\pi_1$ after diffusing for some infinitesimal $\d t$ time. We believe is this reasonable given a trained model with limited capacity. 

\begin{restatable}{thm}{contimprovethm}
    \label{thm:contimprovethm}
    Let $\hat{\pi}_0$ and $\hat{\rho}_0$ be two source distributions and given that Assumption~\ref{ass:contbounded} holds. Then the improvement of the generated target distribution is bounded by the improvement of the source distribution
    \begin{align*}
        \,\phi_{F,\hat{\pi}}(1,x)-\phi_{F,\hat{\rho}}&(1,x)\\
        \leq&\,\phi_{F,\hat{\pi}}(0,x)-\phi_{F,\hat{\rho}}(0,x)+\epsmax-\epsmin.
    \end{align*}
\end{restatable}

The proof can be found in the Appendix \ref{app:theory}. 
Intuitively, Theorem~\ref{thm:contimprovethm} states that if $\hat{\pi}_0$ is a better source distribution than $\hat{\rho}_0$ (i.e.,  $\phi_{F,\hat{\pi}}(0,x) < \phi_{F,\hat{\rho}}(0,x)$), then after diffusion, $\hat{\pi}_1$ is  better than $\hat{\rho}_1$ up to an additive factor of $\epsmax-\epsmin$. To elaborate, let us rewrite the bound as
\begin{align*}
\phi_{F,\hat{\pi}}(1,x) + d(\hat{\pi}_0, \hat{\rho}_0) - (\epsmax-\epsmin) \leq \phi_{F,\hat{\rho}}(1,x) 
\end{align*}
where $d(\hat{\pi}_0, \hat{\rho}_0) = \phi_{F,\hat{\rho}}(0,x)-\phi_{F,\hat{\pi}}(0,x)$ is the difference in F between $\hat{\pi}$ and $\hat{\rho}$ at time $0$. If $\phi_{F,\hat{\pi}}(0,x) < \phi_{F,\hat{\rho}}(0,x)$, then 
$d(\hat{\pi}_0, \hat{\rho}_0) > 0$. The positive factor $\epsmax-\epsmin$ accounts for the variability in  improvements during the diffusion process; a greater disparity in the changes when starting from $\hat{\pi}_0$ versus $\hat{\rho}_0$ can influence the quality of the resulting target distributions. If we further assume that the improvements are equal regardless of the initial source, then this factor disappears and we obtain 
\begin{align*}
\phi_{F,\hat{\pi}}(1,x) + d(\hat{\pi}_0, \hat{\rho}_0) \leq \phi_{F,\hat{\rho}}(1,x).
\end{align*}

In practice, it is necessary to discretize time steps for sampling. Next, we extend our theoretical results to discrete time. Suppose we split the domain of time $[0,1]$ into $K+1$ discrete time steps $0=t_0<t_1<\dots<t_{K-1}<t_K=1$. We make a similar assumption, 
\begin{assumption}
    \label{ass:distbounded}
    There exist constants $\epsmax>\epsmin>0$ such that for all $t,x$,
    \begin{equation*}
        0\geq-\epsmin\delta t_k\geq \phi_{F,\hat{\pi}}(t_k,x)-\phi_{F,\hat{\pi}}(t_{k-1},x)\geq-\epsmax\delta t_k
    \end{equation*}
    where $k\in\{1,\dots,K\}$ and $\delta t_k=t_{k+1}-t_k$.
\end{assumption}
Here, $\epsmin\,\delta t_k$ and $\epsmax\,\delta t_k$ quantify the minimum and maximum improvement in $F$ towards $\pi_1$ after diffusing with step size $\delta t_k$. 

\begin{restatable}{thm}{distimprovethm}
    Let $\hat{\pi}_0$ and $\hat{\rho}_0$ be two source distributions and given that Assumption~\ref{ass:distbounded} holds. Then the improvement of the generated target distribution is bounded by the improvement of the source distribution
    \begin{align*}
        \,\phi_{F,\hat{\pi}}(1,x)-\phi_{F,\hat{\rho}}&(1,x)\\
        \leq&\,\phi_{F,\hat{\pi}}(0,x)-\phi_{F,\hat{\rho}}(0,x)+\epsmax-\epsmin.
    \end{align*}
\end{restatable}

Finally, we establish a similar result for the expected cost $\mathbb{E}[c(a|x)]$ when $F$ is the cross-entropy and the expert is Boltzmann rational. 
\begin{assumption}
    \label{ass:costdist}
    The density of the expert policy
    \begin{equation*}
        \pi_1(a|x)=\frac{1}{Z}\exp(-c(a|x))
    \end{equation*}
    where $Z$ is a normalizing constant.
\end{assumption}
\begin{restatable}{thm}{distcostimprovethm}
 \label{thm:distcostimprovethm}
    Let $\hat{\pi}_0$ and $\hat{\rho}_0$ be two source distributions and given that both Assumption~\ref{ass:distbounded} and \ref{ass:costdist} holds for $F=\H$ (cross-entropy). Then for any observation $x$, we have
    \begin{align*}
        \,\E_{a\sim\hat{\pi}_1}&[c(a|x)]-\E_{a\sim\hat{\rho}_1}[c(a|x)]\\
        \leq&\,\E_{a\sim\hat{\pi}_0}[c(a|x)]-\E_{a\sim\hat{\rho}_0}[c(a|x)]+\epsmax-\epsmin.
    \end{align*}
\end{restatable}
Similar to Theorem~\ref{thm:contimprovethm}, if we let $d_c(\hat{\pi}_0,\hat{\rho}_0) = \E_{a\sim\hat{\rho}_0}[c(a|x)] - \E_{a\sim\hat{\pi}_0}[c(a|x)])$ and $\hat{\pi}_0$ is a lower-cost source policy compared to $\hat{\rho}_0$, i.e., $d_c(\hat{\pi}_0,\hat{\rho}_0) > 0$, then under equal improvement $(\epsmax-\epsmin)=0$,
\begin{align*}
    \E_{\hat{\pi}_1}[c(a|x)] + d_c(\hat{\pi}_0,\hat{\rho}_0)  <  \E_{\hat{\rho}_1}[c(a|x)].
\end{align*} 
In words, the policy derived from the better source policy achieves a lower expected cost. 
In the next section, we discuss how we can practically bridge distributions for imitation learning.

\section{Method: \method{} for Imitation Learning}
\label{sec:method}
In this section, we present a method that can learn to adapt a source policy to match observed demonstrations. The source policy could be a simple Gaussian, or an action distribution hand-crafted using prior knowledge, or a data-driven policy learned from data. We call our method \textbf{B}ehavioral \textbf{R}efinement via \textbf{I}nterpolant-based \textbf{D}iffusion for \textbf{Ge}nerative \textbf{R}obotics (\method{}). 

\method{} is based on Stochastic Interpolants~\cite{albergo2023stochastic}, a recently-proposed framework to bridge densities in finite time. At a high-level, a stochastic interpolant is a continuous-time stochastic process $\{y_t\}_t$ that ``interpolates'' between two arbitrary densities. As a concrete example, consider two densities $p_0$ and $p_1$. A simple linear stochastic interpolant is
\begin{align}
y_t = (1-t)y_0 + ty_1 + \sqrt{2t(1-t)}z
\end{align}
where $y_0$ and $y_1$ are drawn from $p_0$ and $p_1$, respectively, and $z$ is drawn from a standard Gaussian. By construction, the paths of $y_t$ bridge samples from $p_0$ at time $t=0$ and from $p_1$ at $t=1$. Our goal is to learn how to transport samples along the paths of this interpolant. 

In the following, we will present stochastic interpolants more formally and show how they can be adapted to imitation learning. We will then detail the key design elements explored in this work. 

\subsection{Stochastic Interpolants for Imitation Learning}
Let $(C^r(\mathbb{R}^n))^m$ be the space of $r$ continuously differentiable functions from $\mathbb{R}^m$ to $\mathbb{R}^n$ and $(C_0^r(\mathbb{R}^n))^m$ as the space of compactly supported and $r$ continuously differentiable functions from $\mathbb{R}^m$ to $\mathbb{R}^n$. We extend the definition of stochastic interpolant in \cite{albergo2023stochastic} to condition on observation $x$ (e.g., an image or joint angles),  similar to the concurrent work \cite{huang2023conditional}.
\begin{definition}[Stochastic Interpolant]
\label{def:si}
Denote $n_x$ and $n_a$ as the dimension of observation and action respectively. Let $x\in\mathbb{R}^{n_x}$ be an observation. Given two probability density functions $\pi_{0}(a|x)$ and $\pi_{1}(a|x)$, a stochastic interpolant between $\pi_{0}(a|x)$ and $\pi_{1}(a|x)$ is a stochastic process $\{a_t\}_{t\in[0,1]}$ satisfying
\begin{align}
\label{eq:si}
    a_t = I(t, a_0, a_1, x) + \gamma(t)z,\quad t \in \left[0, 1\right]
\end{align}
where
\begin{itemize}
    \item[1.] $I\in C^2([0,1]\times\mathbb{R}^{n_a}\times\mathbb{R}^{n_a}\times\mathbb{R}^{n_x})^{n_a}$ satisfies the boundary conditions $I(0, a_0, a_1, x)=a_0$ and $I(1, a_0, a_1, x)=a_1$
    \item[2.] There exist some $C_1<\infty$ such that $|\partial_t I(t, a_0, a_1, x)| \leq C_1|a_0 - a_1|$ for $t\in[0,1]$.
    \item[3.] $\gamma(t)$ satisfies $\gamma(0)=\gamma(1)=0$, $\gamma(t)>0$ for all $t\in(0,1)$, and $\gamma^2\in C^2([0,1])$.
    \item[4.] The pair $(a_0, a_1)$ is drawn from a probability measure $v$ that marginalizes on $\pi_0(\cdot|x)$ and $\pi_1(\cdot|x)$.
    \item[5.] $z$ is a standard Gaussian random variable independent of $(a_0, a_1)$.
\end{itemize}

\end{definition}

From Definition \ref{def:si} and Theorem 2.6 of \cite{albergo2023stochastic}, the transport equation between the source and target distribution is
\begin{align}
\label{eq:te}
    \partial_t \pi + \nabla_a\cdot(b\pi)=0
\end{align}
where $\pi(t,a,x)\coloneqq\pi_t(a|x)$ and the velocity $b$ is defined as
\begin{align}
\label{eq:b}
    b(t,a, x) = \mathbb{E}\left[\partial_t I(t, a_0, a_1, x) + \dot{\gamma}(t)z\right].
\end{align}

We can rewrite the transport equation (Eq. \ref{eq:te}) as a forward Fokker-Planck equation, along with the corresponding forward stochastic process~\cite{albergo2023stochastic}. For any $\epsilon \in C_0([0,1])$ with $\epsilon(t) \geq 0$ for all $t\in[0,1]$, the probability density $\pi$ in equation \ref{eq:si} satisfies the forward Fokker-Planck equation
\begin{align}
\label{eq:fpe}
    \partial_t \pi + \nabla_a \cdot(b_F\pi) = \epsilon\nabla^2_a\pi,\quad\pi(0)=\pi_0
\end{align}
where 
\begin{align}
    b_F(t, a, x) \coloneqq b(t, a, x) + \epsilon(t)s(t, a, x)
\end{align}
and $s(t, a, x)\coloneqq \nabla_a \log \pi(t,a,x)$ is the score of $\pi_t(\cdot|x)$. The solutions of the forward stochastic differentiable equation (SDE) associated with the Fokker-Planck (Eqn. \ref{eq:fpe}) satisfy  
\begin{align}
\label{eq:fsde}
    \d a_t=b_F(t, a_t, x)\,\d t + \sqrt{2\epsilon(t)}\,\d W_t
\end{align}
solved forward in time from the initial action $a_0\sim\pi_0$.

\para{Action Sampling.} 
Using above properties, we can sample actions by transporting samples from the source distribution $\pi_0$ to the target distribution $\pi_1$, i.e., we solve the forward SDE in Eq.~\ref{eq:fsde} (See Fig. \ref{fig:method}). The time interval $t \in [0,1]$ is discretized into points ${t_0, \ldots, t_K}$, with $K$ denoting the total number of diffusion steps and $\delta t$ representing the uniform time step increment between $t_{k}$ and $t_{k+1}$. The specifics of the sampling process are outlined in Algorithm~\ref{alg:ipi}.

\begin{algorithm}
\caption{\method{} Sampling}\label{alg:ipi}
\begin{algorithmic}
\Require Current observation $x$ and a sample from the source policy $a_0\sim\pi_0(a_0|x)$
\For{\texttt{$k \gets 1$ to $K$}}
\State $z\sim\mathcal{N}(0, I)$
\State $b_F(t_k, a_{t_k}, x)= b(t_k, x) + \epsilon(t_k)s(t_k, a_{t_k}, x)$
\State $a_{t_{k+1}}=a_{t_k} + b_F(t, a_{t_k}, x)\delta t + \sqrt{2\epsilon(t_k)}z$
\EndFor
\Ensure $a_1$
\end{algorithmic}
\end{algorithm}

\para{Model Training. }
Sampling actions as above requires the velocity $b$ and score $s$. Suppose that these two functions are parameterized by function approximators (e.g., neural networks), denoted as $b_\theta$ and $s_\eta$. The velocity $b$ defined in equation \ref{eq:b} can be trained by minimizing, 
\begin{align}
\label{loss:b}
L_b(\theta) = \int_0^1\mathbb{E}\left[b_\theta(t, a_t, x)-(\partial_t I(t, a_0, a_1, x) + \dot{\gamma}(t)z)\right]^2dt
\end{align}
where $a_t$ is defined in Eqn. (\ref{eq:si}) and the expectation is taken independently over $\pi_0(a_0|x)$, $\pi_1(a_1|x)$, and $\mathcal{N}(z|0, I)$. Similarly, the score $s$ of the probability density $\pi(t)$ can be trained by minimizing 
\begin{align}
L_s(\eta)=\int_0^1\mathbb{E}\left[s_\eta(t, a_t, x)+\gamma^{-1}(t)z\right]^2dt.
\end{align}

In practice, predicting $\gamma^{-1}(t)z$ can be difficult since $\gamma^{-1}(t)$ approaches infinity as $t$ approaches $0$ or $1$. To address this issue, we re-parameterize the score as $s_\eta=\hat{s}_\eta\gamma^{-1}(t)$ in our model. To further improve training stability, we decompose the velocity $b$ as suggested in ~\cite{albergo2023stochastic},
\begin{align}
\label{eq:v}
b(t, a, x) = v(t, a, x) - \dot{\gamma}(t)\gamma(t)s(t, a, x)
\end{align}
where $v$ can be trained by minimizing the quadratic objective 
\begin{align}
\label{loss:v}
L_v(\phi) = \int_0^1\mathbb{E}\left(v_\phi(t, a_t, x)-\partial_t I(t, a_0, a_1, x))\right)^2dt
\end{align}
Given samples in the dataset $D$ comprising tuples $(x, a_1)$, the above objectives can be approximated via Monte-Carlo sampling. Our  training algorithm is outlined in Algorithm \ref{alg:ipt}. 

\begin{algorithm}
\caption{\method{} Training}\label{alg:ipt}
\begin{algorithmic}
\Require $D$ and batch size $N$
\While{\texttt{Not Converged}}
\State $(x, a_1)\sim D$ and $a_0\sim\pi_0(a|x)$ \Comment{Sample data}
\State $t\sim \mathrm{U}(0, 1)$ \Comment{Uniformly sample $t$}
\State Sample $a_t\sim I(t, a_0, a_1, x) + \gamma(t)z$
\State Compute losses: 
\State $L_b(\theta)=\frac{1}{N}\sum\left(b_\theta(t, a_t, x)-(\partial_t I + \dot{\gamma}z)\right)^2$
\State $L_s(\eta)=\frac{1}{N}\sum\left(s_\eta(t, a_t, x)+\gamma^{-1}z\right)^2$
\State $L_v(\phi)=\frac{1}{N}\sum\left(v_\phi(t, a_t, x)-\partial_t I\right)^2$
\State Gradient descent on $\theta$, $\eta$ and $\phi$ 
\EndWhile
\Ensure $b_\theta$, $s_\eta$ and $v_\phi$
\end{algorithmic}
\end{algorithm}
\subsection{Design Decisions}
To apply \method{} in practice, we have to design several key components, specifically the source distribution $\pi_0$, the interpolant $I(t, a_0, a_1, x)$, and the noise schedule $\gamma(t)$ and the   $\epsilon(t)$. In this work, we will focus on comparing specific source distributions and interpolants. 


\begin{figure}
    \centering
\includegraphics[width=0.95\linewidth]{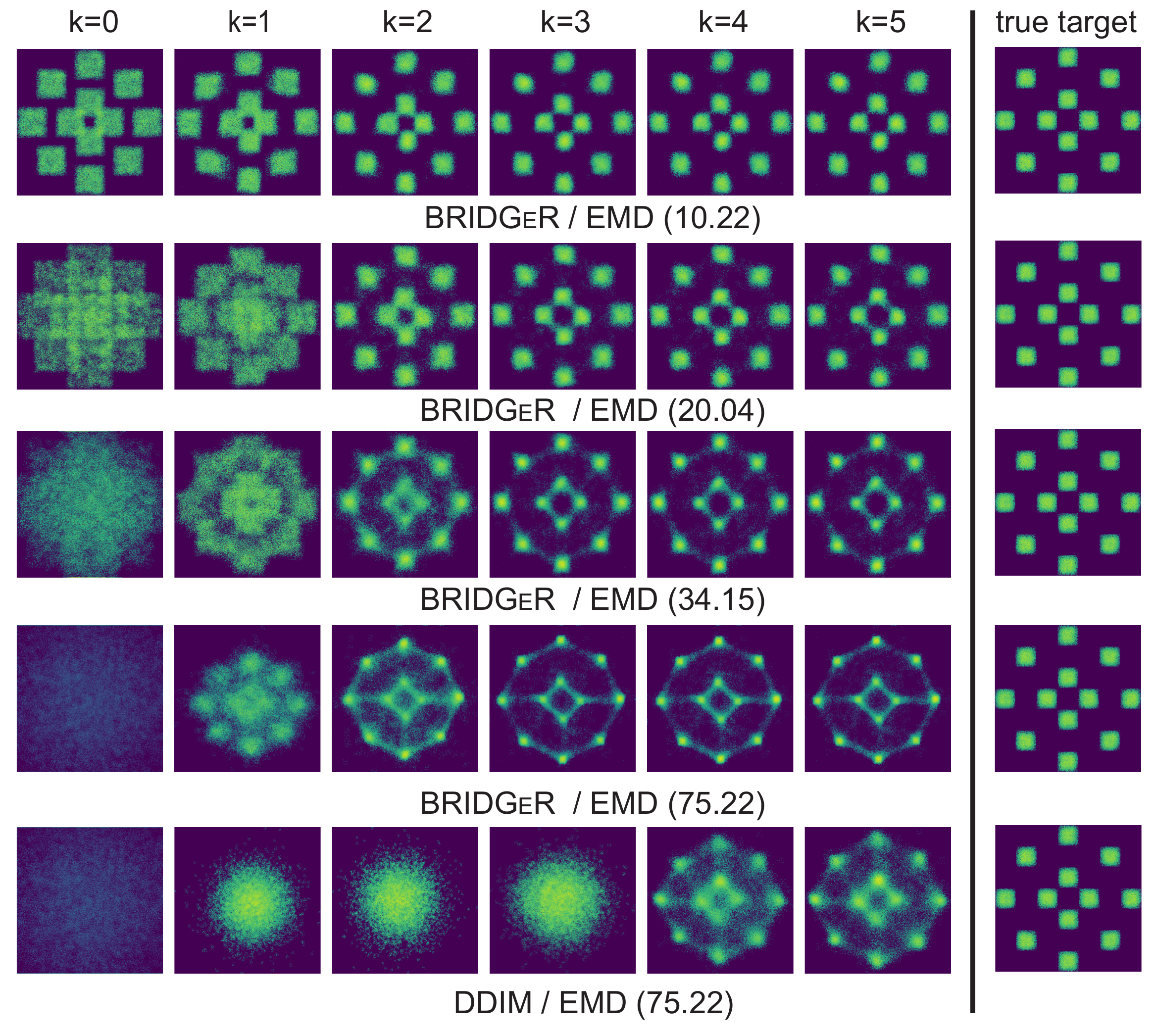}
    \caption{\small Intermediate distributions obtained from \method{} and DDIM trained on 2D synthetic data. With a source distribution that is closer to the target distribution (smaller Earth Mover's Distance (EMD) values), \method{} can better recover the true target distribution. 
    } 
    \label{fig:toy_source}
\end{figure}

\begin{figure}
    \centering    \includegraphics[width=0.95\linewidth]{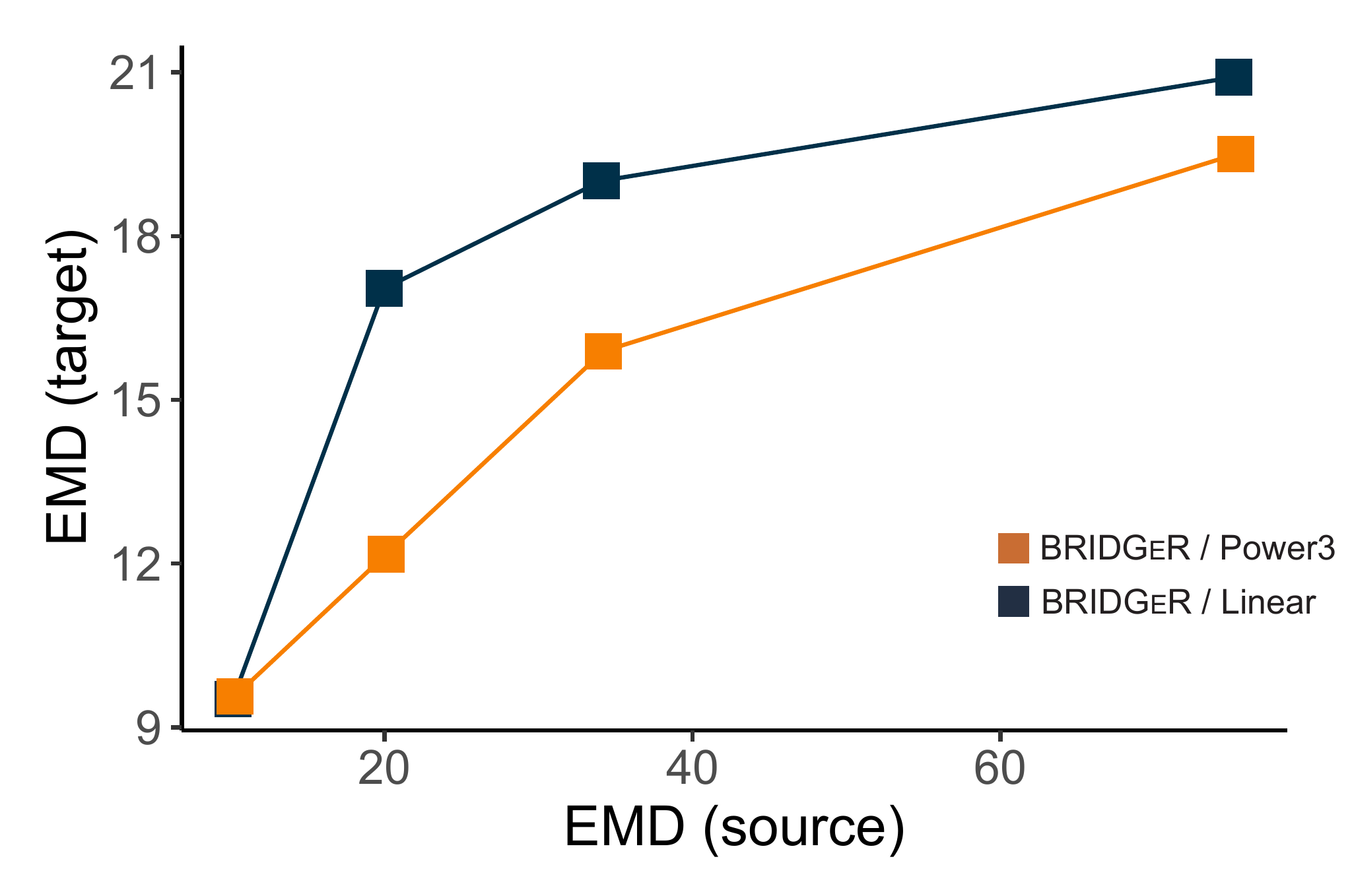}
    \caption{\small Earth Mover's Distance (EMD) of the generated target distributions under different source distributions and interpolant functions on our 2D synthetic dataset. Each point represents the EMD between a source/target distribution and the true target distribution. } 
    \label{fig:toy_source_emd}
\end{figure}

\para{Source Distributions. } As shown by Theorem \ref{thm:contimprovethm} in Sec. \ref{sec:theory}, a source distribution that closer to the target can yield better policies. Our preliminary experiments with 2D synthetic samples supports this notion (Fig. \ref{fig:toy_source} and  \ref{fig:toy_source_emd}) --- we see that source distributions that are closer to the target (lower EMD) are more similar to the true target distribution. Along with standard Gaussians, our experiments will involve two kinds of source policies:
\begin{itemize}
\item \emph{Heuristic policies}: hand-crafted policies (e.g., using rules) based on prior knowledge. Our heuristic policies are task-dependent and detailed in Appendix~\ref{app:sourcepolicies}.
\item \emph{Data-driven policies}: policies learned from a dataset. In our experiments, we use lightweight Conditional Variational Autoencoders (CVAEs)~\cite{sohn2015learning} as a representative policy. CVAEs tend to not fully capture complex target distributions, but are cheap to sample from.
\end{itemize}
These policy types are commonly employed in robotics and by comparing them, we aim to evaluate potential benefits that \method{} may offer under different use-cases.

\begin{figure}
    \centering
\includegraphics[width=0.95\linewidth]{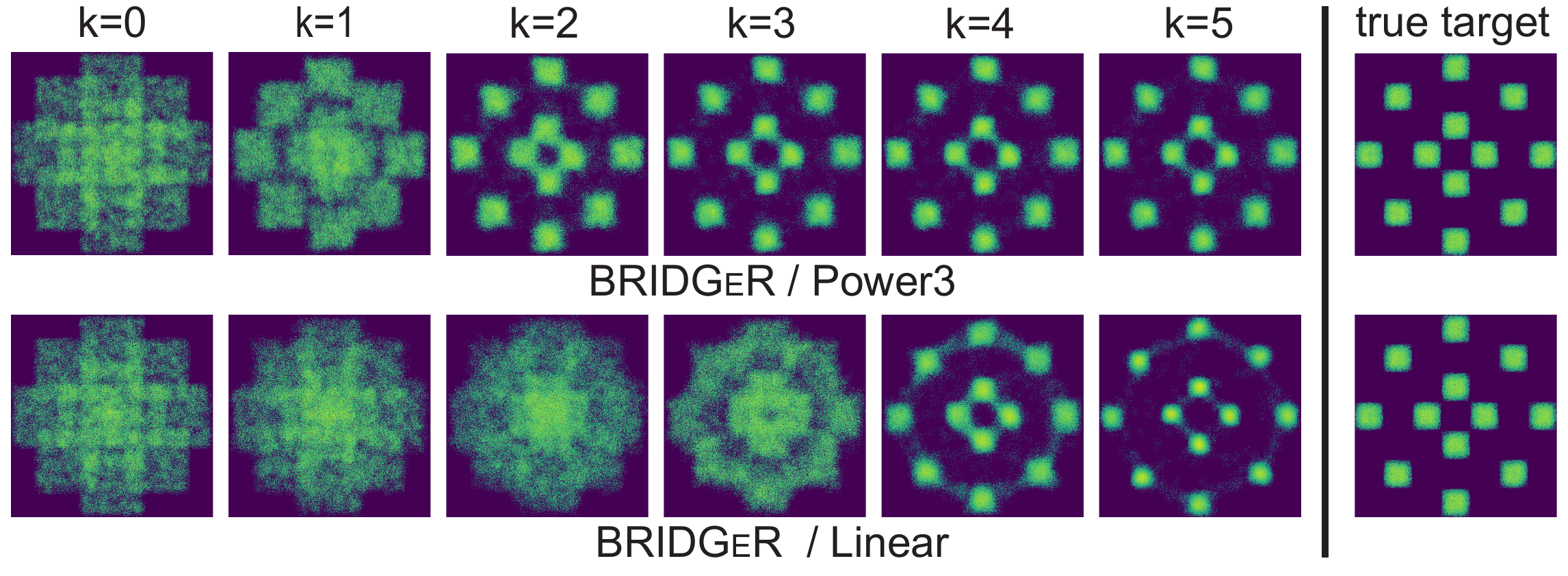}
    \caption{\small Intermediate distributions under different interpolant functions (trained on 2D synthetic data).} 
    \label{fig:toy_interpolant}
    \vspace{-1em}
\end{figure}

\para{Interpolant Function.} The second major component is the interpolant function $I(t, a_0, a_1, x)$. We will use spatially linear interpolants~\cite{albergo2023stochastic}, $a_t = \alpha(t)a_0 + \beta(t)a_1+\gamma(t)z$, specifically, 
\begin{itemize}
\item \emph{Linear Interpolant} where $\alpha(t)=1-t$ and $\beta(t)=t$.
\item \emph{Power3 Interpolant} where $\alpha(t)=(1-t)^m$ and $\beta(t)=1-(1-t)^m$. In our experiments, we set $m=3$, which worked well in preliminary tests. 
\end{itemize}
The linear interpolant uniformly progresses samples from the source to the target distribution, which we observed makes training more stable. Conversely, the Power3 interpolant starts with larger steps that decelerate towards the end. This introduces the target pattern sooner than the linear approach, as shown in Fig. \ref{fig:toy_interpolant}. Qualitatively, we find Power3 worked better in scenarios with highly multi-modal demonstrations. 

\para{Noise Schedule and Diffusion Coefficient. }
\begin{figure}
    \centering
\includegraphics[width=0.95\linewidth]{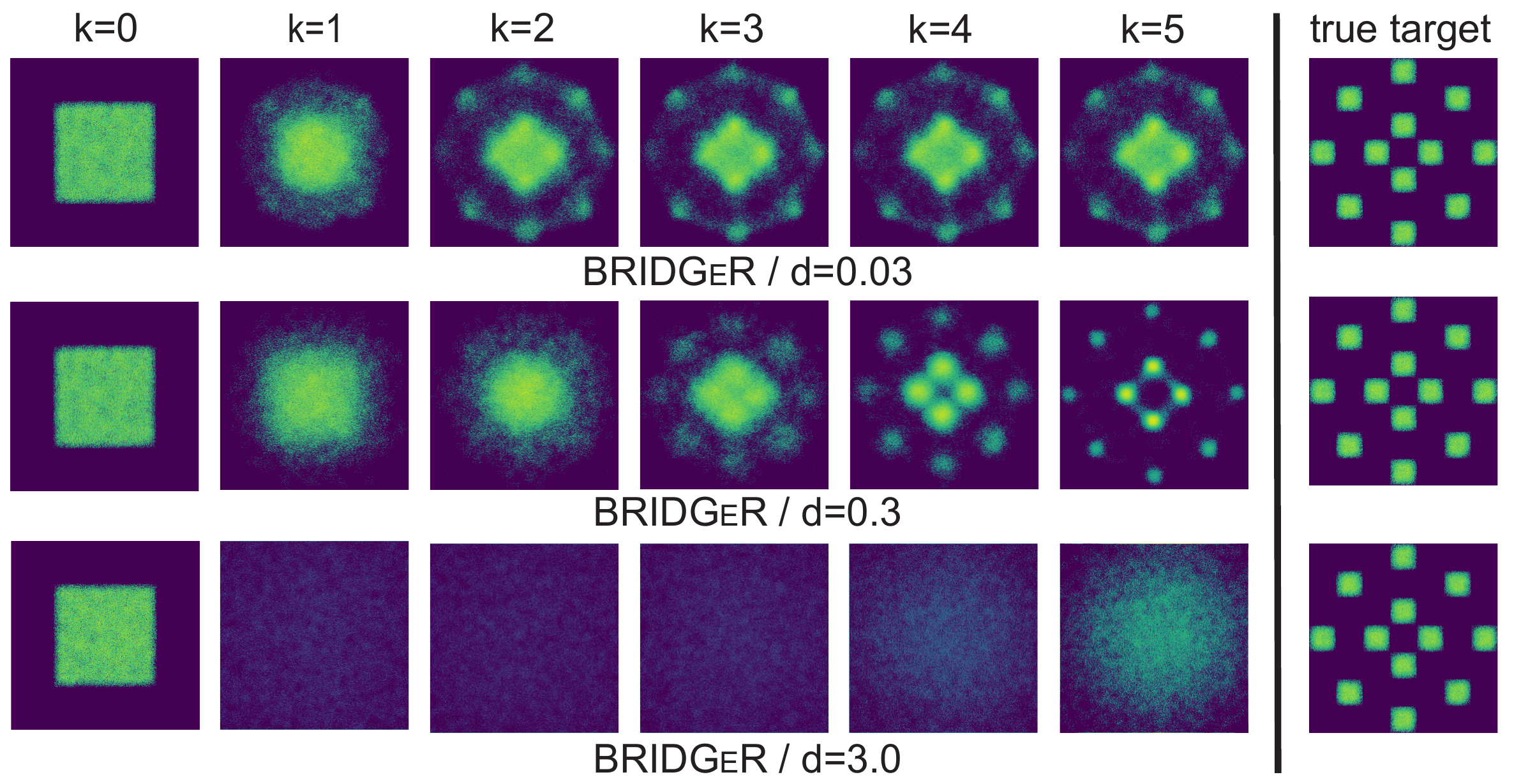}
    \caption{\small Intermediate distributions with varying $\gamma(t)=d\sqrt{2t(1-t)}$. When the support of the source distribution is narrower than the target distribution, selecting a small $\gamma$ value ($d=0.03$) results in samples clustering within the high-density areas of the source. Conversely, an excessively large $\gamma$ ($d=3$) results in overdispersion. However, a well-choosen $\gamma$ ($d=0.3$) facilitates coverage to ensure reasonable recovery of the target.} 
    \label{fig:toy_gamma}
\end{figure}
The noise schedule governs the variance of Gaussian latent noise across the bridge. In our experiments, we set $\gamma(t)=d\sqrt{2t(1-t)}$, with $d$ acting as a scalar to adjust $\gamma$'s magnitude. This configuration results in minimal Gaussian noise at the onset of the transition from the source distribution, increasing to a peak variance before tapering off to zero as samples approach the target distribution. Selecting a larger $d$ facilitates exploration of low-density areas but risks introducing excessive noise, as illustrated in Figure \ref{fig:toy_gamma}. The diffusion coefficient $\epsilon(t)$  controls the level of noise in the forward SDE (Eq. \ref{eq:fsde}). In our experiments, we define $\epsilon$ as $\epsilon(t)=c(1-t)$, where $c$ is a scalar to adjust its magnitude. We set two choices for $c$ (1 and 3) and for $d$ (0.03 and 0.3) and reported the best results. In general, our results were relatively robust to these choices.   

\section{Experiments}
\label{sec:sim_exp}
This section describes experiments designed to evaluate the performance of \method{} relative to recent methods, particularly diffusion-based imitation learning. More importantly, we aimed to test our hypothesis that leveraging better source distributions within a diffusion framework leads to better policies. We further hypothesized that \method{} performs better than DDIM policies given a small number of diffusion steps, and limited data. Finally, we sought to examine the effect of decision decisions, particularly the interpolant choice. We first give an overview of our experimental setup (details relegated to Appendix) followed by our main results.

\subsection{Domains}
\begin{figure*}
    \centering
    \includegraphics[width=1.0\textwidth]{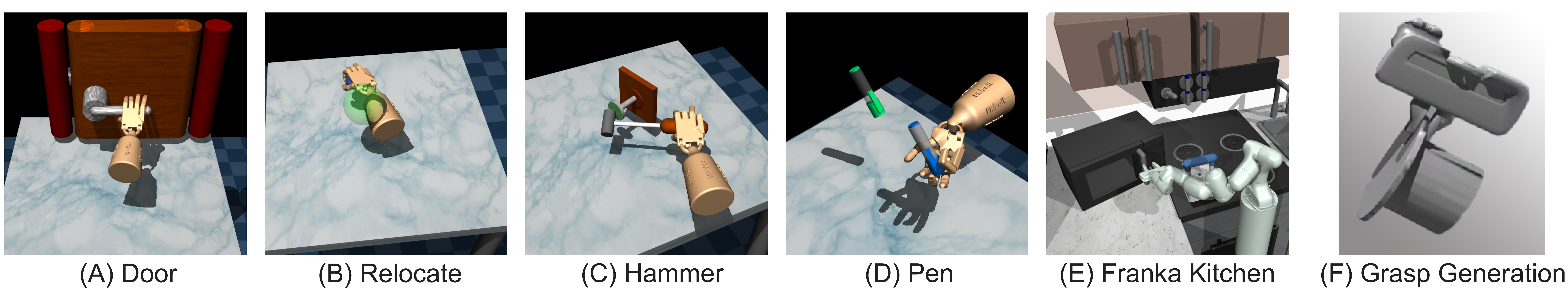}
    \caption{\small Experimental Domains. \textbf{(A)-(D)} Adroit tasks that involve the control of a 24-degree-of-freedom robot hand to accomplish four specific tasks: \textbf{(A)} Door: opening a door, \textbf{(B)} Relocate: moving a ball to a target position, \textbf{(C)} Hammer: driving a nail into a board, and \textbf{(D)} Pen: aligning a pen with a target orientation \cite{fu2020d4rl}. \textbf{(E)} Franka Kitchen includes 7 objects available for interaction and the aim is to accomplish 4 subtasks: opening the microwave, relocating the kettle, flipping the light switch, and sliding open the cabinet door, with arbitrary order. \textbf{(F)} The goal of 6-DoF Grasp-Pose generation is to generate grasp poses capable of successfully picking up an object.} 
    \label{fig:exp_domain}
\end{figure*}

To evaluate our hypotheses above, we chose six challenging robot benchmarks in three domains~\cite{fu2020d4rl, rajeswaran2018learning, gupta2020relay, urain2023se} (See Fig. \ref{fig:exp_domain}). The tasks in these domains feature multi-modal demonstrations, which comprise multiple stages with high-dimensional and high-precision actions. 

\para{Franka Kitchen (State Observations).} The goal in Franka Kitchen is to control a 7 DoF robot arm to solve seven subtasks by reaching a desired state configuration (Fig. \ref{fig:exp_domain}). Franka Kitchen comprises three datasets and we chose the mixed dataset, which is considered as the most challenging; it presents various subtasks being executed with 4 target subtasks that are not completed in sequence~\cite{fu2020d4rl}. The model is trained on three varying data sizes: small (16k sequences), medium (32k sequences), large (64k sequences). 

\para{Adroit (State Observations).} The Adroit tasks require control of a 24-DoF robot hand to accomplish four tasks and is considered one of the most challenging task sets; it requires a model to generate intricate, high-dimensional, and high-precision action sequences over long time horizons. We use the dataset of human demonstrations provided in the DAPG repository~\cite{fu2020d4rl}. Similar to Franka Kitchen, we train the model on three different data sizes; small (1.25k
sequences), medium (2.5k sequences), and large (5k sequences). 

\para{6-DoF Grasp Pose Generation (Point-Cloud Observations).} The goal here is to generate grasp poses capable of picking up an object given  the object's point cloud. The target grasp-pose distribution is multi-modal~\cite{urain2023se}, with high-dimensional point clouds as conditioning observations. \method{} is trained on Acronym dataset~\cite{vahrenkamp2010integrated} (552 objects with 200-2000 grasps per object) after a LogMap transformation~\cite{urain2023se} of the grasp poses. 

\subsection{Compared Methods.}
\method{} was implemented in PyTorch and trained using the Algorithm \ref{alg:ipt}. Additional details regarding hyperparameters and network structure are given in Appendix \ref{app:baselines}.
To evaluate performance, we compared \method{} against strong baselines, specifically: 

\begin{itemize}
    \item \emph{DDIM}: Diffusion policy~\cite{chi2023diffusion, reuss2023goal}, a recent diffusion-based imitation-learning method which trains DDPM and applies DDIM~\cite{song2020denoising} during test-time.
    \item \emph{Residual Policy}: This baseline applies residual learning towards policy learning~\cite{silver2018residual, alakuijala2021residual, johannink2019residual} and is a representative method not based on diffusion.
    \item \emph{SE3}: A state-of-the-art score-based diffusion model designed in SE(3)~ \cite{urain2023se} specifically designed for generating 6-Dof grasp poses. Note that this baseline only applies to the grasp generation domain. 
\end{itemize}

We also evaluated variants of \method{} with different Interpolant functions (Linear and Power3), along with different source policies. As stated in Sec. \ref{sec:method}, we investigate three different kinds of source policies: 
\begin{itemize}
\item \emph{Gaussian}: a standard Gaussian, similar to DDPM. 
\item \emph{CVAE}: a data-driven policy. For fair comparison to the baselines, the CVAE source policy was trained using the same dataset. Qualitatively, we find the CVAE was generally not able to capture the diversity of demonstrations. 
\item \emph{Heuristic}: hand-crafted policies for the Hammer, Door, and Grasp tasks (described in Appendix \ref{app:sourcepolicies}). Heuristic policies were not available for the other tasks due to their complexity and the high-dimension of the action space. 
\end{itemize}
We label each \method{} variant as \method{} / $[\cdots]$ / $[\cdots]$. For example, \method{} / CVAE / Linear represents \method{} with a CVAE source policy and the linear interpolant.

\subsection{Test Methodology.}
Following prior work, we compute success measures for the different tasks. We report success rate (Adroit) and number of successful tasks (Franka Kitchen), averaged over three different seeds. Under each seed, the models are tested with 100 random initializations for each configuration (dataset sizes [small, medium, large] $\times$ the number of diffusion steps). 

For the Grasp task, we evaluated the models in Nvidia Isaac Gym~\cite{makoviychuk2021isaac} and we report the rate of successful grasps; the robot was able to pick up the object without dropping it \cite{urain2023se}. Each object was evaluated using 100 grasps with random initialization of the object pose. We tested our models on both on unseen-objects belonging to the object categories seen in the dataset and unseen-objects from unseen categories. In addition to success rates, we also measured the Earth Mover's Distance (EMD) between the generated grasps and the training data distribution. 

\begin{table*}
	\centering
 	\caption{Average task performance on Adroit (success rate) and Franka Kitchen (number of successful sub-tasks). Best scores in \textbf{bold}. We compare \method{} against state-of-the-art methods under a different number of diffusion steps when trained with the Large dataset. \method{} with $k=0$ indicates the source policy. \method{} generally outperforms the competing methods. Results are similar for the small and medium datasets with complete results in the Appendix.}
        \label{table:d4rl_franka}
	\begin{tabular}{cc | ccc | cc | c}
		\toprule
		& &  \multicolumn{3}{c|}{\method{}}						  & \multicolumn{2}{c|}{Residual Policy}  & \multirow{2}{*}{DDIM} 
		\\
		&  				&  CVAE    & Heuristic   & Gaussian    & CVAE           & Heuristic  &    
		\\    
		\midrule

		\multirow{3}{*}{Door}  
  & $k=0$  & $0.21 \pm 0.04$ & $\mathbf{0.22 \pm 0.06}$  & $0.00 \pm 0.00$ & $0.21 \pm 0.04$ & $\mathbf{0.22 \pm 0.06}$  & $0.00 \pm 0.00$                  
		\\ 
		& $k=5$  	& $\mathbf{0.60 \pm 0.15}$ & $0.00 \pm 0.00$  & $0.08 \pm 0.06$ & $0.04 \pm 0.06$ & $0.00 \pm 0.00$  & $0.02 \pm 0.02$                  
		\\
		& $k=20$ & $\mathbf{0.52 \pm 0.23}$ & $0.45 \pm 0.04$  & $0.10 \pm 0.07$ & $0.04 \pm 0.06$ & $0.00 \pm 0.00$ & $0.38 \pm 0.11$                     
		\\
		& $k=80$ & $0.50 \pm 0.14$ & $\mathbf{0.63 \pm 0.08}$  & $0.12 \pm 0.09$ & $0.04 \pm 0.06$ & $0.00 \pm 0.00$ & $0.05 \pm 0.02$                   
		\\ 
		\midrule

		\multirow{3}{*}{Relocate}    
  & $k=0$  & $\mathbf{0.31 \pm 0.15}$ & - & $0.00 \pm 0.00$ & $\mathbf{0.31 \pm 0.15}$ & - & $0.00 \pm 0.00$                  
		\\ 
		& $k=5$    & $\mathbf{0.75 \pm 0.11}$ & -	 & $0.61 \pm 0.05$ & $0.3 \pm 0.04$ & -	& $0.17 \pm 0.04$                      
		\\
		& $k=20$  & $0.7 \pm 0.11$ & -	 & $\mathbf{0.72 \pm 0.09}$ & $0.34 \pm 0.04$ & -	& $0.37 \pm 0.04$                       
		\\
		& $k=80$  & $0.79 \pm 0.08$ & -	 & $\mathbf{0.81 \pm 0.04}$ & $0.34 \pm 0.04$ & -	& $0.26 \pm 0.03$                      
		\\ 
		\midrule

  		\multirow{3}{*}{Hammer}     
    & $k=0$  & $\mathbf{0.16 \pm 0.03}$ & $0.11 \pm 0.08$  & $0.00 \pm 0.00$ & $\mathbf{0.16 \pm 0.03}$ & $0.11 \pm 0.08$  & $0.00 \pm 0.00$                  
		\\  
		& $k=5$   & $\mathbf{0.44 \pm 0.11}$ & $0.2 \pm 0.24$  & $0.16 \pm 0.04$ & $0.13 \pm 0.06$ & $0.24 \pm 0.07$ & $0.01 \pm 0.01$                      
		\\
		& $k=20$  & $0.63 \pm 0.08$ & $\mathbf{0.74 \pm 0.07}$  & $0.35 \pm 0.02$ & $0.13 \pm 0.06$ & $0.24 \pm 0.07$ & $0.17 \pm 0.12$                       
		\\
		& $k=80$  & $\mathbf{0.72 \pm 0.09}$ & $0.47 \pm 0.25$  & $0.43 \pm 0.09$ & $0.13 \pm 0.06$ & $0.24 \pm 0.07$ & $0.03 \pm 0.04$                      
		\\ 
		\midrule
		
		\multirow{3}{*}{Pen}   
  & $k=0$ & $\mathbf{0.29 \pm 0.05}$ & -  & $0.00 \pm 0.00$ & $\mathbf{0.29 \pm 0.05}$ & -  & $0.00 \pm 0.00$                  
		\\
		& $k=5$   & $0.45 \pm 0.04$ & - & $0.43 \pm 0.02$ & $0.33 \pm 0.12$ & - & $\mathbf{0.54 \pm 0.03}$                       
		\\
		& $k=20$  & $0.49 \pm 0.09$ & - & $0.53 \pm 0.02$ & $0.33 \pm 0.12$ & - & $\mathbf{0.51 \pm 0.03}$                       
		\\
		& $k=80$  & $\mathbf{0.55 \pm 0.06}$ & - & $\mathbf{0.55 \pm 0.03}$ & $0.33 \pm 0.12$ & -                  & $0.52 \pm 0.05$                       
		\\ 
		\midrule

		\multirow{3}{*}{Franka Kitchen} 
  & $k=0$    & $\mathbf{1.53 \pm 0.09}$ & - & $0.00 \pm 0.00$ & $\mathbf{1.53 \pm 0.09}$ & - & $0.00 \pm 0.00$                      
		\\ 
		& $k=5$    & $\mathbf{1.96 \pm 0.03}$ & - & $1.18 \pm 0.02$ & $1.55 \pm 0.10$ & - & $1.84 \pm 0.06$                      
		\\
		& $k=20$  & $\mathbf{2.09 \pm 0.04}$ & - & $1.54 \pm 0.03$ & $1.55 \pm 0.10$ & - & $1.93 \pm 0.07$
		\\
		& $k=80$  & $\mathbf{2.16 \pm 0.03}$ & - & $1.70 \pm 0.05$ & $1.55 \pm 0.10$ & - & $1.92 \pm 0.02$                      
		\\ 
		\bottomrule
		
	\end{tabular}

\end{table*}
\begin{table*}[ht]
\centering
\caption{Success rate (averaged over $100$ grasps on ten test objects). \method{} significantly outperforms DDIM and Residual Policy across the number of diffusion steps. Compared to SE3, \method{} achieve higher success rate when the number of diffusion steps is small. We show up to $k=160$ steps to be consistent with prior reported results~\cite{urain2023se}. Best scores in \textbf{bold}.}
\label{table:grasp}
\addtolength{\tabcolsep}{-2pt}
\begin{tabular}{cc | ccc | cc | cc}
\toprule
		& &  \multicolumn{3}{c|}{\method{}}						  & \multicolumn{2}{c|}{Residual Policy}  & \multirow{2}{*}{DDIM} & \multirow{2}{*}{SE3}
		\\
		&  				&  CVAE    & Heuristic   & Gaussian    & CVAE           & Heuristic  &    
		\\    
		\midrule

\multirow{3}{*}{Seen Categories}    
& $k=0$  	 	& $\mathbf{0.26 \pm 0.02}$ & $0.06 \pm 0.00$ & $0.00 \pm 0.00$ & $\mathbf{0.26 \pm 0.02}$  & $0.06 \pm 0.00$ & $0.00 \pm 0.00$ & $0.00 \pm 0.00$          
\\   
& $k=5$  	 	& $\mathbf{0.73 \pm 0.15}$ & $0.64 \pm 0.17$ & $0.56 \pm 0.17$& $0.09 \pm 0.15$  & $0.01 \pm 0.03$ & $0.52 \pm 0.24$ & $0.38 \pm 0.26$          
\\
& $k=20$  	& $\mathbf{0.93 \pm 0.08}$ & $0.91 \pm 0.07$& $0.83 \pm 0.28$ & $0.09 \pm 0.15$  & $0.01 \pm 0.03$ & $0.64 \pm 0.21$ & $0.78 \pm 0.19$ 							
\\
& $k=160$  	& $0.88 \pm 0.10$ & $\mathbf{0.91 \pm 0.08}$& $0.90 \pm 0.06$ & $0.09 \pm 0.15$ & $0.01 \pm 0.03$ & $0.64 \pm 0.26$ & $\mathbf{0.91 \pm 0.08}$ 							
\\
\midrule

\multirow{3}{*}{Unseen Categories}     
& $k=0$  	 	& $\mathbf{0.23 \pm 0.04}$ & $0.00 \pm 0.00$ & $0.00 \pm 0.00$& $\mathbf{0.23 \pm 0.04}$  & $0.00 \pm 0.00$ & $0.00 \pm 0.00$ & $0.00 \pm 0.00$          
\\  
& $k=5$  	& $\mathbf{0.48 \pm 0.12}$ & $0.43 \pm 0.19$ & $0.45 \pm 0.14$ & $0.10 \pm 0.16$ & $0.12 \pm 0.28$ & $0.33 \pm 0.20$ & $0.20 \pm 0.10$       
\\
& $k=20$  & $\mathbf{0.67 \pm 0.21}$ & $\mathbf{0.67 \pm 0.26}$ & $0.65 \pm 0.25$ & $0.10 \pm 0.16$ & $0.12 \pm 0.28$ & $0.41 \pm 0.23$ & $0.55 \pm 0.24$							
\\
& $k=160$ & $\mathbf{0.71 \pm 0.24}$ & $0.66 \pm 0.23$ & $0.64 \pm 0.23$ & $0.10 \pm 0.16$ & $0.12 \pm 0.28$ & $0.35 \pm 0.19$ & $0.66 \pm 0.25$							
\\

\bottomrule

\end{tabular}

\addtolength{\tabcolsep}{+2pt}

\end{table*}

\subsection{Main Results and Discussion}
\begin{figure}
    \centering
    \includegraphics[width=1.0\linewidth]{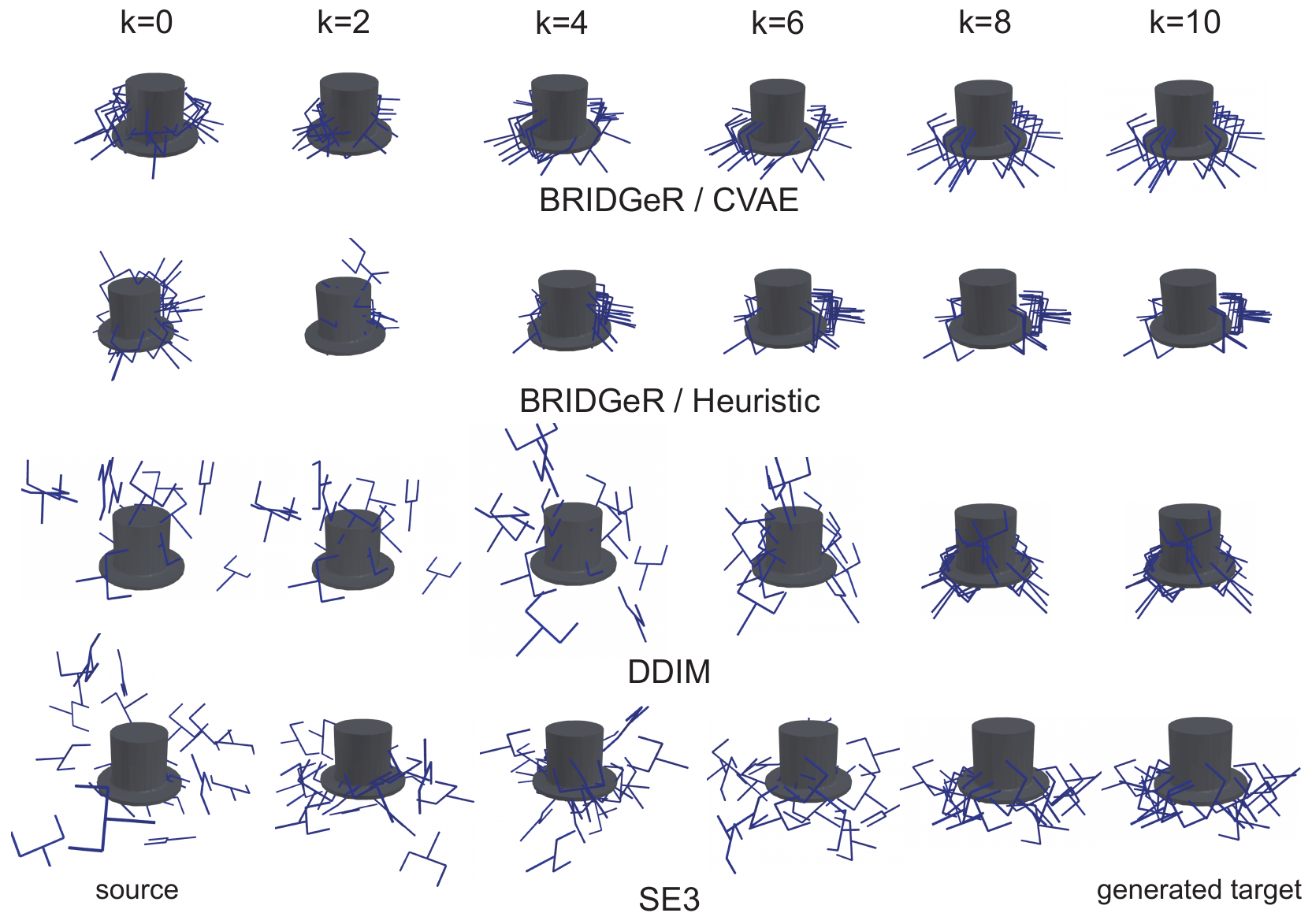}
    \caption{\small Fifteen sampled grasps across the diffusion steps for \method{}, DDIM and SE3. We visualize $15$ grasps samples every two diffusion steps until $k=10$.} 
    \label{fig:grasp_samples}
\end{figure}

In this section, we summarize our key findings, with full results and plots in Appendix \ref{app:results}. We focus on our main hypotheses and significant observations. 

\begin{figure}
    \centering
    \includegraphics[width=0.8\linewidth]{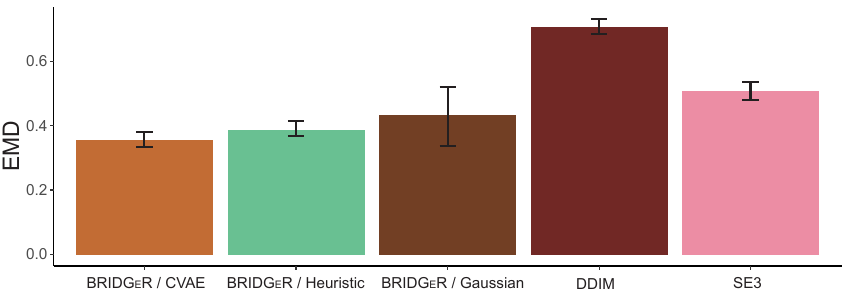}
    \caption{\small Average Earth Mover's Distance between generated grasps pose and target grasp poses on Seen Categories in Grasp Generation task ($k=160$ for SE3, DDIM and \method{}). Lower EMDs  indicate \method{} better mimics the dataset distribution.} 
    \label{fig:emd}
\end{figure}

\para{With a more informative source policy, \method{} outperforms the baselines, especially with a small of diffusion steps.} 
We observed that \method{} achieves the best success rate across the diffusion steps for many of the tasks. The differences in success rates for Adroit and Franka Kitchen tasks (Table \ref{table:d4rl_franka}) are  significant when the number of diffusion steps was small $k=5$; the exception was the Adroit pen task where DDIM performs slightly better. 
The grasp generation results in Table \ref{table:grasp} further supports this finding, where we see \method's average success rates surpassing the state-of-the-art SE3 model with $k=5$ and $k=20$. For higher diffusion steps, SE3 catches up and the methods appear comparable in terms of success rate. However, SE3 achieves poorer EMD scores compared to \method{} (Fig. \ref{fig:emd}). Fig. \ref{fig:grasp_samples} shows sample grasps at different diffusion steps; we see that using an informative source policy enables \method{} to more quickly converge to a reasonable set of grasps. 

Interestingly, we observed that \method{} consistently achieved better scores than DDIM and the Residual Policy regardless of the dataset size (See Fig. \ref{fig:main_datasize} for the Door task, with plots for other tasks in the Appendix). Potentially, larger datasets may negate the benefit afforded by the source policy since more data could enable DDIM to better generalize. 

\begin{figure}
    \centering
    \includegraphics[width=0.8\linewidth]{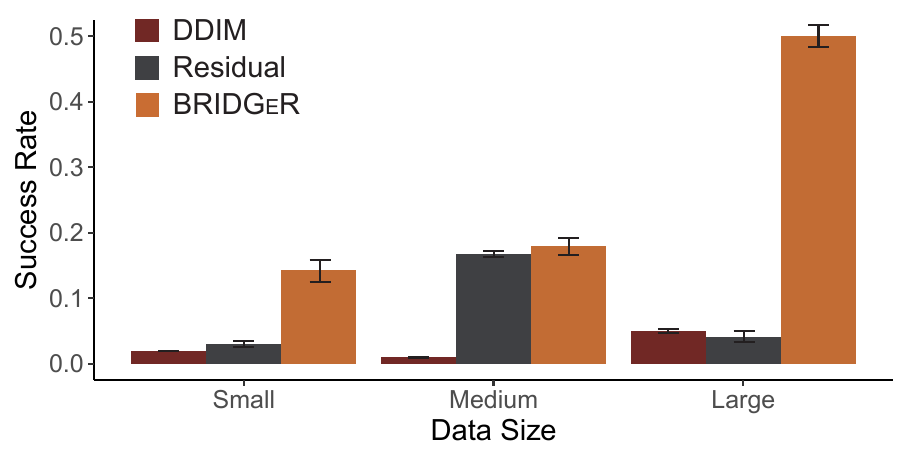}
    \caption{\small Average success rate under different training dataset on Adroit Door task ($k=80$ for DDIM and \method{}). \method{} consistently surpasses baselines across different training data size.} 
    \label{fig:main_datasize}
\end{figure}

\begin{figure}
    \centering
    \includegraphics[width=0.85\columnwidth]{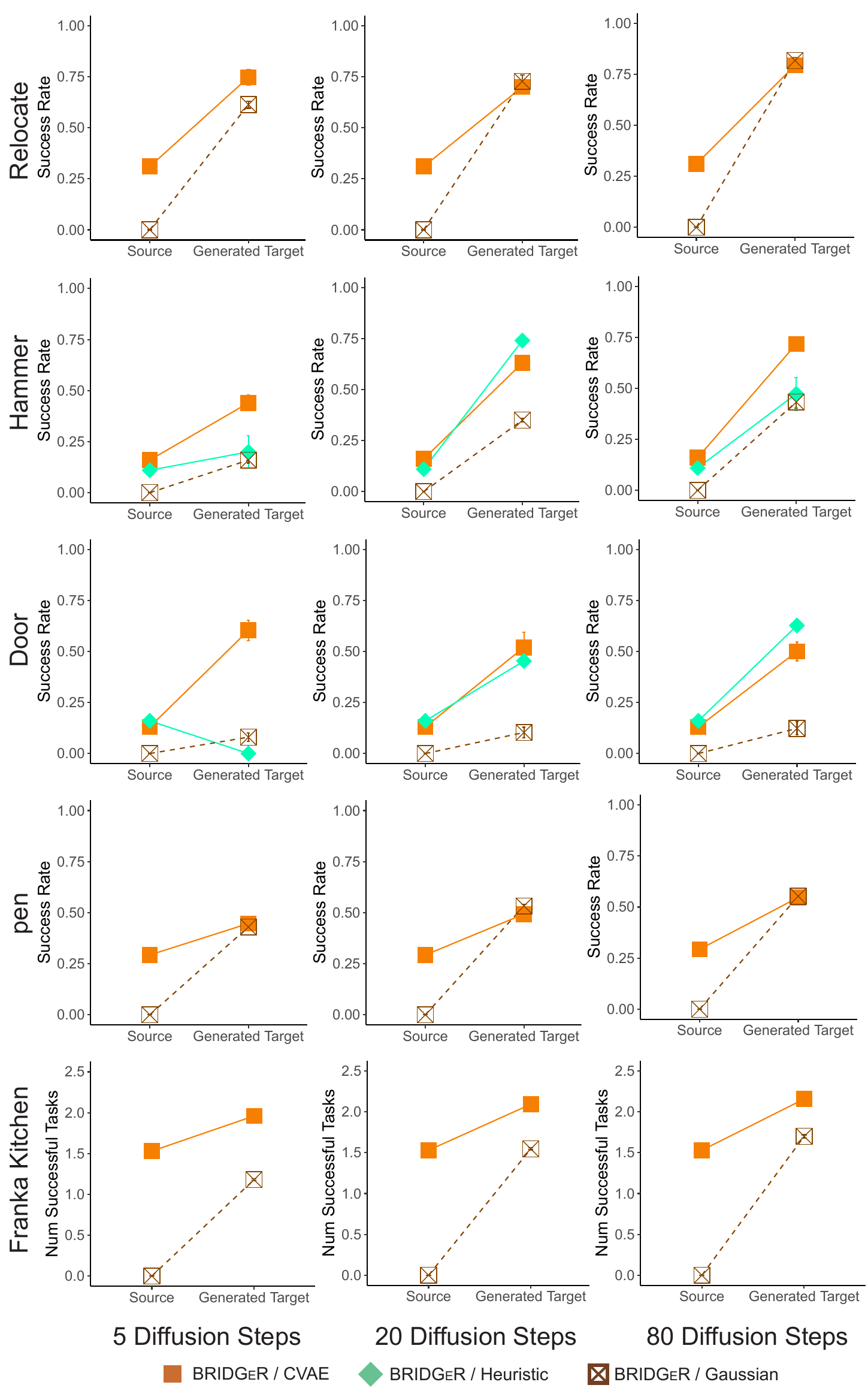}
    \caption{\small Task performance using different source policies. Each colored line shows the performance of a source policy before (``Source'') and after diffusion (``Generated Target'').}
    \label{fig:ablation_source}
    \vspace{-1em}
\end{figure}

\para{\method{} achieves better performance when using better source policies.} Fig. \ref{fig:ablation_source} illustrates the difference in success rates between the source and final action distributions. Overall, starting from a source policy with higher success rates tended to result in a better action distributions. This difference can persists even with $k=80$ steps, with potentially diminishing marginal improvement as illustrated in the Relocate and Franka Kitchen tasks. As suggested by our theoretical results, this could be due to variations in the gradual improvements over the diffusion steps.

\begin{figure}
    \centering
    \includegraphics[width=0.9\columnwidth]{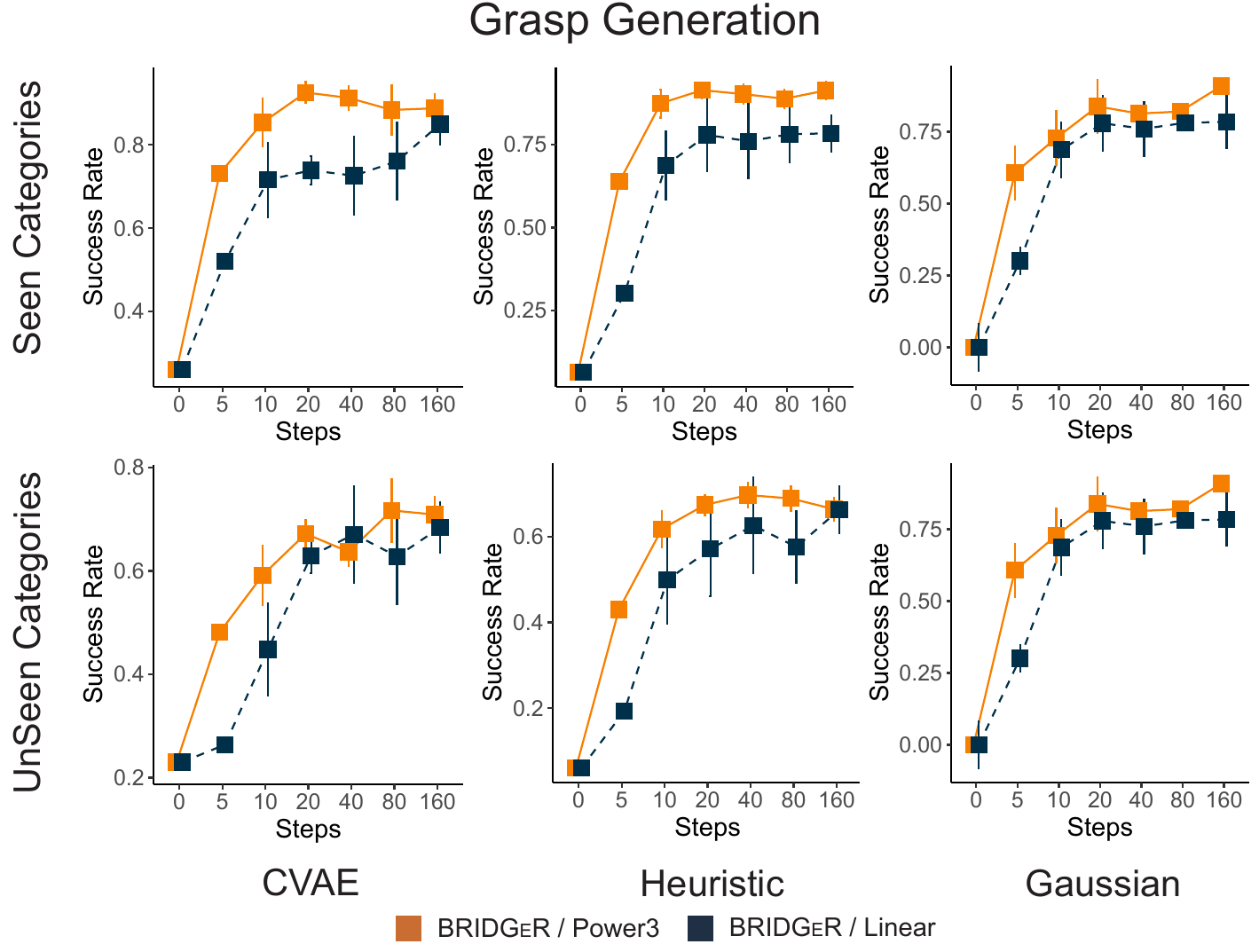}
    \caption{\small The success rate of \method{} with the Power3 and Linear interpolants for grasp generation. In general, Power3 achieves better scores when starting from the same source policy, especially when the number of diffusion steps is small. The error bars represent standard deviations over 10 test objects.}
    \label{fig:ablation_interpolant}
\end{figure}

\begin{figure}
    \centering
    \includegraphics[width=1.0\columnwidth]{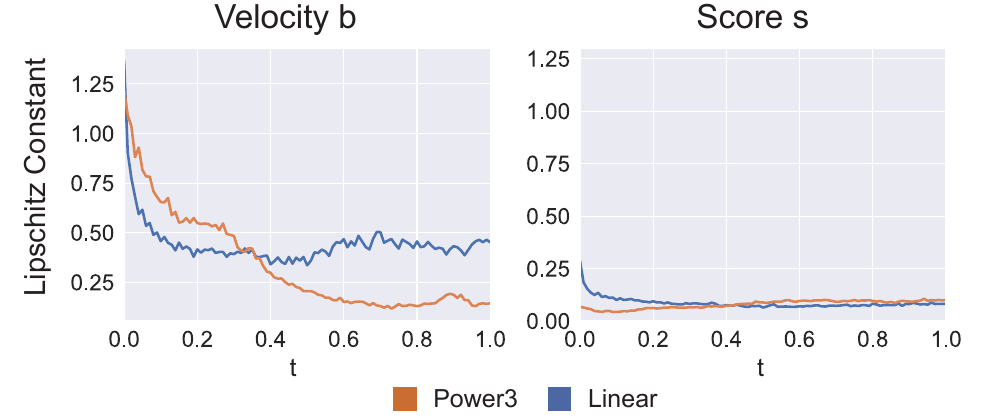}
    \caption{\small Approximate Lipschitz constants of the velocity $b$ and score $s$ for the Grasp task.}
    \label{fig:lip_constant}
    \vspace{-0.5em}
\end{figure}

\para{The Power3 interpolant is more appropriate than the Linear interpolant when behavior distributions exhibit high multi-modality.} 
We find that both interpolants perform comparably in tasks like Adroit, where robot behaviors are intricate, but largely uni-modal. In contrast, the Power3 interpolant significantly outperforms the Linear interpolant in Grasp Generation (Fig. \ref{fig:ablation_interpolant}) where the distribution of end-effector poses is highly multi-modal (Fig. \ref{fig:exp_domain}.F). This efficiency potentially stems from Power3's rapid convergence to a variety of high-density target areas, followed by fine-scale adaptation. This notion is supported by Fig. \ref{fig:lip_constant}, which illustrates the smoothness of the velocity and score functions over time (as measured by approximate Lipschitz constants~\cite{yang2023eliminating}). We observe the Lipschitz constants for Power3 to be larger at the start of the process (a ``rougher'' function) but gradually falls below that of Linear. This suggests that the function is making more rapid changes at the outset and smaller adaptations towards the middle and end of the diffusion process.

\section{Real World Robot Experiments}
\label{sec:rw_experiment}
In this section, we present findings from experiments aimed at evaluating \method{} in real-world domains with noisy and high-dimensional observations (point-cloud and image). As in the previous section, we convey our main results and refer readers to the appendix for further details. Given our simulation results, we hypothesized that \method{} would outperform existing state-of-the-art diffusion-based methods under a computation budget and conducted experiments using two tasks:

\begin{figure}
    \centering
    \includegraphics[width=1.0\linewidth]{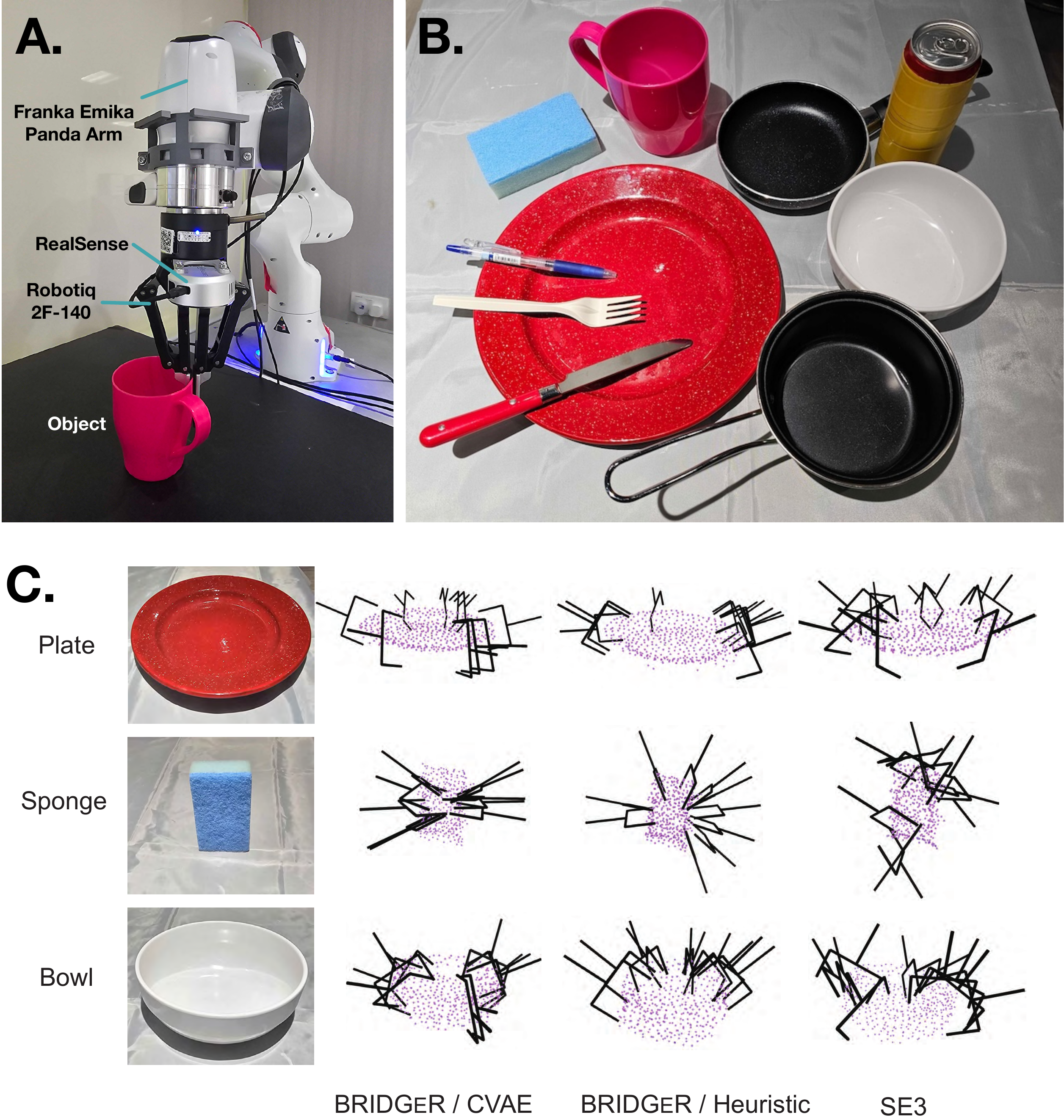}
    \caption{\small (\textbf{A}) Real-world Grasping using a Panda arm with a two-finger gripper. Observations were point clouds obtained from the RealSense Camera on the robot (\textbf{B}) Test objects used in our experiments (unseen during training). (\textbf{C}) Grasp samples from the competing models on three objects (20 diffusion steps).} 
    \label{fig:real_setup_grasp}
\end{figure}
\begin{table}
\centering
\caption{Real-World Grasping Success rate (averaged over 10 grasps on 10 test objects). Best scores shown in \textbf{bold}.}
\label{table:real_grasp_success_rate}
\addtolength{\tabcolsep}{-2pt}
\begin{tabular}{cc | cc | c}
		\toprule 
            & &  \multicolumn{2}{c|}{\method{}} & \multirow{2}{*}{SE3}
		\\  & & CVAE & Heuristic &
		\\  
		\midrule
& $k=0$  	 	& $0.07 \pm 0.12$ & $\mathbf{0.02 \pm 0.00}$ & $0.00 \pm 0.00$  
\\   
& $k=5$  	 	& $0.59 \pm 0.17$ & $\mathbf{0.66 \pm 0.13}$ & $0.05 \pm 0.07$  
\\
& $k=20$  	 	& $0.71 \pm 0.18$ & $\mathbf{0.79 \pm 0.18}$ & $0.56 \pm 0.20$  
\\
& $k=160$  	 	& $0.75 \pm 0.18$ & $\mathbf{0.81 \pm 0.24}$ & $0.73 \pm 0.21$
\\

\bottomrule

\end{tabular}

\addtolength{\tabcolsep}{+2pt}

\end{table}


\para{6-DoF Grasping (Point Cloud Observations)} where a Franka-Emika Panda arm has to grasp and lift objects (Fig. \ref{fig:real_setup_grasp}). The experiment involved $10$ everyday objects with 10 grasp trials per object. Objects were perceived using a RealSense camera that provided point cloud observations. We used the models obtained from our simulation experiments and compared \method{} (with the Power3 interpolant) against the SE3 diffusion model. We used each method to generate a set of end-effector grasp poses, and the MoveIt motion planner to plan and execute a collision-free path to grasp and lift the objects. A trial was considered successful if the robot managed to lift the object without it falling out of the robot's grasp. 

\begin{figure}
    \centering
    \includegraphics[width=1.0\linewidth]{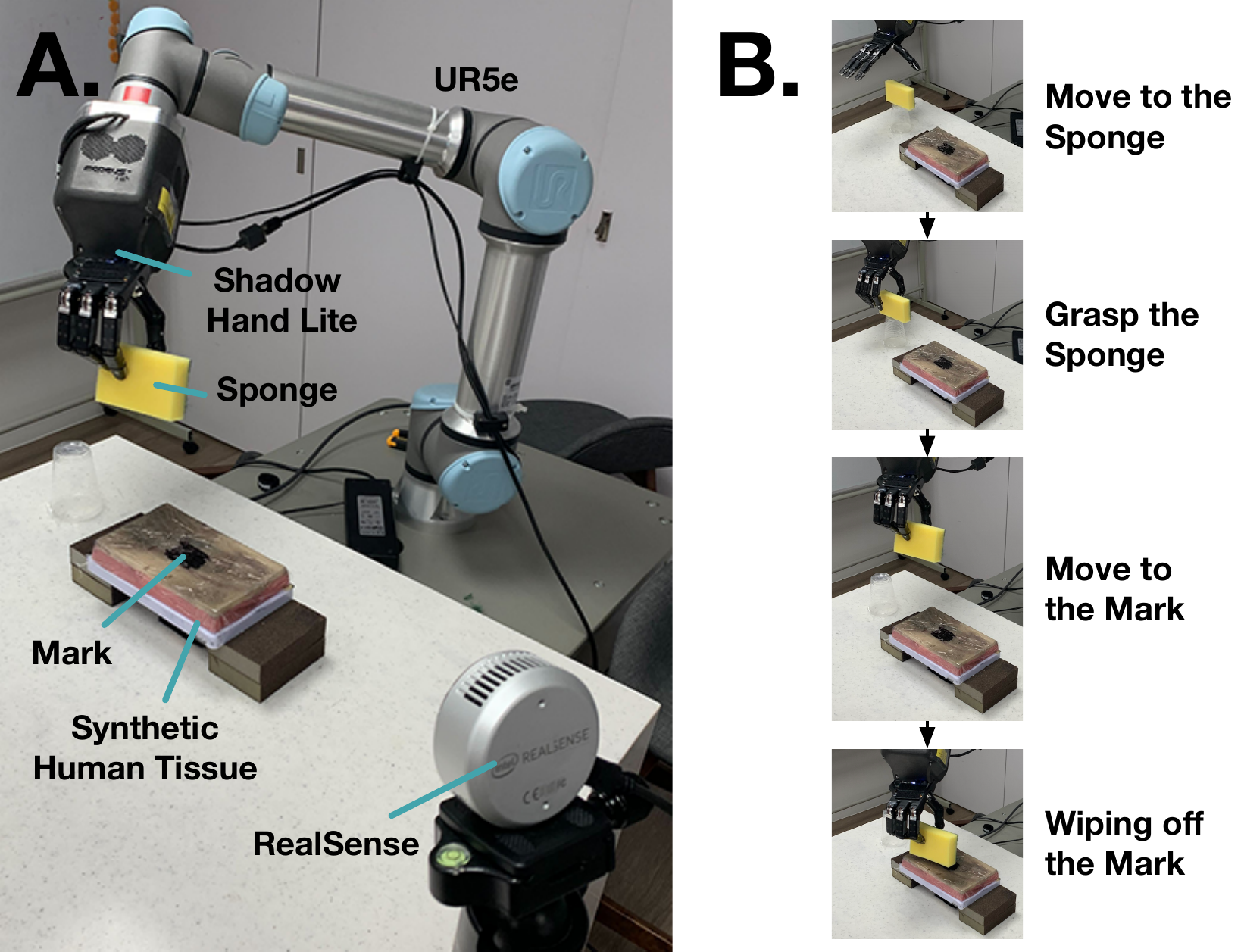}
    \caption{\small (\textbf{A}) Real-world Synthetic Wound Cleaning using a UR5e with Shadow Dextrous Hand Lite. (\textbf{B}) Demonstrations  consisted of moving the hand to the sponge from an initial position, grasping it, then manipulating the sponge to wipe off the mark. Initial positions were randomized in a region roughly 20-30cm above the sponge and the mark was in one of 9 possible positions. The robot had to learn an action policy conditioned upon RGB images from the RealSense Camera and its joint angles (both arm and hand).} 
    \label{fig:real_setup_clean}
    \vspace{-0.5em}
\end{figure}

\para{Cleaning (Image and State Observations)}. Inspired by assistive tasks in healthcare such as wound cleaning and bed-bathing, we used a Shadow Dexterous Hand Lite mounted on a UR-5e arm to perform a synthetic wound cleaning task. 
The robot had to grasp a sponge and wipe marks off a surgical practice model that mimics human tissue (Fig. \ref{fig:real_setup_clean}). This task is challenging as it involves complex high-dimensional action sequences (22 dimensions per time-step, 48 time-steps per prediction), multi-modal demonstrations (e.g., wiping off a mark either from left to right, or from top to bottom), and manipulating a deformable sponge. The mark can be at one of nine possible locations and the robot has to learn appropriate behavior conditioned on visual observations (a RGB image from a RealSense Camera) and its current joint angles. We compared \method{} against Diffusion Policy~\cite{chi2023diffusion} (DDIM) using a normalized cleaned area score, which represents how much of the mark was wiped off. We apply receding-horizon control; the models predict $48$ action steps, of which $16$ steps of actions are executed on the robot without re-planning. The models were trained using 60 demonstrations provided via kinesthetic teaching and replay.

\begin{table}
\centering
\caption{Normalized cleaned area (averaged over 9 positions) for the Cleaning Task. Best scores shown in \textbf{bold}.}
\label{table:real_skin_clean_ratio}
\addtolength{\tabcolsep}{-2pt}
\begin{tabular}{cc | cc | c}
		\toprule 
            & &  \multicolumn{2}{c|}{\method{}} & \multirow{2}{*}{DDIM}
		\\  & & CVAE & Heuristic &
		\\  
		\midrule
& $k=0$  	 	& $0.00 \pm 0.00$ & $\mathbf{0.14 \pm 0.31}$ & $0.00 \pm 0.00$  
\\   
& $k=5$  	 	& $0.46 \pm 0.42$ & $\mathbf{0.50 \pm 0.35}$ & $0.19 \pm 0.25$  
\\
& $k=20$  	 	& $0.47 \pm 0.39$ & $\mathbf{0.51 \pm 0.33}$ & $0.27 \pm 0.33$  
\\
& $k=80$  	 	& $\mathbf{0.32 \pm 0.31}$ & $0.31 \pm 0.29$ & $0.16 \pm 0.15$
\\

\bottomrule

\end{tabular}

\addtolength{\tabcolsep}{+2pt}

\end{table}
\begin{table}
\centering
\caption{Average time to successfully grasp the sponge and roughness of generated action sequences.}
\label{table:real_skin_clean_time_smooth}
\addtolength{\tabcolsep}{-2pt}
\begin{tabular}{cc | cc | c}
		\toprule 
            & &  \multicolumn{2}{c|}{\method{}} & \multirow{2}{*}{DDIM}
		\\  & & CVAE & Heuristic &
		\\  
		\midrule
& Time (Seconds) 	 	& $14.25 \pm 3.09$ & $\mathbf{13.00 \pm 1.87}$ & $36.25 \pm 11.17$  
\\
& Roughness  	 	& $0.12 \pm 0.01$ & $\mathbf{0.08 \pm 0.01}$ & $0.34 \pm 0.02$  
\\
\bottomrule

\end{tabular}

\addtolength{\tabcolsep}{+2pt}

\end{table}



\subsection{Results}

\para{\method{} outperforms the baselines, with larger gaps when the number of diffusion steps is small.} For the grasping task, \method{} achieves significantly higher success rates compared to SE3 (Table. \ref{table:real_grasp_success_rate}). Qualitatively, we observed the grasps generated by \method{} to be more accurately positioned (samples shown in Fig. \ref{fig:real_setup_grasp}), which led to more stable grasps. 

For the Cleaning task, Table. \ref{table:real_skin_clean_ratio} shows that \method{} was better at wiping off the marks compared to the DDIM diffusion policy. Interestingly, performance for all models fell at the largest number of diffusion steps ($k=80$); a potential cause is that errors accumulate during diffusion given the high-dimensional actions. Nevertheless, \method{} attains significantly higher scores. Qualitatively, we observed \method{} generates smoother action trajectories (please see the accompanying supplementary videos for examples). At $k=5$, the DDIM model produced jerkier behavior, with random movements in the arm and fingers. In contrast, \method's trajectories better mimicked the demonstrations, which led to faster completion of the task. Table \ref{table:real_skin_clean_time_smooth} summarizes this observation quantitatively; we see \method{} was quicker to grasp the sponge and had lower trajectory roughness (the average norm of the second derivative of the action trajectories).

\section{Conclusions and Future Work}
\label{sec:conclusion}
In this work, we investigate the potential of integrating informative source distributions into diffusion-style imitation learning. We provide theoretical results that support this idea and propose \method{}, a stochastic interpolant method for imitation learning. Our experiments results show that \method{} outperforms strong baselines, including state-of-the-art diffusion policies on various benchmark tasks and real-world robot experiments. We provide additional analyses on our experimental results to elucidate the effect of various design decisions within \method. 

\para{Limitations and Future Work.} Here, we have shown that leveraging prior knowledge (in the form of source policies) improves learned diffusion policy performance.
\method{} opens up avenues to explore different forms of prior knowledge, e.g., policies designed for other tasks (transfer learning) and those constructed using foundation models. Here, we explored a salient but limited set of design considerations; future work can examine more elaborate interpolant functions, along with more in-depth analysis of noise schedules and diffusion coefficients. Finally, we plan to extend \method{} incorporate considerations such as safety and user preferences. 

\section*{Acknowledgements}

This research is supported by the National Research Foundation, Singapore under its Medium Sized Center for Advanced Robotics Technology Innovation.

\balance
\bibliographystyle{plainnat}
\bibliography{references}
\clearpage
\nobalance
\appendices 
\label{sec:AppendixAdditionalResults}

\section{Proofs} 
\label{app:theory}

\contimprovethm*

\begin{proof}
    By Lagrange's Theorem, there exist $s\in[0,1]$ such that $\partial_t\phi_{F,\hat{\pi}}(s,x)=\phi_{F,\hat{\pi}}(1,x)-\phi_{F,\hat{\pi}}(0,x)$. Rearranging the terms and using the assumption that the derivative is upper bounded by $-\epsmin$, we have
    \begin{align}
        \phi_{F,\hat{\pi}}(1,x)
        &=\phi_{F,\hat{\pi}}(0,x)+\partial_t\phi_{F,\hat{\pi}}(s,x)\nonumber\\
        &\leq\phi_{F,\hat{\pi}}(0,x)-\epsmin.\label{eq:upperphi}
    \end{align}
    Similarly, there must exist $s'$ such that $\partial_t\phi_{F,\hat{\rho}}(s',x)=\phi_{F,\hat{\rho}}(1,x)-\phi_{F,\hat{\rho}}(0,x)$. Rearranging the terms and using the assumption that the derivative is lower bounded by $-\epsmax$,
    \begin{align}
        \phi_{F,\hat{\rho}}(1,x)
        &=\phi_{F,\hat{\rho}}(0,x)+\partial_t\phi_{F,\hat{\rho}}(s',x)\nonumber\\
        &\geq\phi_{F,\hat{\rho}}(0,x)-\epsmax.\label{eq:lowerphi}
    \end{align}
    Putting these results together, we have
    \begin{align*}
        &\,\phi_{F,\hat{\pi}}(1,x)-\phi_{F,\hat{\rho}}(1,x)\\
        \leq&\,(\phi_{F,\hat{\pi}}(0,x)-\epsmin)-(\phi_{F,\hat{\rho}}(0,x)-\epsmax)\\
        =&\,\phi_{F,\hat{\pi}}(0,x)-\phi_{F,\hat{\rho}}(0,x)+\epsmax-\epsmin.
    \end{align*}
    where the inequality is due to Equation~\ref{eq:upperphi} and \ref{eq:lowerphi}.
\end{proof}

\distimprovethm*

\begin{proof}
    We prove by induction. In particular, we will prove that for all $k\in\{1,\dots,K\}$, we have
    \begin{align}
        &\,\phi_{F,\hat{\pi}}(t_k,x)-\phi_{F,\hat{\rho}}(t_k,x)\nonumber\\
        \leq&\,\phi_{F,\hat{\pi}}(0,x)-\phi_{F,\hat{\rho}}(0,x)+(\epsmax-\epsmin)\sum_{i=1}^k\delta t_i.\label{eq:induction}
    \end{align}    
    When $k=1$, we have
    \begin{align*}
        &\,\phi_{F,\hat{\pi}}(t_1,x)-\phi_{F,\hat{\rho}}(t_1,x)\\
        \leq&\,(\phi_{F,\hat{\pi}}(t_0,x)-\epsmin\delta t_1)-(\phi_{F,\hat{\rho}}(t_0,x)-\epsmax\delta t_1)\\
        =&\,(\phi_{F,\hat{\pi}}(0,x)-\phi_{F,\hat{\rho}}(0,x))+(\epsmax-\epsmin)\delta t_1.
    \end{align*}
    Now assume that Equation~\ref{eq:induction} is true for $k$. Then
    \begin{align*}
        &\,\phi_{F,\hat{\pi}}(t_{k+1},x)-\phi_{F,\hat{\rho}}(t_{k+1},x)\\
        \leq&\,(\phi_{F,\hat{\pi}}(t_k,x)-\epsmin\delta t_{k+1})-(\phi_{F,\hat{\rho}}(t_k,x)-\epsmax\delta t_{k+1})\\
        =&\,(\phi_{F,\hat{\pi}}(t_k,x)-\phi_{F,\hat{\rho}}(t_k,x))+(\epsmax-\epsmin)\delta t_{k+1}\\
        \leq&\,\phi_{F,\hat{\pi}}(0,x)-\phi_{F,\hat{\rho}}(0,x)+(\epsmax-\epsmin)\sum_{i=1}^{k+1}\delta t_i.
    \end{align*}
    In particular, when $k=K$, we have
    \begin{align*}
        &\,\phi_{F,\hat{\pi}}(1,x)-\phi_{F,\hat{\rho}}(1,x)\\
        =&\,\phi_{F,\hat{\pi}}(t_{K},x)-\phi_{F,\hat{\rho}}(t_{K},x)\\
        \leq&\,\phi_{F,\hat{\pi}}(0,x)-\phi_{F,\hat{\rho}}(0,x)+(\epsmax-\epsmin)\sum_{i=1}^K\delta t_i\\
        \leq&\,\phi_{F,\hat{\pi}}(0,x)-\phi_{F,\hat{\rho}}(0,x)+\epsmax-\epsmin
    \end{align*}
    where the sum telescopes to $t_K-t_0=1$.
\end{proof}

\distcostimprovethm*

\begin{proof}
    We prove by induction. In particular, we will prove that for all $k\in\{1,\dots,K\}$, we have
    \begin{align*}
        &\,\E_{\hat{\pi}_{[k]}}[c(a|x)]-\E_{\hat{\rho}_{[k]}}[c(a|x)]\nonumber\\
        \leq&\,\E_{\hat{\pi}_{[k-1]}}[c(a|x)]-\E_{\hat{\rho}_{[k-1]}}[c(a|x)]+(\epsmax-\epsmin)\sum_{i=1}^k\delta t_i.
    \end{align*}
    where for brevity, we denote $\hat{\pi}_{t_k}$ as $\hat{\pi}_{[k]}$. Let $\epsilon=\epsmax-\epsmin$. When $k=1$, we have
    \begin{align*}
        &\,\E_{\hat{\pi}_{[1]}}[c(a|x)]-\E_{\hat{\rho}_{[1]}}[c(a|x)]\\
        =&\,\int (\hat{\pi}_{[1]}(a|x)-\hat{\rho}_{[1]}(a|x))\,c(a|x)\,\d a\\
        =&\,-\int (\hat{\pi}_{[1]}(a|x)-\hat{\rho}_{[1]}(a|x))(\ln Z-\ln\pi_1(a|x))\,\d a\\
        =&\,-\int \hat{\pi}_{[1]}(a|x)\ln\pi_1(a|x)-\hat{\rho}_{[1]}(a|x)\ln\pi_1(a|x)\,\d a\\
        =&\,\H(\hat{\pi}_{[1]}(\cdot|x),\pi_1(\cdot|x))-\H(\hat{\rho}_{[1]}(\cdot|x),\pi_1(\cdot|x))\\
        \leq&\,\H(\hat{\pi}_{[0]}(\cdot|x),\pi_1(\cdot|x))-\H(\hat{\rho}_{[0]}(\cdot|x),\pi_1(\cdot|x))+\epsilon\delta t_1\\
        =&\,-\int \hat{\pi}_{[0]}(a|x)\ln\pi_1(a|x)-\hat{\rho}_{[0]}(a|x)\ln\pi_1(a|x)\,\d a+\epsilon\delta t_1\\
        =&\,-\int (\hat{\pi}_{[0]}(a|x)-\hat{\rho}_{[0]}(a|x))(\ln Z-\ln\pi_1(a|x))\,\d a+\epsilon\delta t_1\\
        =&\,\int (\hat{\pi}_{[0]}(a|x)-\hat{\rho}_{[0]}(a|x))\,c(a|x)\,\d a+\epsilon\delta t_1\\
        =&\,\E_{\hat{\pi}_{[0]}}[c(a|x)]-\E_{\hat{\rho}_{[0]}}[c(a|x)]+(\epsmax-\epsmin)\,\delta t_1.
    \end{align*}
    The inductive case can be proven similarly. At $k=K$, the sum telescopes to 1, thus yielding our result.
\end{proof}

\section{Experiment Domains}
\label{app:expdomains}
\para{Franka Kitchen (State).} The goal of Franka Kitchen is to control a  7 DoF robot arm to interact with the various objects to reach a desired state configuration (Fig. \ref{fig:exp_domain}). The environment includes 7 objects available for interaction~\cite{fu2020d4rl}. Franka Kitchen comprises three datasets: 1) a complete dataset that includes demonstrations of all 4 target subtasks performed in a specific order, 2) a partial dataset that includes sub-trajectories where the 4 subtasks are completed sequentially, and 3) a mixed dataset, presenting various subtasks being executed, but the 4 subtasks are not completed in sequence and the demonstrated behaviors are multi-modal~\cite{fu2020d4rl}. We train the models on the mixed dataset, which is considered the most challenging among the three. The model is trained on varying data sizes: 1) small size: $16000$ sequences, 2) medium size: $32000$ sequences, and 3) large size: $64000$ sequences. 
We evaluate CVAE~\cite{sohn2015learning} source distributions trained on the same dataset (using state observations). Similar to \cite{chi2023diffusion}, the models predict action sequences with receding-horizon control ($16$ steps of actions, of which $8$ steps of actions are
executed on the robot without re-planning).

\para{Adroit (State).} Adroit encompasses the control of a 24-degree-of-freedom robotic hand to accomplish four specific tasks (Fig. \ref{fig:exp_domain}). Adroit is one of the most challenging task sets, as it requires a model to generate intricate, high-dimensional, and high-precision action sequences over long time horizons. We use the dataset of human demonstrations provided in the DAPG repository \cite{fu2020d4rl}. We evaluate a model's performance using the proportion  of task successes. An episode is considered successful if the cumulative reward exceeds a threshold ($9$ for Door and $10$ for the rest of the tasks). This threshold is selected to ensure task completion. For instance, in the case of the Adroit Door task, a reward greater than $9$ is received if the door is successfully opened. Similar to Franka Kitchen, we train the model on different data sizes: 1) small size: $1250$ sequences, 2) medium size: $2500$ sequences, and 3) large size: $5000$ sequences. We evaluate two source distributions: 1) CVAE \cite{sohn2015learning} trained on the same dataset (on $4$ tasks) and 2) hand-craft heuristic source distribution (on door and hammer). The models are trained on state observation. Similar to Franka Kitchen, we apply receding-horizon control; the models predict $64$ steps of actions, of which $48$ steps of actions are executed on the robot without re-planning.

\para{6-DoF Grasp Pose Generation (Point-Cloud).} The objective is to generate grasp poses capable of picking up an object based on the object's point cloud (Fig. \ref{fig:exp_domain}). The demonstrated behavior's distribution is intricate and highly multi-modal \cite{urain2023se}, with high-dimensional point clouds as observations. Our model is trained on $11$ object categories sourced from the Acronym dataset \cite{vahrenkamp2010integrated}, encompassing $552$ objects with approximately $200$-$2000$ grasps for each object. Similar to Adroit, we evaluate two different source distributions: 1) CVAE \cite{sohn2015learning} trained on the same dataset and 2) hand-crafted heuristic source distributions. Both source distributions use point-cloud observations. For real-world evaluation, we use the models trained in simulated grasp generation and test it on $10$ real world objects (Fig. \ref{fig:real_setup_grasp}). For each object, we evaluated $10$ grasps uniformly sampled from a set of filtered grasps. Each model generated $2000$ candidate grasps, which we filtered using hand-crated rules (e.g., we remove unreachable grasps). The ratio of filtered grasps were similar among different models (\method{} / CVAE: $57.52\%$ vs. \method{} / Heuristic: $58.12\%$ vs. SE3: $55.03\%$) 

\begin{figure}
    \centering
    \includegraphics[width=0.85\linewidth]{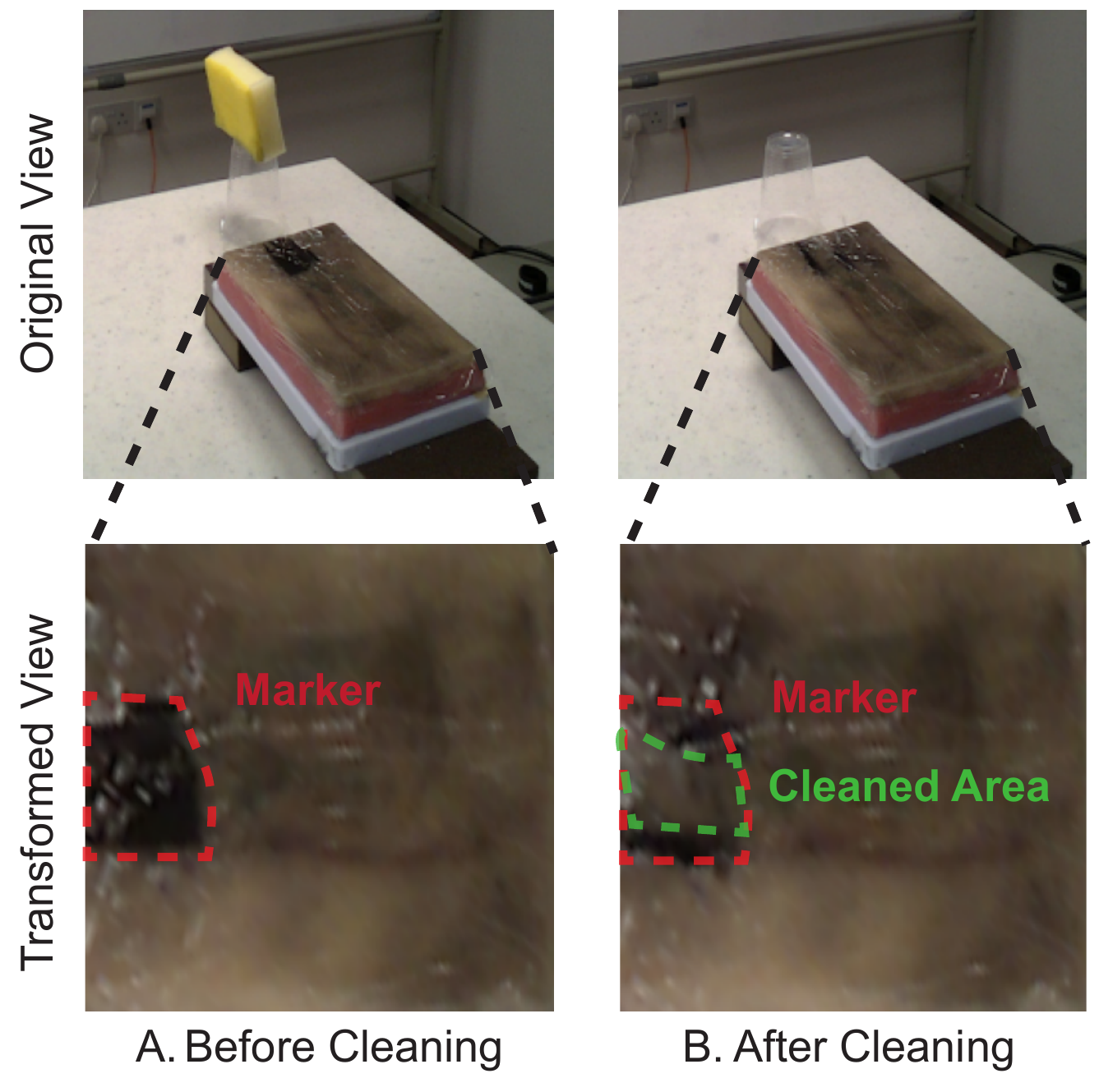}
    \caption{\small The original view of human tissue is transformed by perspective wrap. After transformation, we calculate the normalized cleaned area by the ratio of the cleaned area (green dashed border) to the original area of the marker (red dashed border). The marker and cleaned area are segmented by thresholding the pixel values.} 
    \label{fig:real_clean_metric}
\end{figure}

\para{Cleaning (Image and State).} The robot had to grasp a sponge and wipe marks off a surgical practice model that mimics human tissue. The demonstrated behavior's distribution is high-dimensional (22 dimensions of robot joint angles per time-step), multi-modal demonstrations (e.g., wiping the same mark either from left to right or from up to down), and manipulating a deformable sponge. Our model was trained on $60$ collected demonstrations. Each trajectory involves grasping the sponge and clean the markers which are randomly positioned on the surgical sample. Similar to Adroit, we evaluate two different source distributions: 1) CVAE trained on the same dataset and 2) hand-crafted heuristic source distributions. We apply receding-horizon control; the models predict $48$ steps of actions, of which $16$ steps of actions are executed on the robot without re-planning. We evalute the model performance by normalized cleaned area, which is calculated by the ratio of the area of the cleaned marker to the original area of the marker. (Fig. \ref{fig:real_clean_metric}).

\section{Source Policies}
\label{app:sourcepolicies}
As mentioned above, our experiments involve heuristic policies and data-driven CVAE policies:

\para{Heuristic policies.} We used hand-crafted policies for the following tasks:
\begin{itemize}
\item \textbf{Adroit Hammer}: the policy involves moving the hand towards the hammer, closing the hand, and subsequently rotating the wrist to swing the hammer. 
\item \textbf{Adroit Door}: the policy involves moving the hand towards the handle of the door, closing the hand, rotating the wrist, and pulling the hand back. 
\item \textbf{Grasp Generation}: We sample grasp poses directed at the object, distributing them uniformly on a sphere with its center at the object. The sphere's radius is determined by adding $0.5$ to the average distance from the normalized point cloud to its center.
\item \textbf{Cleaning}: We recorded a single fixed reference trajectory, which moves the robot arm from an initial pose to grasp the sponge and cleans a specific position on the surgical model. To sample from source policy, we compute the closest robot state (joint angles) in the reference trajectory to the current robot state, extract the subsequent $48$ actions, and add Gaussian noise. The  methods must learn to grasp the sponge from various initial poses and clean marks positioned at different locations.
\end{itemize}

\para{Data-driven policies.} We use lightweight Conditional Variational Autoencoders (CVAEs) as  representative policies. The CVAEs are trained to generate action sequences conditioned on observations specific to each domain. In Franka Kitchen and Adroit, the CVAEs are parameterized by multi-layer perceptrons with  $\approx 3\times10^6$ parameters ($10$ times fewer parameters compared to the diffusion models). In Grasp Generation, CVAE is parameterized using the Grasp SE(3)-DiffusionFields neural network~\cite{urain2023se} with $\approx 3.5\times 10^6$ parameters (approximately half of the number of parameters of the diffusion models).
In Cleaning, the CVAE is parameterized by a multi-layer perceptron with $\approx 3\times10^6$ parameters, along with image and state encoders with $\approx 1\times10^7$ parameters.
\begin{table*}
\centering
\caption{Values of $\gamma$ scale $d$ and $\epsilon$ scale $c$ in \method{} under different source policies.}
\label{table:bridger_hyper_params}
\addtolength{\tabcolsep}{-2pt}
\begin{tabular}{cc | cc | cc | cc}
		\toprule 
            & &  \multicolumn{2}{c|}{CVAE} & \multicolumn{2}{c|}{Heuristic} & \multicolumn{2}{c}{Gaussian}
		\\
		& &  $\gamma$ scale $d$  & $\epsilon$ scale $c$ & $\gamma$ scale $d$  & $\epsilon$ scale $c$ & $\gamma$ scale $d$  & $\epsilon$ scale $c$   
		\\    
		\midrule

& Door  	 	& $0.03$ & $1.0$ & $0.03$ & $1.0$  & $0.03$ & $1.0$      
\\
& Relocate  	 	& $0.3$ & $1.0$ & - & -  & $0.3$ & $1.0$      
\\
& Hammer  	 	& $0.03$ & $3.0$ & $0.3$ & $1.0$  & $0.03$ & $3.0$      
\\
& Pen  	   & $0.03$ & $1.0$ & - & -  & $0.03$ & $1.0$
\\
& Franka Kitchen  	 	& $0.03$ & $3.0$ & - & - & $0.03$ & $3.0$
\\
& Grasp Generation  	 	& $0.3$ & $1.0$ & $0.3$ & $1.0$  & $0.3$ & $1.0$  
\\
& Wound Cleaning  	 	& $0.03$ & $1.0$ & $0.03$ & $1.0$  & - & -  
\\
\bottomrule

\end{tabular}

\addtolength{\tabcolsep}{+2pt}

\end{table*}

\section{Baselines}
\label{app:baselines}
We compare \method{} Policy with the current state-of-the-art diffusion-style behavior cloning. Since \method{} can be viewed as a type of (probabilistic) residual model, we also compare it against a residual policy. For ablation studies, we evaluate \method{} under different configurations, focusing on the source distributions and interpolants. In the following, we present an overview of each methods. 

\para{DDIM.} Diffusion Policy \cite{chi2023diffusion, reuss2023goal} trains a Denoising Diffusion Probabilistic Model \cite{ho2020denoising} and applied Denoising Diffusion Implicit Model (DDIM) \cite{song2020denoising} during testing. In Adroit and Franka Kitchen tasks, the model is parameterized by the CNN-based U-net~\cite{chi2023diffusion}. In Grasp Generation tasks, the model is parameterized by Grasp SE(3)-Diffusion Fields neural network \cite{urain2023se}.
In Cleaning, the model is parameterized by the CNN-based U-net \cite{chi2023diffusion} similar to the Adroit and Franka Kitchen tasks, but with an additional image and state encoders~\cite{chi2023diffusion}.

\para{SE3 for Grasping.} Score-based diffusion model designed in SE(3) space \cite{urain2023se}. The model is specifically designed for generating 6-DoF grasp poses and parameterized by Grasp SE(3)-DiffusionFields neural network. It represents a state-of-the-art generative model for grasps.

\para{Residual Policy.} Residual Policy \cite{silver2018residual, alakuijala2021residual, johannink2019residual} takes in the sample from the source distribution as input and predicts the difference between the source sample and the target sample. The model is parameterized by the CNN-based U-net \cite{chi2023diffusion} similar to the network used in DDIM and is trained by Mean Square Error (MSE).  For Grasp Generation tasks, it is parameterized by the Grasp SE(3)-DiffusionFields neural network \cite{urain2023se}.

\para{\method{}.} Our stochastic interpolant policy. The model is parameterized by the CNN-based U-net \cite{chi2023diffusion} for the Adroit, Franka Kitchen and Cleaning (with an additional image and state encoders similar to DDIM) tasks and parameterized by Grasp SE(3)-DiffusionFields neural network \cite{urain2023se} for Grasp Generation. When compared with other baselines, we set $\gamma$ scale $d$ and $\epsilon$ scale $c$ as in Table. \ref{table:bridger_hyper_params}.

For fair comparison, each method has approximately the same number of parameters. In Adroit and Franka Kitchen, methods have $\approx 6.7\times 10^7$ parameters each. In Grasp Generation, the parameter count for each method is  $\approx 6.2 \times 10^6$ and in Cleaning, the number of parameters is $3.0\times10^7$ for each method.

\section{Model Training}
\label{app:training}
In both Adroit, Wound Cleaning and Franka Kitchen, we utilized AdamW \cite{loshchilov2017decoupled} as the optimizer. The model was trained for $3000$ epochs for Adroit and Wound Cleaning, and $5000$ epochs for Franka Kitchen. The learning rate was set to $5\times10^{-6}$ with a learning rate decay of $0.5$ every $500$ epochs. The batch size was fixed at $256$.

For Grasp Generation, we employed Adam \cite{kingma2014adam} as the optimizer. The model was trained for 4000 epochs with the learning rate $5\times10^{-4}$. Similar to Adroit and Franka Kitchen, the learning rate decay was set to $0.5$ every $500$ epochs. The batch contains $100$ objects with $80$ grasps per object.


\section{Additional Results and Plots}
\label{app:results}
In this section, we present additional results, complementing the main paper.

\para{\method{} against baselines.} Performance comparisons between \method{} v.s. the baselines on different training data sizes and diffusion steps for Adroit and Franka Kitchen are shown in Table. \ref{table:full_d4rl_franka_large_main} (large training data size), \ref{table:full_d4rl_franka_medium_main}  (medium training data size) and \ref{table:full_d4rl_franka_small_main}  (small training data size). Results for Grasp Generation with varying number of steps are presented in Table. \ref{table:full_grasp_success_main} and \ref{table:full_grasp_emd_main}. We report the results of \method{} with Power3 interpolant function and decomposed velocity $b$. The values of $\gamma$ scale and $\epsilon$ scale are listed in Table. \ref{table:bridger_hyper_params}. 

\para{\method{} with Power3 and Linear interpolant function.} The performance metrics comparing the different interpolant functions on the Adroit and Franka Kitchen tasks are shown in Table. \ref{table:full_d4rl_franka_large_interpolant}  (large training data size), \ref{table:full_d4rl_franka_medium_interpolant}  (medium training data size) and \ref{table:full_d4rl_franka_small_interpolant}  (small training data size). The results for grasp generation  presented in Table. \ref{table:full_grasp_success_interpolant} and \ref{table:full_grasp_emd_interpolant}.

\begin{figure}
    \centering
    \includegraphics[width=0.75\linewidth]{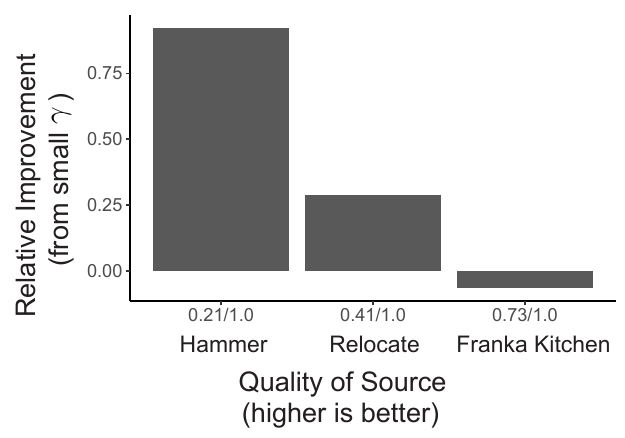}
    \caption{Relative improvement is the ratio of the source distribution's performance to the best performance achieved by \method{}. 
    Increasing the $\gamma$ scale parameter from $d=0.03$ to $d=0.3$ yields relative improvements in scenarios where the source distribution significantly deviates from the target distribution, as observed in the Hammer and Relocate tasks. This adjustment promotes exploration, enhancing performance. Conversely, in cases where the source distribution closely matches the target, such as in the Franka Kitchen task, increasing the $\gamma$ scale parameter does not contribute to performance gains. }
    \label{fig:ablation_gamma}
    \vspace{-1.5em}
\end{figure}

\begin{figure}
    \centering
    \includegraphics[width=1.0\linewidth, keepaspectratio=True]{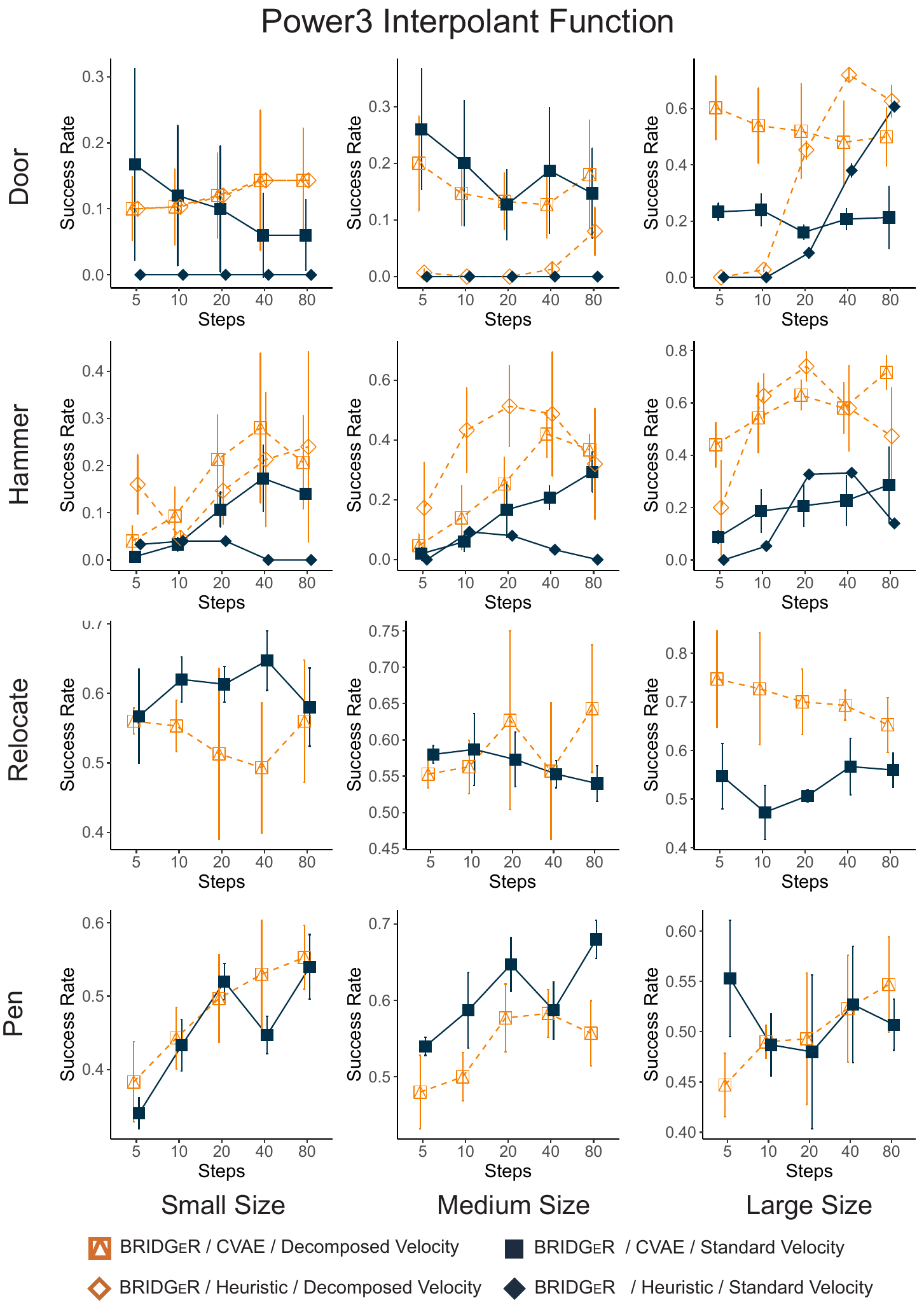}
    \caption{Success rate of \method{} / Power3 in the Adroit tasks with standard and decomposed velocity functions. Error bars represent standard deviation over three seeds.}
    \label{fig:ablation_bs_vs_power3}
\end{figure}

\begin{figure}
    \centering
    \includegraphics[width=1.0\linewidth, keepaspectratio=True]{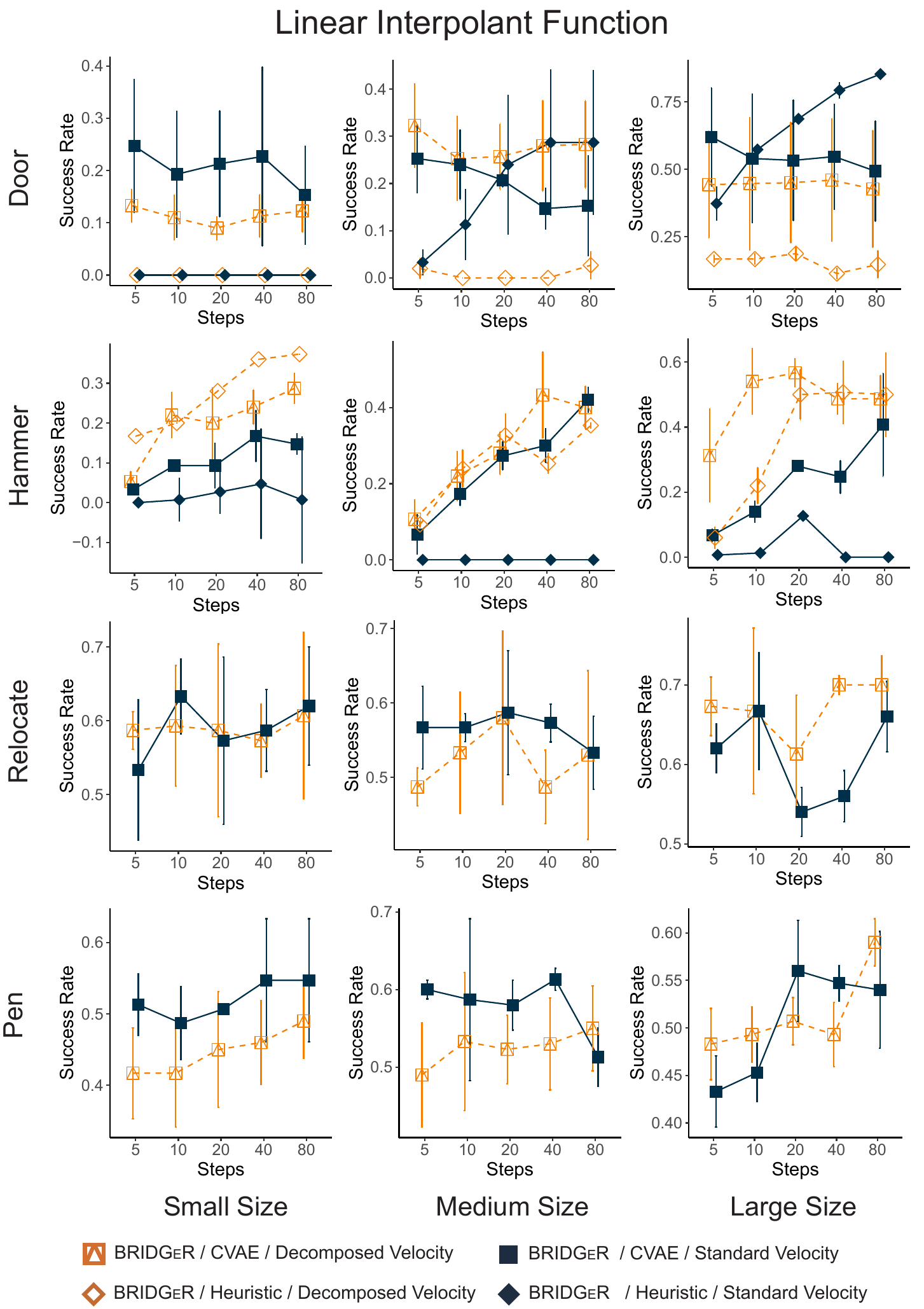}
    \caption{Success rate of \method{} / Linear in Adroit tasks with standard and decomposed velocity functions. Error bars represent standard deviation over three seeds.}
    \label{fig:ablation_bs_vs_linear}
\end{figure}

\para{Larger $\gamma$ value encourages exploration.}
Our findings indicate that increasing the $\gamma$ scale $d$ is beneficial when there is a significant discrepancy between the source and target distributions. Specifically, in experiments, the heuristic source policy in Hammer and the CVAE source policy in Relocate exhibited suboptimal performance, with relative ratios of source policy performance to the best performance of our method being $0.41$ and $0.21$, respectively. Comparing $\gamma$ scale values ($0.03$ vs. $0.3$), we observed that an increased $\gamma$ scale led to improved outcomes, as depicted in Figure \ref{fig:ablation_gamma}. We believe this is due to  larger exploration in the sampling process during training. However, when the source distribution closely resembles the target distribution, increasing $\gamma$ does not further enhance performance. This is exemplified in the Franka Kitchen scenario, where the relative ratio between source-only and optimal performance of our method stands at $0.73$, suggesting a relatively minor gap. In such contexts, our method yields comparable results regardless of the increased $\gamma$ scale, as illustrated in Figure \ref{fig:ablation_gamma}.

\begin{table*}
	\centering
 	\caption{Full results of average task performance of \method{} against current state-of-the-art methods under varying number of diffusion steps when trained with the Large dataset.}
        \label{table:full_d4rl_franka_large_main}
	\begin{tabular}{cc | ccc | cc | c}
		\toprule
		& &  \multicolumn{3}{c|}{\method{}}						  & \multicolumn{2}{c|}{Residual Policy}  & \multirow{2}{*}{DDIM} 
		\\
		&  				&  CVAE    & Heuristic   & Gaussian    & CVAE           & Heuristic  &    
		\\

		\midrule
		\multirow{6}{*}{Door}     
		& $k=0$  & $0.21 \pm 0.04$ & $\mathbf{0.22 \pm 0.06}$  & $0.00 \pm 0.00$ & $0.21 \pm 0.04$ & $\mathbf{0.22 \pm 0.06}$  & $0.00 \pm 0.00$                  
		\\     
		& $k=5$  & $\mathbf{0.60 \pm 0.15}$ & $0.00 \pm 0.00$  & $0.08 \pm 0.06$ & $0.04 \pm 0.06$ & $0.00 \pm 0.00$  & $0.02 \pm 0.02$                  
		\\
		& $k=10$ & $\mathbf{0.54 \pm 0.18}$ & $0.27 \pm 0.01$  & $0.07 \pm 0.05$ & $0.04 \pm 0.06$ & $0.00 \pm 0.00$ & $0.31 \pm 0.08$                     
		\\
		& $k=20$ & $\mathbf{0.52 \pm 0.23}$ & $0.45 \pm 0.04$  & $0.10 \pm 0.07$ & $0.04 \pm 0.06$ & $0.00 \pm 0.00$ & $0.38 \pm 0.11$                     
		\\
		& $k=40$ & $0.48 \pm 0.19$ & $\mathbf{0.72 \pm 0.02}$  & $0.07 \pm 0.05$ & $0.04 \pm 0.06$ & $0.00 \pm 0.00$ & $0.03 \pm 0.00$                     
		\\
		& $k=80$ & $0.50 \pm 0.14$ & $\mathbf{0.63 \pm 0.08}$  & $0.12 \pm 0.09$ & $0.04 \pm 0.06$ & $0.00 \pm 0.00$ & $0.05 \pm 0.02$           
		\\ 
		\midrule

		\multirow{6}{*}{Relocate}     
		& $k=0$  & $\mathbf{0.31 \pm 0.15}$ & - & $0.00 \pm 0.00$ & $\mathbf{0.31 \pm 0.15}$ & - & $0.00 \pm 0.00$                  
		\\         
		& $k=5$  & $\mathbf{0.75 \pm 0.11}$ & -	& $0.61 \pm 0.05$ & $0.30 \pm 0.04$ & -	& $0.17 \pm 0.04$                      
		\\
		& $k=10$ & $\mathbf{0.75 \pm 0.08}$ & - & $0.73 \pm 0.11$ & $0.30 \pm 0.04$ & - & $0.38 \pm 0.11$                     
		\\
		& $k=20$ & $0.70 \pm 0.11$ & - & $\mathbf{0.72 \pm 0.09}$ & $0.30 \pm 0.04$ & -	& $0.37 \pm 0.04$                       
		\\
		& $k=40$ & $0.77 \pm 0.11$ & - & $\mathbf{0.78 \pm 0.05}$ & $0.30 \pm 0.04$ & - & $0.38 \pm 0.11$                     
		\\
		& $k=80$ & $0.79 \pm 0.08$ & - & $\mathbf{0.81 \pm 0.04}$ & $0.30 \pm 0.04$ & -	& $0.26 \pm 0.03$                      
		\\ 
		\midrule
  
  		\multirow{6}{*}{Hammer}     
		& $k=0$  & $\mathbf{0.16 \pm 0.03}$ & $0.11 \pm 0.08$  & $0.00 \pm 0.00$ & $\mathbf{0.16 \pm 0.03}$ & $0.11 \pm 0.08$  & $0.00 \pm 0.00$                  
		\\         
		& $k=5$  & $\mathbf{0.44 \pm 0.11}$ & $0.20 \pm 0.24$  & $0.16 \pm 0.04$ & $0.13 \pm 0.06$ & $0.24 \pm 0.07$ & $0.01 \pm 0.01$                      
		\\
		& $k=10$ & $0.54 \pm 0.17$ & $\mathbf{0.62 \pm 0.11}$  & $0.30 \pm 0.04$ & $0.13 \pm 0.06$ & $0.24 \pm 0.07$ & $0.11 \pm 0.08$                     
		\\
		& $k=20$ & $0.63 \pm 0.08$ & $\mathbf{0.74 \pm 0.07}$  & $0.35 \pm 0.02$ & $0.13 \pm 0.06$ & $0.24 \pm 0.07$ & $0.17 \pm 0.12$                       
		\\
		& $k=40$ & $\mathbf{0.58 \pm 0.12}$ & $\mathbf{0.58 \pm 0.21}$  & $0.36 \pm 0.05$ & $0.13 \pm 0.06$ & $0.24 \pm 0.07$ & $0.01 \pm 0.01$                     
		\\
		& $k=80$ & $\mathbf{0.72 \pm 0.09}$ & $0.47 \pm 0.25$  & $0.43 \pm 0.09$ & $0.13 \pm 0.06$ & $0.24 \pm 0.07$ & $0.03 \pm 0.04$                      
		\\ 
		\midrule
		\multirow{6}{*}{Pen}     
		& $k=0$ & $\mathbf{0.29 \pm 0.05}$ & -  & $0.00 \pm 0.00$ & $\mathbf{0.29 \pm 0.05}$ & -  & $0.00 \pm 0.00$                  
		\\         
		& $k=5$ & $0.45 \pm 0.04$ & - & $0.43 \pm 0.02$ & $0.33 \pm 0.12$ & - & $\mathbf{0.54 \pm 0.03}$                       
		\\
		& $k=10$ & $0.49 \pm 0.02$ & -  & $0.46 \pm 0.04$ & $0.33 \pm 0.12$ & - & $\mathbf{0.54 \pm 0.04}$                     
		\\
		& $k=20$ & $0.49 \pm 0.09$ & - & $\mathbf{0.53 \pm 0.02}$ & $0.33 \pm 0.12$ & - & $0.51 \pm 0.03$                       
		\\
		& $k=40$ & $0.52 \pm 0.07$ & -  & $\mathbf{0.54 \pm 0.05}$ & $0.33 \pm 0.12$ & - & $0.53 \pm 0.02$                     
		\\
		& $k=80$ & $\mathbf{0.55 \pm 0.06}$ & - & $\mathbf{0.55 \pm 0.03}$ & $0.33 \pm 0.12$ & - & $0.52 \pm 0.05$                       
		\\ 
		\midrule
  
            \multirow{6}{*}{Franka Kitchen}      
		& $k=0$    & $\mathbf{1.53 \pm 0.09}$ & - & $0.00 \pm 0.00$ & $\mathbf{1.53 \pm 0.09}$ & - & $0.00 \pm 0.00$                      
		\\    
		& $k=5$    & $\mathbf{1.96 \pm 0.03}$ & - & $1.18 \pm 0.02$ & $1.55 \pm 0.10$ & - & $1.84 \pm 0.06$                      
		\\     
		& $k=10$   & $\mathbf{2.13 \pm 0.01}$ & - & $1.28 \pm 0.24$ & $1.55 \pm 0.10$ & - & $1.91 \pm 0.01$                      
		\\
		& $k=20$  & $\mathbf{2.09 \pm 0.04}$ & - & $1.54 \pm 0.03$ & $1.55 \pm 0.10$ & - & $1.93 \pm 0.07$
		\\     
		& $k=40$  & $\mathbf{2.13 \pm 0.06}$ & - & $1.67 \pm 0.01$ & $1.55 \pm 0.10$ & - & $1.99 \pm 0.04$                      
		\\
		& $k=80$  & $\mathbf{2.16 \pm 0.03}$ & - & $1.70 \pm 0.05$ & $1.55 \pm 0.10$ & - & $1.92 \pm 0.02$                      
		\\ 
		\bottomrule
		
	\end{tabular}
\end{table*}
\begin{table*}
	\centering
 	\caption{Full results of average task performance of \method{} against current state-of-the-art methods under varying number of diffusion steps when trained with the Medium dataset.}
        \label{table:full_d4rl_franka_medium_main}
	\begin{tabular}{cc | ccc | cc | c}
		\toprule
		& &  \multicolumn{3}{c|}{\method{}}						  & \multicolumn{2}{c|}{Residual Policy}  & \multirow{2}{*}{DDIM} 
		\\
		&  				&  CVAE    & Heuristic   & Gaussian    & CVAE           & Heuristic  &    
		\\

		\midrule
		\multirow{6}{*}{Door}     
		& $k=0$  & $0.20 \pm 0.08$ & $\mathbf{0.22 \pm 0.06}$  & $0.00 \pm 0.00$ & $0.20 \pm 0.08$ & $\mathbf{0.22 \pm 0.06}$  & $0.00 \pm 0.00$                  
		\\     
		& $k=5$  & $\mathbf{0.20 \pm 0.11}$ & $0.07 \pm 0.00$  & $0.08 \pm 0.00$ & $0.16 \pm 0.11$ & $0.02 \pm 0.02$  & $0.00 \pm 0.00$                  
		\\
		& $k=10$ & $0.15 \pm 0.07$ & $0.00 \pm 0.00$  & $0.02 \pm 0.02$ & $\mathbf{0.16 \pm 0.11}$ & $0.02 \pm 0.02$ & $0.02 \pm 0.01$                     
		\\
		& $k=20$ & $\mathbf{0.13 \pm 0.06}$ & $0.00 \pm 0.00$  & $0.08 \pm 0.10$ & $0.16 \pm 0.11$ & $0.02 \pm 0.02$ & $0.03 \pm 0.02$                     
		\\
		& $k=40$ & $0.13 \pm 0.08$ & $0.01 \pm 0.02$ & $0.12 \pm 0.11$ & $\mathbf{0.16 \pm 0.11}$ & $0.02 \pm 0.02$ & $0.00 \pm 0.00$                     
		\\
		& $k=80$ & $\mathbf{0.18 \pm 0.13}$ & $0.08 \pm 0.06$  & $0.13 \pm 0.11$ & $0.16 \pm 0.11$ & $0.02 \pm 0.02$ & $0.00 \pm 0.00$           
		\\ 
		\midrule

		\multirow{6}{*}{Relocate}     
		& $k=0$  & $\mathbf{0.24 \pm 0.15}$ & - & $0.00 \pm 0.00$ & $\mathbf{0.24 \pm 0.12}$ & - & $0.00 \pm 0.00$                  
		\\         
		& $k=5$  & $\mathbf{0.55 \pm 0.02}$ & -	& $0.52 \pm 0.08$ & $0.29 \pm 0.07$ & -	& $0.01 \pm 0.01$                      
		\\
		& $k=10$ & $\mathbf{0.56 \pm 0.04}$ & - & $0.61 \pm 0.09$ & $0.29 \pm 0.07$ & - & $0.09 \pm 0.01$                     
		\\
		& $k=20$ & $0.62 \pm 0.16$ & - & $\mathbf{0.63 \pm 0.12}$ & $0.29 \pm 0.07$ & -	& $0.12 \pm 0.04$                       
		\\
		& $k=40$ & $0.55 \pm 0.12$ & - & $\mathbf{0.58 \pm 0.15}$ & $0.29 \pm 0.07$ & - & $0.020 \pm 0.01$                     
		\\
		& $k=80$ & $\mathbf{0.64 \pm 0.11}$ & - & $0.54 \pm 0.10$ & $0.29 \pm 0.07$ & -	& $0.03 \pm 0.01$                      
		\\ 
		\midrule

  		\multirow{6}{*}{Hammer}     
		& $k=0$  & $0.09 \pm 0.15$ & $\mathbf{0.11 \pm 0.08}$  & $0.00 \pm 0.00$ & $0.09 \pm 0.15$ & $\mathbf{0.11 \pm 0.08}$  & $0.00 \pm 0.00$                  
		\\         
		& $k=5$  & $0.04 \pm 0.05$ & $\mathbf{0.17 \pm 0.20}$  & $0.04 \pm 0.01$ & $0.11 \pm 0.02$ & $0.12 \pm 0.05$ & $0.00 \pm 0.00$                      
		\\
		& $k=10$ & $0.14 \pm 0.14$ & $\mathbf{0.43 \pm 0.18}$  & $0.11 \pm 0.03$ & $0.11 \pm 0.02$ & $0.12 \pm 0.05$ & $0.03 \pm 0.02$                     
		\\
		& $k=20$ & $0.25 \pm 0.11$ & $\mathbf{0.51 \pm 0.18}$  & $0.24 \pm 0.03$ & $0.11 \pm 0.02$ & $0.12 \pm 0.05$ & $0.04 \pm 0.04$                       
		\\
		& $k=40$ & $0.42 \pm 0.10$ & $\mathbf{0.48 \pm 0.27}$  & $0.29 \pm 0.02$ & $0.11 \pm 0.02$ & $0.12 \pm 0.05$ & $0.00 \pm 0.00$                     
		\\
		& $k=80$ & $\mathbf{0.37 \pm 0.06}$ & $0.32 \pm 0.24$  & $0.31 \pm 0.07$ & $0.11 \pm 0.02$ & $0.12 \pm 0.05$ & $0.00 \pm 0.00$                      
		\\ 
		\midrule
  
		\multirow{6}{*}{Pen}     
		& $k=0$ & $\mathbf{0.21 \pm 0.05}$ & -  & $0.00 \pm 0.00$ & $\mathbf{0.21 \pm 0.05}$ & -  & $0.00 \pm 0.00$                  
		\\         
		& $k=5$ & $\mathbf{0.48 \pm 0.06}$ & - & $\mathbf{0.48 \pm 0.02}$ & $0.19 \pm 0.05$ & - & $0.42 \pm 0.02$                       
		\\
		& $k=10$ & $\mathbf{0.50 \pm 0.04}$ & -  & $0.46 \pm 0.07$ & $0.19 \pm 0.05$ & - & $\mathbf{0.50 \pm 0.07}$                     
		\\
		& $k=20$ & $\mathbf{0.57 \pm 0.05}$ & - & $0.46 \pm 0.05$ & $0.19 \pm 0.05$ & - & $0.49 \pm 0.03$                       
		\\
		& $k=40$ & $\mathbf{0.58 \pm 0.04}$ & -  & $0.45 \pm 0.07$ & $0.19 \pm 0.05$ & - & $0.52 \pm 0.03$                     
		\\
		& $k=80$ & $0.56 \pm 0.05$ & - & $0.49 \pm 0.06$ & $0.19 \pm 0.05$ & - & $\mathbf{0.57 \pm 0.04}$                       
		\\ 
		\midrule

            \multirow{6}{*}{Franka Kitchen}      
		& $k=0$    & $\mathbf{1.43 \pm 0.17}$ & - & $0.00 \pm 0.00$ & $\mathbf{1.43 \pm 0.17}$ & - & $0.00 \pm 0.00$                      
		\\    
		& $k=5$    & $\mathbf{1.88 \pm 0.04}$ & - & $1.57 \pm 0.14$ & $1.4 \pm 0.06$ & - & $1.82 \pm 0.08$                      
		\\     
		& $k=10$   & $\mathbf{2.04 \pm 0.02}$ & - & $1.62 \pm 0.22$ & $1.4 \pm 0.06$ & - & $1.74 \pm 0.04$                      
		\\
		& $k=20$  & $\mathbf{2.12 \pm 0.05}$ & - & $1.48 \pm 0.23$ & $1.4 \pm 0.06$ & - & $1.81 \pm 0.04$
		\\     
		& $k=40$  & $\mathbf{2.09 \pm 0.05}$ & - & $1.51 \pm 0.36$ & $1.4 \pm 0.06$ & - & $1.79 \pm 0.07$                      
		\\
		& $k=80$  & $\mathbf{2.11 \pm 0.05}$ & - & $1.44 \pm 0.28$ & $1.4 \pm 0.06$ & - & $1.79 \pm 0.06$                      
		\\ 
		\bottomrule
		
	\end{tabular}
\end{table*}
\begin{table*}
	\centering
 	\caption{Full results of average task performance of \method{} against current state-of-the-art methods under varying number of diffusion steps when trained with the Small dataset.}
        \label{table:full_d4rl_franka_small_main}
	\begin{tabular}{cc | ccc | cc | c}
		\toprule
		& &  \multicolumn{3}{c|}{\method{}}						  & \multicolumn{2}{c|}{Residual Policy}  & \multirow{2}{*}{DDIM} 
		\\
		&  				&  CVAE    & Heuristic   & Gaussian    & CVAE           & Heuristic  &    
		\\    
		\midrule

		\multirow{6}{*}{Door}     
		& $k=0$ & $0.08 \pm 0.12$ & $\mathbf{0.22 \pm 0.06}$  & $0.00 \pm 0.00$ & $0.08 \pm 0.12$ & $\mathbf{0.22 \pm 0.06}$ & $0.00 \pm 0.00$                  
		\\     
		& $k=5$ & $\mathbf{0.10 \pm 0.06}$ & $0.00 \pm 0.00$  & $0.00 \pm 0.00$ & $0.00 \pm 0.00$ & $0.01 \pm 0.02$  & $0.00 \pm 0.00$                  
		\\
		& $k=10$ & $\mathbf{0.10 \pm 0.07}$ & $0.00 \pm 0.00$  & $0.01 \pm 0.01$ & $0.00 \pm 0.00$ & $0.01 \pm 0.02$ & $0.00 \pm 0.00$                     
		\\
		& $k=20$ & $\mathbf{0.12 \pm 0.08}$ & $0.00 \pm 0.00$  & $0.05 \pm 0.02$ & $0.00 \pm 0.00$ & $0.01 \pm 0.02$ & $0.00 \pm 0.00$                     
		\\
		& $k=40$ & $\mathbf{0.14 \pm 0.14}$ & $0.00 \pm 0.00$  & $0.06 \pm 0.01$ & $0.00 \pm 0.00$ & $0.01 \pm 0.02$ & $0.00 \pm 0.00$                     
		\\
		& $k=80$ & $\mathbf{0.14 \pm 0.14}$ & $0.00 \pm 0.00$  & $0.06 \pm 0.01$ & $0.00 \pm 0.00$ & $0.01 \pm 0.02$ & $0.00 \pm 0.00$                   
		\\ 
		\midrule
  
		\multirow{6}{*}{Relocate}     
		& $k=0$  & $\mathbf{0.18 \pm 0.13}$ & - & $0.00 \pm 0.00$ & $\mathbf{0.18 \pm 0.13}$ & - & $0.00 \pm 0.00$                  
		\\         
		& $k=5$  & $\mathbf{0.56 \pm 0.06}$ & -	& $0.23 \pm 0.05$ & $0.20 \pm 0.07$ & -	& $0.00 \pm 0.00$                      
		\\
		& $k=10$ & $\mathbf{0.55 \pm 0.06}$ & - & $0.31 \pm 0.05$ & $0.20 \pm 0.07$ & - & $0.00 \pm 0.00$                     
		\\
		& $k=20$ & $\mathbf{0.51 \pm 0.03}$ & - & $0.28 \pm 0.04$ & $0.20 \pm 0.07$ & -	& $0.00 \pm 0.00$                       
		\\
		& $k=40$ & $\mathbf{0.49 \pm 0.04}$ & - & $0.37 \pm 0.06$ & $0.20 \pm 0.07$ & - & $0.00 \pm 0.00$                     
		\\
		& $k=80$ & $\mathbf{0.56 \pm 0.14}$ & - & $0.31 \pm 0.04$ & $0.20 \pm 0.07$ & -	& $0.01 \pm 0.01$                      
		\\ 
		\midrule
  
  		\multirow{6}{*}{Hammer}     
		& $k=0$  & $0.09 \pm 0.15$ & $\mathbf{0.11 \pm 0.08}$  & $0.00 \pm 0.00$ & $0.07 \pm 0.12$ & $\mathbf{0.11 \pm 0.08}$  & $0.00 \pm 0.00$                  
		\\         
		& $k=5$  & $0.04 \pm 0.04$ & $\mathbf{0.16 \pm 0.08}$  & $0.03 \pm 0.03$ & $0.07 \pm 0.04$ & $0.08 \pm 0.04$ & $0.00 \pm 0.00$                      
		\\
		& $k=10$ & $\mathbf{0.09 \pm 0.08}$ & $0.04 \pm 0.01$  & $0.06 \pm 0.02$ & $0.07 \pm 0.04$ & $0.08 \pm 0.04$ & $0.01 \pm 0.01$                     
		\\
		& $k=20$ & $\mathbf{0.21 \pm 0.12}$ & $0.14 \pm 0.09$  & $0.18 \pm 0.07$ & $0.07 \pm 0.04$ & $0.08 \pm 0.04$ & $0.01 \pm 0.01$                       
		\\
		& $k=40$ & $\mathbf{0.28 \pm 0.21}$ & $0.21 \pm 0.18$  & $0.18 \pm 0.04$ & $0.07 \pm 0.04$ & $0.08 \pm 0.04$ & $0.01 \pm 0.01$                     
		\\
		& $k=80$ & $0.21 \pm 0.13$ & $\mathbf{0.24 \pm 0.26}$  & $0.17 \pm 0.01$ & $0.07 \pm 0.04$ & $0.08 \pm 0.04$ & $0.01 \pm 0.00$                      
		\\ 
		\midrule

		\multirow{6}{*}{Pen}     
		& $k=0$ & $\mathbf{0.24 \pm 0.11}$ & -  & $0.00 \pm 0.00$ & $\mathbf{0.24 \pm 0.11}$ & -  & $0.00 \pm 0.00$                  
		\\         
		& $k=5$ & $0.38 \pm 0.07$ & - & $0.34 \pm 0.03$ & $\mathbf{0.48 \pm 0.10}$ & - & $0.41 \pm 0.05$                       
		\\
		& $k=10$ & $0.44 \pm 0.05$ & -  & $0.43 \pm 0.03$ & $\mathbf{0.48 \pm 0.10}$ & - & $0.45 \pm 0.03$                     
		\\
		& $k=20$ & $\mathbf{0.50 \pm 0.08}$ & - & $0.40 \pm 0.03$ & $0.48 \pm 0.10$ & - & $0.47 \pm 0.01$                       
		\\
		& $k=40$ & $\mathbf{0.53 \pm 0.09}$ & -  & $0.48 \pm 0.01$ & $0.48 \pm 0.10$ & - & $0.47 \pm 0.04$                     
		\\
		& $k=80$ & $\mathbf{0.55 \pm 0.05}$ & - & $0.38 \pm 0.08$ & $0.48 \pm 0.10$ & - & $0.47 \pm 0.03$                       
		\\ 
		\midrule

            \multirow{6}{*}{Franka Kitchen}      
		& $k=0$    & $\mathbf{1.43 \pm 0.17}$ & - & $0.00 \pm 0.00$ & $\mathbf{1.43 \pm 0.17}$ & - & $0.00 \pm 0.00$                      
		\\    
		& $k=5$    & $\mathbf{1.88 \pm 0.04}$ & - & $1.41 \pm 0.06$ & $1.33 \pm 0.07$ & - & $1.81 \pm 0.03$                      
		\\     
		& $k=10$   & $\mathbf{2.02 \pm 0.11}$ & - & $1.66 \pm 0.07$ & $1.33 \pm 0.07$ & - & $1.94 \pm 0.12$                      
		\\
		& $k=20$  & $\mathbf{1.99 \pm 0.06}$ & - & $1.60 \pm 0.09$ & $1.33 \pm 0.07$ & - & $1.92 \pm 0.04$
		\\     
		& $k=40$  & $\mathbf{2.01 \pm 0.03}$ & - & $1.70 \pm 0.07$ & $1.33 \pm 0.07$ & - & $1.71 \pm 0.05$                      
		\\
		& $k=80$  & $\mathbf{2.00 \pm 0.01}$ & - & $1.60 \pm 0.03$ & $1.33 \pm 0.07$ & - & $1.78 \pm 0.09$                      
		\\ 
		\bottomrule
		
	\end{tabular}
\end{table*}
\begin{table*}
\centering
\caption{Success rate (averaged over $100$ grasps on ten test objects). \method{} significantly outperforms DDIM and Residual Policy across the number of diffusion steps. Compared to SE3, \method{} achieve higher success rate when the number of diffusion steps is small. We show up to $k=160$ steps to be consistent with prior reported results~\cite{urain2023se}.}
\label{table:full_grasp_success_main}
\addtolength{\tabcolsep}{-2pt}
\begin{tabular}{cc | ccc | cc | cc}
\toprule
		& &  \multicolumn{3}{c|}{\method{}}						  & \multicolumn{2}{c|}{Residual Policy}  & \multirow{2}{*}{DDIM} & \multirow{2}{*}{SE3}
		\\
		&  				&  CVAE    &  Heuristic  & Gaussian    & CVAE &  Heuristic &    
		\\    
		\midrule

\multirow{7}{*}{Seen Categories}     
& $k=0$  	 	& $\mathbf{0.26 \pm 0.02}$ & $0.06 \pm 0.00$ & $0.00 \pm 0.00$ & $\mathbf{0.26 \pm 0.02}$  & $0.06 \pm 0.00$ & $0.00 \pm 0.00$ & $0.00 \pm 0.00$          
\\     
& $k=5$  	 	& $\mathbf{0.73 \pm 0.15}$ & $0.64 \pm 0.17$ & $0.56 \pm 0.17$ & $0.09 \pm 0.15$  & $0.01 \pm 0.03$ & $0.52 \pm 0.24$ & $0.38 \pm 0.26$          
\\
& $k=10$  	 	& $0.85 \pm 0.17$ & $\mathbf{0.87 \pm 0.12}$ & $0.72 \pm 0.27$ & $0.09 \pm 0.15$  & $0.01 \pm 0.03$ & $0.61 \pm 0.27$ & $0.56 \pm 0.20$          
\\
& $k=20$  	    & $\mathbf{0.93 \pm 0.08}$ & $0.91 \pm 0.07$ & $0.83 \pm 0.28$ & $0.09 \pm 0.15$  & $0.01 \pm 0.03$ & $0.64 \pm 0.21$ & $0.78 \pm 0.19$
\\     
& $k=40$  	 	& $\mathbf{0.91 \pm 0.08}$ & $0.90 \pm 0.08$ & $0.81 \pm 0.27$ & $0.09 \pm 0.15$  & $0.01 \pm 0.03$ & $0.61 \pm 0.26$ & $0.89 \pm 0.04$          
\\     
& $k=80$  	 	& $\mathbf{0.88 \pm 0.18}$ & $\mathbf{0.88 \pm 0.08}$ & $0.82 \pm 0.05$ & $0.09 \pm 0.15$  & $0.01 \pm 0.03$ & $0.64 \pm 0.24$ & $0.83 \pm 0.20$          
\\
& $k=160$  	    & $0.88 \pm 0.10$ & $\mathbf{0.91 \pm 0.08}$& $0.90 \pm 0.06$ & $0.09 \pm 0.15$ & $0.01 \pm 0.03$ & $0.64 \pm 0.26$ & $\mathbf{0.91 \pm 0.08}$ 							
\\
\midrule
												
\multirow{7}{*}{Unseen Categories}     
& $k=0$  	 	& $\mathbf{0.23 \pm 0.04}$ & $0.00 \pm 0.00$ & $0.00 \pm 0.00$& $\mathbf{0.23 \pm 0.04}$  & $0.00 \pm 0.00$ & $0.00 \pm 0.00$ & $0.00 \pm 0.00$          
\\     
& $k=5$  	    & $\mathbf{0.48 \pm 0.12}$ & $0.43 \pm 0.19$ & $0.45 \pm 0.14$ & $0.09 \pm 0.15$ & $0.01 \pm 0.03$ & $0.33 \pm 0.20$ & $0.20 \pm 0.10$
\\  
& $k=10$  	 	& $0.59 \pm 0.31$ & $\mathbf{0.61 \pm 0.26}$ & $0.60 \pm 0.21$& $0.09 \pm 0.15$  & $0.01 \pm 0.03$ & $0.38 \pm 0.21$ & $0.39 \pm 0.15$          
\\
& $k=20$        & $\mathbf{0.67 \pm 0.21}$ & $\mathbf{0.67 \pm 0.26}$ & $0.65 \pm 0.25$ & $0.09 \pm 0.15$ & $0.01 \pm 0.03$ & $0.41 \pm 0.23$ & $0.55 \pm 0.24$							
\\     
& $k=40$  	 	& $0.63 \pm 0.29$ & $\mathbf{0.69 \pm 0.22}$ & $\mathbf{0.69 \pm 0.20}$& $0.09 \pm 0.15$  & $0.01 \pm 0.03$ & $0.33 \pm 0.19$ & $0.61 \pm 0.21$          
\\     
& $k=80$  	 	& $\mathbf{0.71 \pm 0.20}$ & $0.68 \pm 0.25$ & $0.69 \pm 0.21$& $0.09 \pm 0.15$  & $0.01 \pm 0.03$ & $0.38 \pm 0.23$ & $0.63 \pm 0.29$          
\\
& $k=160$       & $\mathbf{0.71 \pm 0.24}$ & $0.66 \pm 0.23$ & $0.64 \pm 0.23$ & $0.09 \pm 0.15$ & $0.01 \pm 0.03$ & $0.35 \pm 0.19$ & $0.66 \pm 0.25$							
\\

\bottomrule

\end{tabular}

\addtolength{\tabcolsep}{+2pt}

\end{table*}
\begin{table*}
\centering
\caption{Average Earth Mover’s Distance between generated grasp poses and target grasp poses. \method{} significantly
outperforms SE3, DDIM and Residual Policy across the number of diffusion steps..}
\label{table:full_grasp_emd_main}
\addtolength{\tabcolsep}{-2pt}
\begin{tabular}{cc | ccc | cc | cc}
\toprule
		& &  \multicolumn{3}{c|}{\method{}}						  & \multicolumn{2}{c|}{Residual Policy}  & \multirow{2}{*}{DDIM} & \multirow{2}{*}{SE3}
		\\
		&  				& CVAE  & Heuristicx & Gaussian & CVAE & Heuristic &    
		\\    
		\midrule

\multirow{7}{*}{Seen Categories}
& $k=0$  	 	& $\mathbf{0.66 \pm 0.09}$ & $1.21 \pm 0.06$ & $1.45 \pm 0.08$& $\mathbf{0.66 \pm 0.09}$  & $1.21 \pm 0.06$ & $1.45 \pm 0.08$ & $1.45 \pm 0.08$      
\\
& $k=5$  	 	& $0.38 \pm 0.03$ & $\mathbf{0.36 \pm 0.03}$ & $0.39 \pm 0.05$& $0.69 \pm 0.02$  & $1.55 \pm 0.03$ & $1.41 \pm 0.14$ & $0.44 \pm 0.03$          
\\
& $k=10$  	 	& $0.38 \pm 0.03$ & $\mathbf{0.36 \pm 0.03}$ & $0.37 \pm 0.04$& $0.69 \pm 0.02$  & $1.55 \pm 0.03$ & $1.41 \pm 0.14$ & $0.41 \pm 0.03$          
\\
& $k=20$  	    & $0.39 \pm 0.05$ & $\mathbf{0.35 \pm 0.03}$ & $0.39 \pm 0.05$ & $0.79 \pm 0.03$  & $1.55 \pm 0.03$ & $0.64 \pm 0.21$ & $0.40 \pm 0.02$ 		
\\
& $k=40$  	 	& $0.37 \pm 0.02$ & $\mathbf{0.35 \pm 0.02}$ & $0.40 \pm 0.06$ & $0.73 \pm 0.03$  & $1.55 \pm 0.03$ & $0.52 \pm 0.24$ & $0.40 \pm 0.01$          
\\
& $k=80$  	 	& $0.38 \pm 0.02$ & $\mathbf{0.35 \pm 0.02}$ & $0.40 \pm 0.03$ & $0.74 \pm 0.02$  & $1.55 \pm 0.03$ & $0.52 \pm 0.24$ & $0.45 \pm 0.02$          
\\
& $k=160$  	    & $0.38 \pm 0.02$ & $\mathbf{0.35 \pm 0.02}$ & $0.43 \pm 0.08$ & $0.70 \pm 0.02$  &  $1.55 \pm 0.03$ & $0.64 \pm 0.26$ & $0.50 \pm 0.03$ 							
\\
\midrule

\multirow{7}{*}{Unseen Categories}     
& $k=0$  	 	& $\mathbf{0.83 \pm 0.15}$ & $1.21 \pm 0.12$ & $1.42 \pm 0.04$ & $\mathbf{0.83 \pm 0.15}$  & $1.21 \pm 0.12$ & $1.42 \pm 0.04$ & $1.38 \pm 0.02$      
\\
& $k=5$  	 	& $0.48 \pm 0.13$ & $\mathbf{0.44 \pm 0.09}$ & $0.45 \pm 0.09$ & $1.48 \pm 0.06$  & $1.57 \pm 0.06$ & $0.73 \pm 0.08$ & $0.49 \pm 0.07$          
\\
& $k=10$  	 	& $0.49 \pm 0.15$ & $\mathbf{0.45 \pm 0.11}$ & $0.44 \pm 0.09$& $1.48 \pm 0.06$  & $1.57 \pm 0.06$ & $0.80 \pm 0.10$ & $0.45 \pm 0.06$          
\\
& $k=20$  	    & $0.50 \pm 0.15$ & $\mathbf{0.42 \pm 0.10}$ & $0.48 \pm 0.11$ & $1.48 \pm 0.06$  & $1.57 \pm 0.06$ & $0.76 \pm 0.09$ & $0.45 \pm 0.09$
\\
& $k=40$  	 	& $0.49 \pm 0.13$ & $\mathbf{0.43 \pm 0.08}$ & $0.48 \pm 0.12$& $1.48 \pm 0.06$  & $1.57 \pm 0.06$ & $0.80 \pm 0.13$ & $0.47 \pm 0.06$          
\\
& $k=80$  	 	& $0.50 \pm 0.15$ & $\mathbf{0.45 \pm 0.11}$ & $0.49 \pm 0.12$& $1.48 \pm 0.06$  & $1.57 \pm 0.06$ & $0.81 \pm 0.12$ & $0.56 \pm 0.10$          
\\
& $k=160$  	    & $0.49 \pm 0.141$ & $\mathbf{0.44 \pm 0.11}$& $0.52 \pm 0.15$ & $1.48 \pm 0.06$ & $1.57 \pm 0.06$ & $0.78 \pm 0.14$ & $0.58 \pm 0.10$ 							
\\

\bottomrule

\end{tabular}

\addtolength{\tabcolsep}{+2pt}

\end{table*}

\begin{table*}
	\centering
 	\caption{Full results of average task performance of \method{} with Power3 and Linear interpolant functions under varying number of diffusion steps when trained with the Large dataset.}
        \label{table:full_d4rl_franka_large_interpolant}
	\begin{tabular}{cc | ccc | ccc}
		\toprule
		& &  \multicolumn{3}{c|}{\method{} / Power3}						  & \multicolumn{3}{c}{\method{} / Linear} 
		\\
		&  				&  CVAE    & Heuristic   & Gaussian    & CVAE           & Heuristic  &   Gaussian 
		\\ 

		\midrule
		\multirow{6}{*}{Door}     
		& $k=0$  & $0.21 \pm 0.04$ & $\mathbf{0.22 \pm 0.06}$  & $0.00 \pm 0.00$                  & $0.21 \pm 0.04$ & $\mathbf{0.22 \pm 0.06}$  & $0.00 \pm 0.00$                  
		\\     
		& $k=5$  & $\mathbf{0.60 \pm 0.15}$ & $0.00 \pm 0.00$  & $0.08 \pm 0.06$                  & $0.44 \pm 0.26$ & $0.16 \pm 0.01$  & $0.04 \pm 0.06$                  
		\\
		& $k=10$ & $\mathbf{0.54 \pm 0.18}$ & $0.27 \pm 0.01$  & $0.07 \pm 0.05$                  & $0.44 \pm 0.32$ & $0.16 \pm 0.02$ & $0.21 \pm 0.15$                     
		\\
		& $k=20$ & $\mathbf{0.52 \pm 0.23}$ & $0.45 \pm 0.04$  & $0.10 \pm 0.07$                  & $0.45 \pm 0.29$ & $0.18 \pm 0.03$ & $0.31 \pm 0.20$                     
		\\
		& $k=40$ & $0.48 \pm 0.19$ & $\mathbf{0.72 \pm 0.02}$  & $0.07 \pm 0.05$                  & $0.46 \pm 0.30$ & $0.11 \pm 0.02$ & $0.33 \pm 0.18$                     
		\\
		& $k=80$ & $0.50 \pm 0.14$ & $\mathbf{0.63 \pm 0.08}$  & $0.12 \pm 0.09$                  & $0.42 \pm 0.29$ & $0.14 \pm 0.06$ & $0.20 \pm 0.09$           
		\\ 
		\midrule
  
		\multirow{6}{*}{Relocate}     
		& $k=0$  & $\mathbf{0.31 \pm 0.15}$ & - & $0.00 \pm 0.00$ &                  $\mathbf{0.31 \pm 0.15}$ & - & $0.00 \pm 0.00$                  
		\\         
		& $k=5$  & $\mathbf{0.75 \pm 0.11}$ & -	& $0.61 \pm 0.05$ &                  $0.64 \pm 0.12$ & -	& $0.29 \pm 0.12$                      
		\\
		& $k=10$ & $\mathbf{0.75 \pm 0.08}$ & - & $0.73 \pm 0.11$ &                  $0.68 \pm 0.16$ & - & $0.74 \pm 0.04$                     
		\\
		& $k=20$ & $0.70 \pm 0.11$ & - & $0.72 \pm 0.09$ &                  $0.64 \pm 0.12$ & -	& $\mathbf{0.76 \pm 0.07}$                       
		\\
		& $k=40$ & $0.77 \pm 0.11$ & - & $\mathbf{0.78 \pm 0.05}$ &                  $0.63 \pm 0.19$ & - & $0.71 \pm 0.04$                     
		\\
		& $k=80$ & $0.79 \pm 0.08$ & - & $\mathbf{0.81 \pm 0.04}$ &                  $0.68 \pm 0.19$ & -	& $0.74 \pm 0.01$                      
		\\ 
		\midrule

  		\multirow{6}{*}{Hammer}     
		& $k=0$  & $\mathbf{0.16 \pm 0.03}$ & $0.11 \pm 0.08$  & $0.00 \pm 0.00$                  & $\mathbf{0.16 \pm 0.03}$ & $0.11 \pm 0.08$  & $0.00 \pm 0.00$                  
		\\         
		& $k=5$  & $\mathbf{0.44 \pm 0.11}$ & $0.20 \pm 0.24$  & $0.16 \pm 0.04$                  & $0.31 \pm 0.18$ & $0.06 \pm 0.04$ & $0.04 \pm 0.02$                      
		\\
		& $k=10$ & $0.54 \pm 0.17$ & $\mathbf{0.62 \pm 0.11}$  & $0.30 \pm 0.04$                  & $0.54 \pm 0.13$ & $0.22 \pm 0.07$ & $0.08 \pm 0.05$                     
		\\
		& $k=20$ & $0.63 \pm 0.08$ & $\mathbf{0.74 \pm 0.07}$  & $0.35 \pm 0.02$                  & $0.56 \pm 0.05$ & $0.50 \pm 0.10$ & $0.14 \pm 0.04$                       
		\\
		& $k=40$ & $\mathbf{0.58 \pm 0.12}$ & $\mathbf{0.58 \pm 0.21}$  & $0.36 \pm 0.05$                  & $0.48 \pm 0.06$ & $0.50 \pm 0.12$ & $0.24 \pm 0.06$                   
		\\
		& $k=80$ & $\mathbf{0.72 \pm 0.09}$ & $0.47 \pm 0.25$  & $0.43 \pm 0.09$                  & $0.48 \pm 0.09$ & $0.50 \pm 0.17$ & $0.29 \pm 0.10$                      
		\\ 
		\midrule

		\multirow{6}{*}{Pen}     
		& $k=0$ & $\mathbf{0.29 \pm 0.05}$ & -  & $0.00 \pm 0.00$                  & $\mathbf{0.29 \pm 0.05}$ & -  & $0.00 \pm 0.00$                  
		\\         
		& $k=5$ & $0.45 \pm 0.04$ & - & $0.43 \pm 0.02$                  & $\mathbf{0.48 \pm 0.05}$ & - & $0.42 \pm 0.03$                       
		\\
		& $k=10$ & $\mathbf{0.49 \pm 0.02}$ & -  & $0.46 \pm 0.04$                  & $\mathbf{0.49 \pm 0.03}$ & - & $\mathbf{0.49 \pm 0.05}$                     
		\\
		& $k=20$ & $0.49 \pm 0.09$ & - & $\mathbf{0.53 \pm 0.02}$                  & $0.51 \pm 0.03$ & - & $0.52 \pm 0.04$                       
		\\
		& $k=40$ & $0.52 \pm 0.07$ & -  & $0.54 \pm 0.05$                  & $0.49 \pm 0.04$ & - & $\mathbf{0.51 \pm 0.04}$                     
		\\
		& $k=80$ & $0.55 \pm 0.06$ & - & $0.55 \pm 0.03$                  & $\mathbf{0.59 \pm 0.03}$ & - & $0.57 \pm 0.05$                       
		\\ 
		\midrule

            \multirow{6}{*}{Franka Kitchen}      
		& $k=0$    & $\mathbf{1.53 \pm 0.09}$ & - & $0.00 \pm 0.00$                  & $\mathbf{1.53 \pm 0.09}$ & - & $0.00 \pm 0.00$                      
		\\    
		& $k=5$    & $1.96 \pm 0.03$ & - & $1.18 \pm 0.02$                  & $\mathbf{2.03 \pm 0.11}$ & - & $1.12 \pm 0.09$                      
		\\     
		& $k=10$   & $\mathbf{2.13 \pm 0.01}$ & - & $1.28 \pm 0.24$                  & $2.10 \pm 0.01$ & - & $1.55 \pm 0.09$                      
		\\
		& $k=20$  & $2.09 \pm 0.04$ & - & $1.54 \pm 0.03$                  & $\mathbf{2.11 \pm 0.05}$ & - & $1.70 \pm 0.04$
		\\     
		& $k=40$  & $\mathbf{2.13 \pm 0.06}$ & - & $1.67 \pm 0.01$                  & $2.08 \pm 0.06$ & - & $1.79 \pm 0.04$                      
		\\
		& $k=80$  & $\mathbf{2.16 \pm 0.03}$ & - & $1.70 \pm 0.05$                  & $2.06 \pm 0.06$ & - & $1.79 \pm 0.01$                      
		\\ 
		\bottomrule
		
	\end{tabular}
\end{table*}
\begin{table*}
	\centering
 	\caption{Full results of average task performance of \method{} with Power3 and Linear interpolant functions under varying number of diffusion steps when trained with the Medium dataset.}
        \label{table:full_d4rl_franka_medium_interpolant}
	\begin{tabular}{cc | ccc | ccc}
		\toprule
		& &  \multicolumn{3}{c|}{\method{} / Power3}						  & \multicolumn{3}{c}{\method{} / Linear} 
		\\
		&  				&  CVAE    & Heuristic   & Gaussian    & CVAE           & Heuristic  &   Gaussian   
		\\ 

		\midrule
		\multirow{6}{*}{Door}     
		& $k=0$  & $0.20 \pm 0.08$ & $\mathbf{0.22 \pm 0.06}$  & $0.00 \pm 0.00$                  & $0.20 \pm 0.08$ & $\mathbf{0.22 \pm 0.06}$  & $0.00 \pm 0.00$                  
		\\     
		& $k=5$  & $0.20 \pm 0.11$ & $0.07 \pm 0.00$  & $0.08 \pm 0.00$                  & $\mathbf{0.32 \pm 0.11}$ & $0.02 \pm 0.02$  & $0.04 \pm 0.05$                  
		\\
		& $k=10$ & $0.15 \pm 0.07$ & $0.00 \pm 0.00$  & $0.02 \pm 0.02$                  & $\mathbf{0.25 \pm 0.11}$ & $0.00 \pm 0.00$ & $0.22 \pm 0.12$                     
		\\
		& $k=20$ & $0.13 \pm 0.06$ & $0.00 \pm 0.00$  & $0.08 \pm 0.10$                  & $\mathbf{0.25 \pm 0.09}$ & $0.00 \pm 0.00$ & $\mathbf{0.25 \pm 0.16}$                     
		\\
		& $k=40$ & $0.13 \pm 0.08$ & $0.01 \pm 0.02$ & $0.12 \pm 0.11$                  & $\mathbf{0.28 \pm 0.12}$ & $0.00 \pm 0.00$ & $0.22 \pm 0.12$                     
		\\
		& $k=80$ & $0.18 \pm 0.13$ & $0.08 \pm 0.06$  & $0.13 \pm 0.11$                  & $\mathbf{0.28 \pm 0.12}$ & $0.02 \pm 0.03$ & $0.26 \pm 0.16$           
		\\ 
		\midrule

		\multirow{6}{*}{Relocate}     
		& $k=0$  & $\mathbf{0.24 \pm 0.15}$ & - & $0.00 \pm 0.00$                  & $\mathbf{0.24 \pm 0.12}$ & - & $0.00 \pm 0.00$                  
		\\         
		& $k=5$  & $0.55 \pm 0.02$ & -	& $0.52 \pm 0.08$                  & $\mathbf{0.63 \pm 0.09}$ & -	& $0.52 \pm 0.08$                      
		\\
		& $k=10$ & $0.56 \pm 0.04$ & - & $0.61 \pm 0.09$                  & $\mathbf{0.65 \pm 0.06}$ & - & $\mathbf{0.65 \pm 0.02}$                     
		\\
		& $k=20$ & $0.62 \pm 0.16$ & - & $0.63 \pm 0.12$                  & $0.61 \pm 0.05$ & -	& $\mathbf{0.65 \pm 0.02}$                       
		\\
		& $k=40$ & $0.55 \pm 0.12$ & - & $0.58 \pm 0.15$                  & $0.57 \pm 0.01$ & - & $\mathbf{0.60 \pm 0.08}$                     
		\\
		& $k=80$ & $0.64 \pm 0.11$ & - & $0.54 \pm 0.10$                  & $0.65 \pm 0.05$ & -	& $\mathbf{0.66 \pm 0.06}$                      
		\\ 
		\midrule

  		\multirow{6}{*}{Hammer}     
		& $k=0$  & $0.09 \pm 0.15$ & $\mathbf{0.11 \pm 0.08}$  & $0.00 \pm 0.00$                  & $0.09 \pm 0.15$ & $\mathbf{0.11 \pm 0.08}$  & $0.00 \pm 0.00$                  
		\\         
		& $k=5$  & $0.04 \pm 0.05$ & $\mathbf{0.17 \pm 0.20}$  & $0.04 \pm 0.01$                  & $0.10 \pm 0.07$ & $0.09 \pm 0.01$ & $0.03 \pm 0.01$                      
		\\
		& $k=10$ & $0.14 \pm 0.14$ & $\mathbf{0.43 \pm 0.18}$  & $0.11 \pm 0.03$                  & $0.22 \pm 0.09$ & $0.24 \pm 0.07$ & $0.20 \pm 0.02$                     
		\\
		& $k=20$ & $0.25 \pm 0.11$ & $\mathbf{0.51 \pm 0.18}$  & $0.24 \pm 0.03$                  & $0.28 \pm 0.08$ & $0.33 \pm 0.07$ & $0.22 \pm 0.07$                       
		\\
		& $k=40$ & $0.42 \pm 0.10$ & $\mathbf{0.48 \pm 0.27}$  & $0.29 \pm 0.02$                  & $0.43 \pm 0.15$ & $0.25 \pm 0.03$ & $0.36 \pm 0.05$                     
		\\
		& $k=80$ & $0.37 \pm 0.06$ & $0.32 \pm 0.24$  & $0.31 \pm 0.07$                  & $\mathbf{0.40 \pm 0.08}$ & $0.35 \pm 0.01$ & $0.28 \pm 0.10$                      
		\\ 
		\midrule

		\multirow{6}{*}{Pen}     
		& $k=0$ & $\mathbf{0.21 \pm 0.05}$ & -  & $0.00 \pm 0.00$                  & $\mathbf{0.21 \pm 0.05}$ & -  & $0.00 \pm 0.00$                  
		\\         
		& $k=5$ & $\mathbf{0.48 \pm 0.06}$ & - & $\mathbf{0.48 \pm 0.02}$                  & $\mathbf{0.48 \pm 0.05}$ & - & $0.42 \pm 0.03$                       
		\\
		& $k=10$ & $\mathbf{0.50 \pm 0.04}$ & -  & $0.46 \pm 0.07$                  & $0.49 \pm 0.03$ & - & $0.49 \pm 0.05$                     
		\\
		& $k=20$ & $\mathbf{0.57 \pm 0.05}$ & - & $0.46 \pm 0.05$                  & $0.51 \pm 0.03$ & - & $0.52 \pm 0.04$                       
		\\
		& $k=40$ & $\mathbf{0.58 \pm 0.04}$ & -  & $0.45 \pm 0.07$                  & $0.49 \pm 0.04$ & - & $0.51 \pm 0.04$                     
		\\
		& $k=80$ & $0.56 \pm 0.05$ & - & $0.49 \pm 0.06$                  & $\mathbf{0.59 \pm 0.03}$ & - & $0.56 \pm 0.05$                       
		\\ 
		\midrule

            \multirow{6}{*}{Franka Kitchen}      
		& $k=0$    & $\mathbf{1.43 \pm 0.17}$ & - & $0.00 \pm 0.00$                  & $\mathbf{1.43 \pm 0.17}$ & - & $0.00 \pm 0.00$                      
		\\    
		& $k=5$    & $1.88 \pm 0.04$ & - & $1.57 \pm 0.14$                  & $\mathbf{2.06 \pm 0.02}$ & - & $1.02 \pm 0.04$                      
		\\     
		& $k=10$   & $2.04 \pm 0.02$ & - & $1.62 \pm 0.22$                  & $\mathbf{2.09 \pm 0.01}$ & - & $1.41 \pm 0.09$                      
		\\
		& $k=20$  & $\mathbf{2.12 \pm 0.05}$ & - & $1.48 \pm 0.23$                  & $2.04 \pm 0.02$ & - & $1.75 \pm 0.02$
		\\     
		& $k=40$  & $2.09 \pm 0.05$ & - & $1.51 \pm 0.36$                  & $\mathbf{2.11 \pm 0.05}$ & - & $1.81 \pm 0.16$                      
		\\
		& $k=80$  & $\mathbf{2.11 \pm 0.05}$ & - & $1.44 \pm 0.28$                  & $2.03 \pm 0.07$ & - & $1.82 \pm 0.06$                      
		\\ 
		\bottomrule
		
	\end{tabular}
\end{table*}
\begin{table*}
	\centering
 	\caption{Full results of average task performance of \method{} with Power3 and Linear interpolant functions under varying number of diffusion steps when trained with the Small dataset.}
        \label{table:full_d4rl_franka_small_interpolant}
	\begin{tabular}{cc | ccc | ccc}
		\toprule
		& &  \multicolumn{3}{c|}{\method{} / Power3}						  & \multicolumn{3}{c}{\method{} / Linear} 
		\\
		&  				&  CVAE    & Heuristic   & Gaussian    & CVAE           & Heuristic  &   Gaussian    
		\\    
		\midrule
  
		\multirow{6}{*}{Door}     
		& $k=0$ & $0.08 \pm 0.12$ & $\mathbf{0.22 \pm 0.06}$  & $0.00 \pm 0.00$                  & $0.08 \pm 0.12$ & $\mathbf{0.22 \pm 0.06}$ & $0.00 \pm 0.00$                  
		\\     
		& $k=5$ & $0.10 \pm 0.06$ & $0.00 \pm 0.00$  & $0.00 \pm 0.00$                  & $\mathbf{0.13 \pm 0.04}$ & $0.00 \pm 0.00$  & $0.03 \pm 0.03$                  
		\\
		& $k=10$ & $0.10 \pm 0.07$ & $0.00 \pm 0.00$  & $0.01 \pm 0.01$                  & $\mathbf{0.11 \pm 0.06}$ & $0.00 \pm 0.00$ & $0.06 \pm 0.07$                     
		\\
		& $k=20$ & $\mathbf{0.12 \pm 0.08}$ & $0.00 \pm 0.00$  & $0.05 \pm 0.02$                  & $0.09 \pm 0.03$ & $0.00 \pm 0.00$ & $0.08 \pm 0.06$                     
		\\
		& $k=40$ & $\mathbf{0.14 \pm 0.14}$ & $0.00 \pm 0.00$  & $0.06 \pm 0.01$                  & $0.11 \pm 0.05$ & $0.00 \pm 0.00$ & $0.12 \pm 0.02$                     
		\\
		& $k=80$ & $\mathbf{0.14 \pm 0.14}$ & $0.00 \pm 0.00$  & $0.06 \pm 0.01$                  & $0.12 \pm 0.05$ & $0.00 \pm 0.00$ & $0.13 \pm 0.03$                   
		\\ 
		\midrule

		\multirow{6}{*}{Relocate}     
		& $k=0$  & $\mathbf{0.18 \pm 0.13}$ & - & $0.00 \pm 0.00$                  & $\mathbf{0.18 \pm 0.13}$ & - & $0.00 \pm 0.00$                  
		\\         
		& $k=5$  & $0.56 \pm 0.06$ & -	& $0.23 \pm 0.05$                  & $\mathbf{0.59 \pm 0.14}$ & -	& $0.11 \pm 0.10$                      
		\\
		& $k=10$ & $0.55 \pm 0.06$ & - & $0.31 \pm 0.05$                  & $\mathbf{0.60 \pm 0.05}$ & - & $0.59 \pm 0.16$                     
		\\
		& $k=20$ & $0.51 \pm 0.03$ & - & $0.28 \pm 0.04$                  & $0.59 \pm 0.02$ & -	& $\mathbf{0.63 \pm 0.06}$                       
		\\
		& $k=40$ & $0.49 \pm 0.04$ & - & $0.37 \pm 0.06$                  & $0.57 \pm 0.11$ & - & $\mathbf{0.58 \pm 0.04}$                     
		\\
		& $k=80$ & $0.56 \pm 0.14$ & - & $0.31 \pm 0.04$                  & $\mathbf{0.61 \pm 0.02}$ & -	& $0.55 \pm 0.02$                      
		\\ 
		\midrule
  		\multirow{6}{*}{Hammer}     
		& $k=0$  & $0.09 \pm 0.15$ & $\mathbf{0.11 \pm 0.08}$  & $0.00 \pm 0.00$                  & $0.07 \pm 0.12$ & $\mathbf{0.11 \pm 0.08}$  & $0.00 \pm 0.00$                  
		\\         
		& $k=5$  & $0.04 \pm 0.04$ & $0.16 \pm 0.08$  & $0.03 \pm 0.03$                  & $0.05 \pm 0.01$ & $\mathbf{0.17 \pm 0.01}$ & $0.04 \pm 0.06$                      
		\\
		& $k=10$ & $0.09 \pm 0.08$ & $0.04 \pm 0.01$  & $0.06 \pm 0.02$                  & $\mathbf{0.22 \pm 0.01}$ & $0.20 \pm 0.07$ & $0.12 \pm 0.10$                     
		\\
		& $k=20$ & $0.21 \pm 0.12$ & $0.14 \pm 0.09$  & $0.18 \pm 0.07$                  & $0.20 \pm 0.08$ & $\mathbf{0.28 \pm 0.07}$ & $0.19 \pm 0.11$                       
		\\
		& $k=40$ & $0.28 \pm 0.21$ & $0.21 \pm 0.18$  & $0.18 \pm 0.04$                  & $0.24 \pm 0.09$ & $\mathbf{0.36 \pm 0.18}$ & $0.17 \pm 0.11$                     
		\\
		& $k=80$ & $0.21 \pm 0.13$ & $0.24 \pm 0.26$  & $0.17 \pm 0.01$                  & $0.29 \pm 0.00$ & $\mathbf{0.37 \pm 0.21}$ & $0.21 \pm 0.08$                      
		\\ 
		\midrule

		\multirow{6}{*}{Pen}     
		& $k=0$ & $\mathbf{0.24 \pm 0.11}$ & -  & $0.00 \pm 0.00$                  & $\mathbf{0.24 \pm 0.11}$ & -  & $0.00 \pm 0.00$                  
		\\         
		& $k=5$ & $0.38 \pm 0.07$ & - & $\mathbf{0.34 \pm 0.03}$                  & $\mathbf{0.42 \pm 0.09}$ & - & $0.35 \pm 0.03$                       
		\\
		& $k=10$ & $0.44 \pm 0.05$ & -  & $0.43 \pm 0.03$                  & $0.42 \pm 0.10$ & - & $\mathbf{0.45 \pm 0.03}$                     
		\\
		& $k=20$ & $\mathbf{0.50 \pm 0.08}$ & - & $0.40 \pm 0.03$                  & $0.45 \pm 0.11$ & - & $0.47 \pm 0.01$                       
		\\
		& $k=40$ & $\mathbf{0.53 \pm 0.09}$ & -  & $0.48 \pm 0.01$                  & $0.46 \pm 0.08$ & - & $0.46 \pm 0.09$                     
		\\
		& $k=80$ & $\mathbf{0.55 \pm 0.05}$ & - & $0.38 \pm 0.08$                  & $0.49 \pm 0.07$ & - & $0.52 \pm 0.05$                       
		\\ 
		\midrule

            \multirow{6}{*}{Franka Kitchen}      
		& $k=0$    & $\mathbf{1.43 \pm 0.17}$ & - & $0.00 \pm 0.00$                  & $\mathbf{1.43 \pm 0.17}$ & - & $0.00 \pm 0.00$                      
		\\    
		& $k=5$    & $1.88 \pm 0.04$ & - & $1.41 \pm 0.06$                  & $\mathbf{2.02 \pm 0.04}$ & - & $1.04 \pm 0.07$                      
		\\     
		& $k=10$   & $\mathbf{2.02 \pm 0.11}$ & - & $1.66 \pm 0.07$                  & $\mathbf{2.02 \pm 0.04}$ & - & $1.42 \pm 0.03$                      
		\\
		& $k=20$  & $1.99 \pm 0.06$ & - & $1.60 \pm 0.09$                  & $\mathbf{2.07 \pm 0.04}$ & - & $1.69 \pm 0.05$
		\\     
		& $k=40$  & $\mathbf{2.01 \pm 0.03}$ & - & $1.70 \pm 0.07$                  & $2.00 \pm 0.02$ & - & $1.78 \pm 0.07$                      
		\\
		& $k=80$  & $\mathbf{2.00 \pm 0.01}$ & - & $1.60 \pm 0.03$                  & $1.93 \pm 0.04$ & - & $1.57 \pm 0.11$                      
		\\ 
		\bottomrule
		
	\end{tabular}
\end{table*}
\begin{table*}
\centering
\caption{Success rate (averaged over $100$ grasps on ten test objects). \method{} with Power3 interpolant function performs better than Linear interpolant function. We show up to $k=160$ steps to be consistent with prior reported results~\cite{urain2023se}.}
\label{table:full_grasp_success_interpolant}
\addtolength{\tabcolsep}{-2pt}
\begin{tabular}{cc | ccc | ccc}
		\toprule
		& &  \multicolumn{3}{c|}{\method{} / Power3} & \multicolumn{3}{c}{\method{} / Linear} 
		\\
		&  				&  CVAE    & Heuristic   & Gaussian    & CVAE           & Heuristic  &   Gaussian   
		\\    
		\midrule

\multirow{7}{*}{Seen Categories}     
& $k=0$  	 	& $\mathbf{0.26 \pm 0.02}$ & $0.06 \pm 0.00$ & $0.00 \pm 0.00$      & $\mathbf{0.26 \pm 0.02}$  & $0.06 \pm 0.00$ & $0.00 \pm 0.00$          
\\     
& $k=5$  	 	& $\mathbf{0.73 \pm 0.15}$ & $0.64 \pm 0.17$ & $0.56 \pm 0.17$      & $0.52 \pm 0.29$  & $0.31 \pm 0.28$ & $0.30 \pm 0.30$          
\\
& $k=10$  	 	& $0.85 \pm 0.17$ & $\mathbf{0.87 \pm 0.12}$ & $0.72 \pm 0.27$      & $0.71 \pm 0.30$  & $0.53 \pm 0.27$ & $0.68 \pm 0.24$          
\\
& $k=20$  	    & $\mathbf{0.93 \pm 0.08}$ & $0.91 \pm 0.07$ & $0.83 \pm 0.28$      & $0.73 \pm 0.33$  & $0.86 \pm 0.10$ & $0.77 \pm 0.14$
\\     
& $k=40$  	 	& $\mathbf{0.91 \pm 0.08}$ & $0.90 \pm 0.08$ & $0.81 \pm 0.27$      & $0.72 \pm 0.33$  & $0.77 \pm 0.28$ & $0.76 \pm 0.28$          
\\     
& $k=80$  	 	& $\mathbf{0.88 \pm 0.18}$ & $\mathbf{0.88 \pm 0.08}$ & $0.82 \pm 0.05$      & $0.76 \pm 0.25$  & $0.79 \pm 0.27$ & $0.78 \pm 0.29$          
\\
& $k=160$  	    & $0.88 \pm 0.10$ & $\mathbf{0.91 \pm 0.08}$ & $0.90 \pm 0.06$      & $0.85 \pm 0.16$ & $0.86 \pm 0.14$ & $0.78 \pm 0.28$ 							
\\
\midrule

\multirow{7}{*}{Unseen Categories}     
& $k=0$  	 	& $\mathbf{0.23 \pm 0.04}$ & $0.00 \pm 0.00$ & $0.00 \pm 0.00$       & $\mathbf{0.23 \pm 0.04}$  & $0.00 \pm 0.00$ & $0.00 \pm 0.00$         
\\     
& $k=5$  	    & $\mathbf{0.48 \pm 0.12}$ & $0.43 \pm 0.19$ & $0.45 \pm 0.14$       & $0.20 \pm 0.14$ & $0.19 \pm 0.18$ & $0.20 \pm 0.14$
\\  
& $k=10$  	 	& $0.59 \pm 0.31$ & $\mathbf{0.61 \pm 0.26}$ & $0.60 \pm 0.21$       & $0.52 \pm 0.25$  & $0.50 \pm 0.22$ & $0.52 \pm 0.25$         
\\
& $k=20$        & $\mathbf{0.67 \pm 0.21}$ & $\mathbf{0.67 \pm 0.26}$ & $0.65 \pm 0.25$       & $0.63 \pm 0.26$ & $0.57 \pm 0.24$ & $0.63 \pm 0.26$						
\\     
& $k=40$  	 	& $0.63 \pm 0.29$ & $\mathbf{0.69 \pm 0.22}$ & $\mathbf{0.69 \pm 0.20}$       & $0.63 \pm 0.27$  & $0.62 \pm 0.25$ & $0.63 \pm 0.27$         
\\     
& $k=80$  	 	& $\mathbf{0.71 \pm 0.20}$ & $0.68 \pm 0.25$ & $0.69 \pm 0.21$       & $0.60 \pm 0.28$  & $0.57 \pm 0.25$ & $0.60 \pm 0.28$        
\\
& $k=160$       & $\mathbf{0.71 \pm 0.24}$ & $0.66 \pm 0.23$ & $0.64 \pm 0.23$        & $0.60 \pm 0.31$ & $0.66 \pm 0.23$ & $0.60 \pm 0.31$						
\\

\bottomrule

\end{tabular}

\addtolength{\tabcolsep}{+2pt}

\end{table*}
\begin{table*}
\centering
\caption{Average Earth Mover’s Distance between generated grasp
poses and target grasp poses. \method{} with Power3 interpolant function performs better than Linear interpolant function.}
\label{table:full_grasp_emd_interpolant}
\addtolength{\tabcolsep}{-2pt}
\begin{tabular}{cc | ccc | ccc}
		\toprule
		& &  \multicolumn{3}{c|}{\method{} / Power3}						  & \multicolumn{3}{c}{\method{} / Linear} 
		\\
		&  				&  CVAE    & Heuristic   & Gaussian    & CVAE           & Heuristic  &   Gaussian   
		\\    
		\midrule

\multirow{7}{*}{Seen Categories}
& $k=0$  	 	& $\mathbf{0.66 \pm 0.09}$ & $1.21 \pm 0.06$ & $1.45 \pm 0.08$          & $\mathbf{0.66 \pm 0.09}$  & $1.21 \pm 0.06$ & $1.45 \pm 0.08$   
\\
& $k=5$  	 	& $0.38 \pm 0.03$ & $\mathbf{0.36 \pm 0.03}$ & $0.39 \pm 0.05$           & $0.56 \pm 0.12$  & $0.42 \pm 0.04$ & $0.42 \pm 0.04$    
\\
& $k=10$  	 	& $0.38 \pm 0.03$ & $\mathbf{0.36 \pm 0.03}$ & $0.37 \pm 0.04$           & $0.49 \pm 0.09$  & $0.42 \pm 0.06$ & $0.43 \pm 0.05$       
\\
& $k=20$  	    & $0.39 \pm 0.05$ & $\mathbf{0.35 \pm 0.03}$ & $0.39 \pm 0.05$           & $0.57 \pm 0.15$  & $0.39 \pm 0.04$ & $0.51 \pm 0.06$	
\\
& $k=40$  	 	& $0.37 \pm 0.02$ & $\mathbf{0.35 \pm 0.02}$ & $0.40 \pm 0.06$           & $0.54 \pm 0.12$  & $0.39 \pm 0.05$ & $0.51 \pm 0.12$     
\\
& $k=80$  	 	& $0.38 \pm 0.02$ & $\mathbf{0.35 \pm 0.02}$ & $0.40 \pm 0.03$           & $0.49 \pm 0.07$  & $0.39 \pm 0.05$ & $0.54 \pm 0.13$    
\\
& $k=160$  	    & $0.38 \pm 0.02$ & $\mathbf{0.35 \pm 0.02}$ & $0.43 \pm 0.08$           & $0.50 \pm 0.07$  &  $0.38 \pm 0.04$ & $0.51 \pm 0.08$						
\\
\midrule

\multirow{7}{*}{Unseen Categories}     
& $k=0$  	 	& $\mathbf{0.83 \pm 0.15}$ & $1.21 \pm 0.12$ & $1.42 \pm 0.04$              & $\mathbf{0.83 \pm 0.15}$  & $1.21 \pm 0.12$ & $1.42 \pm 0.04$      
\\
& $k=5$  	 	& $0.48 \pm 0.13$ & $\mathbf{0.44 \pm 0.09}$ & $0.45 \pm 0.09$              & $0.56 \pm 0.08$  & $0.51 \pm 0.10$ & $0.54 \pm 0.05$      
\\
& $k=10$  	 	& $0.49 \pm 0.15$ & $\mathbf{0.45 \pm 0.11}$ & $0.44 \pm 0.09$              & $0.62 \pm 0.18$  & $0.48 \pm 0.10$ & $0.54 \pm 0.10$      
\\
& $k=20$  	    & $0.50 \pm 0.15$ & $\mathbf{0.42 \pm 0.10}$ & $0.48 \pm 0.11$              & $0.64 \pm 0.18$  & $0.45 \pm 0.06$ & $0.53 \pm 0.13$
\\
& $k=40$  	 	& $0.49 \pm 0.13$ & $\mathbf{0.43 \pm 0.08}$ & $0.48 \pm 0.12$              & $0.59 \pm 0.15$  & $0.48 \pm 0.08$ & $0.57 \pm 0.11$      
\\
& $k=80$  	 	& $0.50 \pm 0.15$ & $\mathbf{0.45 \pm 0.11}$ & $0.49 \pm 0.12$              & $0.66 \pm 0.18$  & $0.45 \pm 0.08$ & $0.64 \pm 0.19$       
\\
& $k=160$  	    & $0.49 \pm 0.14$ & $\mathbf{0.44 \pm 0.11}$& $0.52 \pm 0.15$              & $0.58 \pm 0.10$ & $0.46 \pm 0.08$ & $0.69 \pm 0.11$					
\\

\bottomrule

\end{tabular}

\addtolength{\tabcolsep}{+2pt}

\end{table*}

\end{document}